\documentclass{article}
\usepackage{amssymb}

\usepackage{times} % assumes new font selection scheme installed
\usepackage{graphicx}
\usepackage{amsmath} 
\usepackage{bbm}
\usepackage{makecell}
\usepackage{multirow}
\usepackage{booktabs}
\usepackage{wrapfig}
\usepackage{here}
\usepackage[T1]{fontenc}
\usepackage{color}
\usepackage{pifont}% http://ctan.org/pkg/pifont

\usepackage{caption} % ぶち抜きeye-catchのた
\usepackage{bm}
 \usepackage{float}
 \usepackage{makecell}
\usepackage[preprint]{corl_2025} % Uncomment for the camera-ready ``final'' version.

\title{GENNAV: Polygon Mask Generation for \\ Generalized Referring Navigable Regions}

\author{
  Kei~Katsumata$^{1}$ \, 
  Yui~Iioka$^{1}$ \, 
  \textbf{Naoki~Hosomi}$^{2}$ \,
  \textbf{Teruhisa~Misu}$^{3}$ \, \\
  \textbf{Kentaro~Yamada}$^{2}$ \,
  \textbf{Komei~Sugiura}$^{1}$ \\
  $^{1}$Keio University, Japan, 
  $^{2}$Honda R\&D Co., Ltd., Japan, 
  $^{3}$Honda Research Institute USA, USA, \\
  \texttt{\{ke59ka77, kmngrd1805\}@keio.jp}, \,
  \texttt{naoki\_hosomi@jp.honda}, \\ 
  \texttt{tmisu@honda-ri.com}, \,
  \texttt{kentaro\_yamada@jp.honda}, \,
  \texttt{komei.sugiura@keio.jp}
}

\begin{document}
\maketitle

%===============================================================================

\begin{abstract}
We focus on the task of identifying the location of target regions from a natural language instruction and a front camera image captured by a mobility.
This task is challenging because it requires both existence prediction and segmentation, particularly for stuff-type target regions with ambiguous boundaries.
Existing methods often underperform in handling stuff-type target regions, in addition to absent or multiple targets.
To overcome these limitations, we propose GENNAV, which predicts target existence and generates segmentation masks for multiple stuff-type target regions. 
To evaluate GENNAV, we constructed a novel benchmark called GRiN-Drive, which includes three distinct types of samples: no-target, single-target, and multi-target.
GENNAV achieved superior performance over baseline methods on standard evaluation metrics.
Furthermore, we conducted real-world experiments with four automobiles operated in five geographically distinct urban areas to validate its zero-shot transfer performance.
In these experiments, GENNAV outperformed baseline methods and demonstrated its robustness across diverse real-world environments. The project page is available at { \url{https://gennav.vercel.app/}}.

\end{abstract}

\vspace{-4mm}
\keywords{Autonomous Driving, Vision and Language, Semantic Understanding} 
%===============================================================================
\vspace{-2mm}

\setlength{\baselineskip}{4.1mm}

\vspace{-3mm}
\section{Introduction} \label{intro}
\vspace{-3mm}

Shortages of professional labor pose critical challenges to providing equitable access to transportation, particularly where public transportation systems are fragile.
To address this issue, numerous efforts have been made to develop autonomous personal mobility that assists people with mobility impairments~\citep{rufus2021grounding, tnrsmral24}.
Such systems are expected to facilitate user-friendly interactions between users and autonomous systems.
In particular, appropriately understanding navigation instructions is crucial for enabling such interactions.
However, most existing systems exhibit limited performance in understanding navigation instructions based on the surrounding environments.

\begin{figure}[t]
    \centering
    \includegraphics[width=0.9\linewidth]{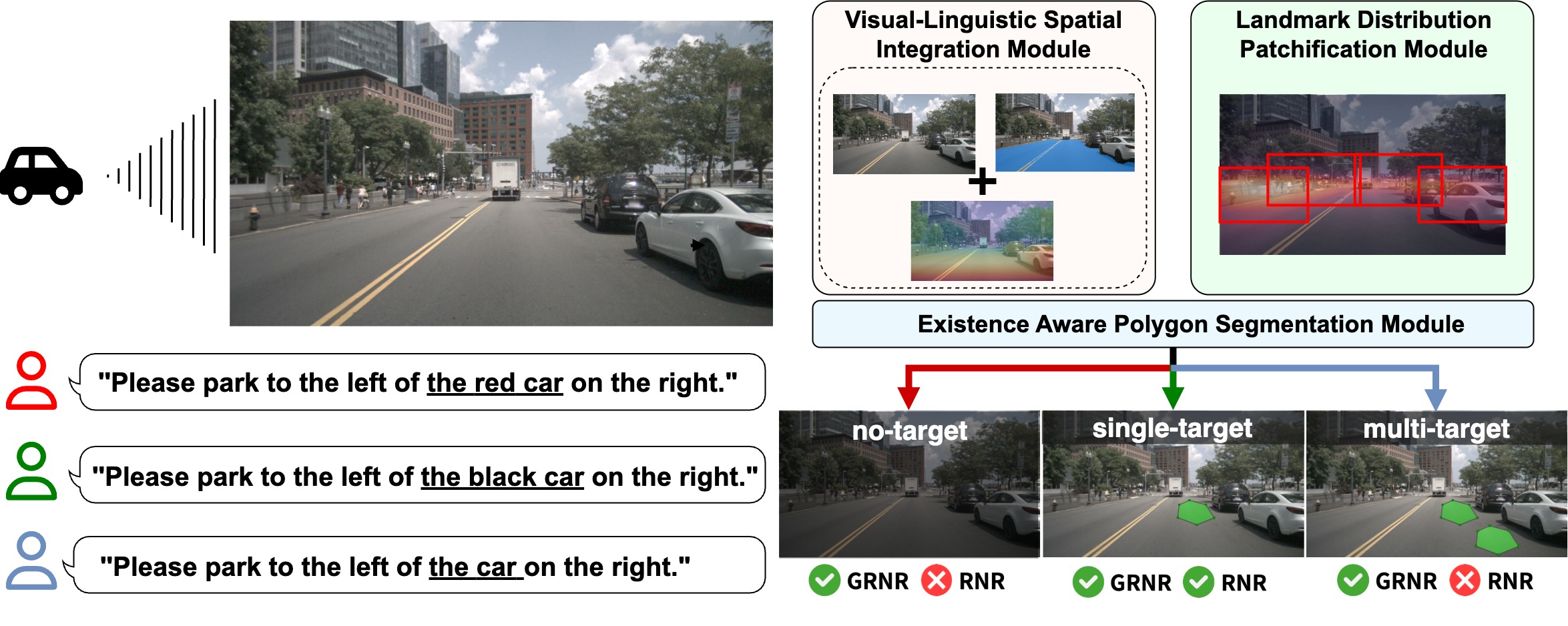}
    % \vspace{-20pt}
    \caption{
    \small
    Overview of GENNAV. 
    The model predicts target regions from a natural language instruction and a front camera image captured by a moving mobility.
    }
    \label{fig:eye-catch}
    \vspace{-6mm}
\end{figure}

% 1-2
\vspace{-2mm}
In this study, we focus on the task of identifying the location of target region(s) from both a natural language instruction and a front camera image captured by a mobility.
In autonomous driving scenarios, such instructions may not always remain valid if the landmark no longer exists in the image due to the movement of the mobility itself.
Accordingly, the model is expected to predict a ``no-target'' response when no region in the image corresponds to the instruction.
% 1-3
Fig. \ref{fig:eye-catch} shows a typical use case for our task.
In the example, when a user gives the instruction, ``Please park to the left of the black car on the right,'' the model generates a mask that corresponds to the region on the left of the black car.
By contrast, the model is expected to predict ``no-target'' when the user requests ``Park left of the red car on the right'' but no red car exists.
% 1-4

% \vspace{-8pt}
% \vspace{-8pt}

\vspace{-2mm}
Our target task is challenging because it requires both predicting the existence of target regions and generating segmentation masks for them.
This difficulty is particularly pronounced for stuff-type target regions (e.g., road) \citep{kirillov2019panoptic}, which lack clear boundaries, in contrast to thing-type regions (e.g., person, sign or mobility) that correspond to countable objects with clear boundaries.
Moreover, existing methods \citep{rufus2021grounding, tnrsmral24} including MLLMs remain insufficient, particularly in handling cases where the target is absent or multiple targets exist.
This is partially because these methods lack an explicit mechanism for predicting no-target or multi-target responses, as we show in Section \ref{sec:quantitative}.
By contrast, models that support multiple or no-target outputs~\citep{liu2023gres, xia2024gsva} are primarily designed for thing-type scenarios and do not account for the stuff-type target regions.

% 1-6
\vspace{-2mm}
To address these limitations, we propose GENNAV which explicitly predicts whether target regions exist and generates segmentation masks for stuff-type target regions. 
% 1-7
% It differs from existing approaches in the following aspect.
% It simultaneously predicts the existence of target regions and generates polygon-based segmentation masks by integrating three types of information: linguistic features, estimated road surface features, and spatially grounded multimodal representations.
% GENNAV differs from existing approaches in that it integrates polygon-based segmentation with spatially grounded multimodal representations and patchification driven by landmark distribution, which enables an appropriate semantic and spatial understanding of landmarks while maintaining computational efficiency.
GENNAV differs from existing approaches in the use of polygon based mask generation extended by existence prediction, which allows us to predict an arbitrary number of target regions in a computationally efficient manner.
Classic pixel-based methods~\citep{rufus2021grounding, tnrsmral24} incur significant computational overhead by predicting every pixel, whereas polygon-based methods~\citep{nishimura24iros, liu2023polyformer} are more efficient but typically fail in no-target cases.
In contrast, GENNAV extends existence prediction into polygon-based mask generation, effectively overcoming these limitations.
% Second, the proposed method models fine-grained visual representations related to potential landmarks by patchifying images based on the spatial distribution of landmarks.
% 1-8
% Compared to the existing methods which requires inference over all pixels, our method only needs to generates a fixed number of coordinates (e.g., 12 points), making it significantly more efficient.
% This approach not only reduces computational cost but also preserves essential features by focusing on semantically important regions.
% Our method only needs to generate a fixed number of coordinates (e.g., 12 points), making it significantly more efficient.
% This approach not only reduces computational cost but also preserves essential features by focusing on semantically important regions.
% 1-9

\vspace{-2mm}
The main contributions of this study are as follows:
\vspace{-3mm}
\begin{itemize}
    \setlength{\parskip}{0mm} % 段落間
    \setlength{\itemsep}{0.01mm} % 項目間
    \item We introduce Existence Aware Polygon Segmentation Module, which predicts the existence of target regions and generates polygon-based segmentation masks.
    % by integrating four types of information: linguistic features, estimated road surface features, fine-grained visual representations and spatially grounded multimodal representations.
    \item We incorporate Landmark Distribution Patchification Module, which models fine-grained visual representations related to potential landmarks by patchifying images based on the spatial distribution of landmarks.
    % \item We introduce Visual-Linguistic Spatial Integration Module, which models the relationship between language and spatial information by integrating features derived from an image overlaid with a pseudo-depth image and a segmentation mask of the road region, aligning with linguistic features.
    % \item We introduce two key modules: (i) the Landmark Distribution Patchification Module, which models fine-grained visual representations by patchifying images based on the spatial distribution of potential landmarks, and (ii) the Visual-Linguistic Spatial Integration Module, which models the relationship between language and spatial information.
    \item We construct the GRiN-Drive benchmark, which includes three distinct types of samples: single-target, no-target, and multi-target. 
    \item We validate the zero-shot transfer performance of GENNAV in real-world experiments involving four automobiles operating across five geographically distinct urban areas.
\end{itemize}
\vspace{-6mm}
\section{Related Work}
\vspace{-3mm}
% 2-1
Research on multimodal language processing for autonomous driving has been widely conducted~\citep{cui2024survey, datasetsurveyMingyu, huang2022multi, zhou2024vision}.
% \citet{zhou2024vision} reviews and evaluates vision-language models for autonomous driving tasks, such as object referring and vision-and-language navigation (VLN).
% Survey by~\citet{cui2024survey} highlight the potential of large language models in perception, planning, and decision-making for autonomous driving scenarios.
\citet{zhou2024vision} and \citet{cui2024survey} provide a comprehensive review of the use of LLMs and vision-language models in autonomous driving.

\vspace{-2mm}
\textbf{Visual Grounding.}
Research on visual grounding in driving scenarios has been widely conducted, encompassing various tasks such as referring expression comprehension (REC) tasks~\citep{ikun_Du_2024_CVPR, NEURIPS2023_typetotrack, deruyttere-etal-2019-talk2car, wu2023referringRMOT, zhang2024bootstrapping, kamath2021mdetr, zeng2022motr, zhang2021fairmot}, referring expression segmentation (RES) tasks~\citep{zhu2022seqtr, liu2023polyformer, cheng2024parallel, nishimura24iros, liu2023gres, xia2024gsva} and Referring Navigable Regions (RNR) tasks~\citep{rufus2021grounding, tnrsmral24}.
REC aims to predict a rectangular bounding box that localizes the object described by a natural language phrase.
Beyond single‑object settings, multi‑object and temporally dynamic variants such as Referring Multi‑Object Tracking extend REC to sequences in which several landmarks are to be tracked simultaneously~\citep{wu2023referringRMOT, zhang2024bootstrapping, ikun_Du_2024_CVPR, NEURIPS2023_typetotrack}.

\vspace{-2mm}
Although REC has made significant progress, bounding box‑level localization is often too coarse for the navigation task.
To address this limitation, 
% several segmentation tasks that incorporate linguistic instructions have been introduced.
RES models aim to generate a pixel‑level mask for the region referred to by the expression ~\citep{zhu2022seqtr, liu2023polyformer, cheng2024parallel, nishimura24iros, yang2022lavt, liu2023gres, xia2024gsva, lai2024lisa, imagespirit, hu2023beyond, luo2024cohd, li2024bring}. 
Recent polygon‑generation frameworks~\citep{zhu2022seqtr, liu2023polyformer, cheng2024parallel, nishimura24iros} reformulate RES as sequential polygon prediction and achieve promising results. 
To accommodate diverse real‑world scenarios, segmentation models that handle the existence of one or more targets have also been proposed~\citep{liu2023gres}.
Closely related to our study, the RNR task requires the segmentation of the destination region indicated by a navigation instruction~\citep{rufus2021grounding, tnrsmral24}.
To solve the RNR and related tasks, TNRSM~\citep{tnrsmral24} integrates language, image, and semantic-mask modalities in parallel to predict a destination region, whereas TRiP~\citep{hosomitrip2025} models relative positional relationships between referred landmarks and target positions.

\vspace{-2mm}
\textbf{Vision and Language Navigation.}
While most VLN studies have focused on indoor environments~\citep{anderson2018vision, zhu2021soon, qi2020reverie, zhang2024navid, krantz2020beyond}, recent works have also explored its application to autonomous driving, where vehicles follow natural language instructions to reach specified destinations~\citep{deruyttere2022talk2car, omama2023alt, shah2022lmnav, jain2023ground, xiang-etal-2020-learning, schumann-riezler-2022-analyzing}.
Such tasks often require language-conditioned localization and path planning, where appropriately grounding the target location mentioned in the instruction is critical for navigation.
Talk to the Vehicle~\citep{deruyttere2022talk2car} introduces a waypoint generation conditioned on natural language instructions.
Furthermore, some approaches integrate foundation models  to facilitate free-form and zero-shot navigation~\citep{omama2023alt, shah2022lmnav}. 
For instance, Grounded then Navigate~\citep{jain2023ground} uses CLIP~\citep{dosovitskiy2021an} to identify navigable regions aligned with language instructions.
While VLN focuses on the generation of a sequence of steps to reach a destination, RNR and GRNR require the generation of segmentation masks that directly identify the target region within an image,  which serves as the designated goal.

\vspace{-1mm}
\textbf{Benchmarks.} The survey by~\citet{datasetsurveyMingyu} reviews benchmark datasets for autonomous‑driving scenes and highlights representative benchmarks such as the BDD100K~\citep{yu2020bdd100k}, Cityscapes~\citep{cordts2016cityscapes}, KITTI~\citep{kittiGeiger2012CVPR}, and nuScenes~\citep{caesar2020nuscenes}.
% BDD100K comprises roughly 100k driving videos collected from more than 50k rides and supports a suite of tasks that includes object detection, tracking, semantic segmentation, and lane detection.
% Cityscapes is a standard benchmark for semantic segmentation that provides fine‑grained, pixel‑level annotations for 30 visual classes across diverse urban scenes.
The KITTI vision benchmark suite is a comprehensive driving-scene dataset with ground truth for a variety of tasks, such as semantic segmentation and depth estimation.
Refer‑KITTI~\citep{wu2023referringRMOT} and Refer‑KITTI-V2~\citep{zhang2024bootstrapping} were built on top of KITTI to establish the RMOT benchmark in which an arbitrary number of objects are tracked according to the given expression.
The nuScenes dataset contains full‑surround data across various cities and weather conditions, and is widely adopted for multi‑view 3D object‑detection benchmarks.
Built on top of nuScenes, Talk2Car~\citep{deruyttere-etal-2019-talk2car} introduces navigation commands that contain landmark‑based referring expressions together with the locations of the referred landmarks and Talk2Car‑RegSeg~\citep{rufus2021grounding} further augments the dataset with pixel‑wise segmentation masks of the intended navigable regions.
\vspace{-3mm}
\section{Method} \label{method}
\vspace{-2mm}
% 4-1
% \section{Problem Statements
% \label{sec:problem}
% }
\subsection{Problem Statements}
\vspace{-2mm}
% 3-1
In this paper, we focus on the Generalized Referring Navigable Regions (GRNR) task, which predicts both the existence and location of target regions based on a natural language instruction and a front camera image taken from a moving mobility. 
% 3-2
In this task, the model is expected to determine whether any target region specified in the navigation instruction exists in the given image.
If one or more target regions exist, the model is required to generate a segmentation mask for each target region.
% 3-3

\vspace{-2mm}
Fig. \ref{fig:eye-catch} shows typical scenes from the GRNR task.
Unlike the RNR task, the GRNR task involves instructions that specify any number of target regions, including multi-target and no-target instructions. 
In this task, we do not focus on trajectory prediction that corresponds to the given target region because existing lower-level functionalities can already perform such trajectory prediction sufficiently in practical applications.
Further details of the task examples are available in the supplementary material.

\vspace{-2mm}
The terminology used in this paper is defined as follows.
An ``navigation instruction'' refers to an instruction in natural language to navigate the mobility to the destination.
The ``target region'' refers to the region specified by a given navigation instruction. The relationship between the instruction and the target region follows a one-to-many mapping (if any).
``Landmark'' is an object used as a reference when specifying a target region.
% Terms used in this study are defined as follows:
% A ``Navigation Instruction'' refers to instruction in natural language to navigate the mobility to the destination, and a ``target region'' refers to the region specified by a given navigation instruction. The relationship between the instruction and the target region follows a one-to-many mapping (if any).
% In addition, a ``landmark'' refers to an object used as a reference when specifying a target region.
% We use the following evaluation metrics: msIoU (mean stuff intersection over union), $\mathrm{P}@K$ (precision$@K$), accuracy and inference speed.
% A detailed explanation of msIoU is provided in Section \ref{evaluation_metrics}.
% 3-7

% \vspace{-2mm}
\vspace{-4mm}
\subsection{GENNAV}
\begin{figure*}[t]
    \centering
    \includegraphics[width=0.9\linewidth]{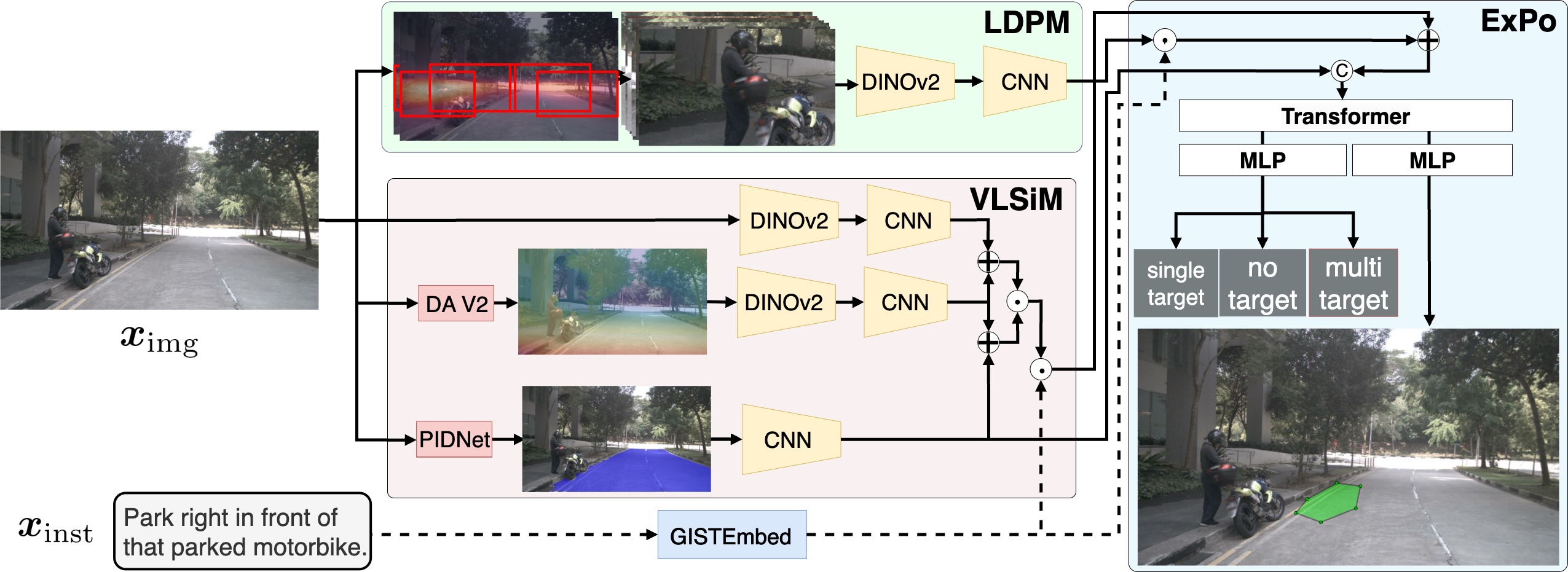}
    \caption{Overall architecture of GENNAV. DA represents Depth Anything~\citep{depth_anything_v2}. The green, red, and blue regions in this figure represent the LDPM, VLSiM, and ExPo modules, respectively.}
    \vspace{-5mm}
    \label{fig:model}
\end{figure*}

\vspace{-3mm}
% 4-2
% In GENNAV, we introduce a unified framework that integrates polygon-based segmentation with spatially grounded multimodal representations and landmark distribution driven patchification, enabling both appropriate semantic and spatial understanding of landmarks while maintaining computational efficiency.
% This framework models both the semantic and spatial characteristics with low computational overhead.
% The model is widely applicable to various RES and RNR models.
% The proposed method, GENNAV, is inspired by polygon-based referring expression segmentation (RES) approaches \citep{zhu2022seqtr, liu2023polyformer, cheng2024parallel, nishimura24iros}.
We propose GENNAV which integrates polygon-based segmentation with spatially grounded multimodal representations and landmark-distribution-driven patchification.
GENNAV models both the semantic and spatial characteristics with low computational overhead.
% {\color{blue}
It is inspired by polygon-based referring expression segmentation (RES) approaches \citep{zhu2022seqtr, liu2023polyformer, cheng2024parallel, nishimura24iros}, and is widely applicable to various RES and RNR models.
% }
Fig.~\ref{fig:model} shows the structure of GENNAV. It consists of three modules: Existence Aware Polygon Segmentation module (ExPo), Landmark Distribution Patchification Module (LDPM), and Visual-Linguistic Spatial Integration Module (VLSiM).

% 4-6
\vspace{-2mm}
% We define the input $\bm{x}$ to the model as follows:
%  $\bm{x} =\{ \bm{x}_\text{img}, \bm{x}_\text{inst} \}$, where $\bm{x}_\text{img} \in \mathbb{R}^{3 \times W \times H}$ and $\bm{x}_\text{inst} \in \{0,1 \}^{V \times L}$ denote a front camera image and a navigation instruction, respectively.
% Here, $W$, $H$, $V$ and $L$ denote the image width, image height, vocabulary size, and maximum token length, respectively.
We define the input $\bm{x}$ to the model as follows: $\bm{x}=\{\bm{x}_{\text{img}},\,\bm{x}_{\text{inst}}\}$, where
$\bm{x}_{\text{img}}\in\mathbb{R}^{3\times W\times H}$ and
$\bm{x}_{\text{inst}}\in\{0,1\}^{V\times L}$ denote a front-camera image and a navigation instruction, respectively, with the instruction represented as a sequence of one-hot vectors with the sequence length limited to a maximum of $L$ over a vocabulary of size $V$. Here, $W$ and $H$ denote the image width and height.
% 4-7
We preprocess $\bm{x}_{\text{inst}}$ with GISTEmbed \cite{solatorio2024gistembed} to obtain the linguistic feature $\bm{h}_{\text{inst}} \in \mathbb{R}^{C_{\text{inst}}}$, where $C_{\text{inst}}$ denotes the dimensionality of the linguistic feature.

% \clearpage
% \vspace{-3mm}
% \subsection{Landmark Distribution Patchification Module}
\vspace{-1mm}
% \textbf{LDPM.}
\textbf{Landmark Distribution Patchification Module.}
This module models fine-grained visual representations related to potential landmarks by efficiently partitioning patches based on the spatial distribution of landmarks.
Such representations are crucial because $\bm{x}_\text{inst}$ sometimes specify pedestrians/vehicles, however, most existing methods (e.g., ~\citep{rufus2021grounding, tnrsmral24}) resize input images to a low resolution, resulting in an insufficient number of pixels being allocated to distant landmarks.

\vspace{-2mm}
The input to this module is $\bm{x}_\text{img}$.
We adopt a patchification strategy for utilizing high-resolution images.
We introduce landmark-distribution-based patchification that covers regions where landmarks are likely to be densely distributed because classic uniform partitioning is inefficient.
The left image within the LDPM (green) in Fig.~\ref{fig:model} shows the patches (shown in red bounding boxes) and the distribution of landmarks in the training set of the Talk2Car dataset~\citep{deruyttere-etal-2019-talk2car}.
Each patch region is encoded by DINOv2~\citep{oquab2023dinov2}, followed by a CNN. The final feature $\bm{h_{\text{ldp}}}$ is obtained by integrating the resulting features across all patch regions.

\vspace{-1mm}
% \textbf{VLSiM.}
\textbf{Visual-Linguistic Spatial Integration Module.}
% \subsection{Visual-Linguistic Spatial Integration Module}
% \vspace{-3mm}
This module models the relationship between linguistic and spatial information by integrating $\bm{x}_\text{img}$, $\bm{h}_\text{inst}$, a pseudo-depth image, and road region. 
The inputs to VLSiM are $\bm{x}_\text{img}$ and $\bm{h}_{\text{inst}}$.
First, we obtain visual features from $\bm{x}_\text{img}$ using a frozen DINOv2~\citep{oquab2023dinov2}, followed by a CNN. This procedure is denoted by $f_{\text{vis}}(\bm{x}_\text{img})$.

\vspace{-2mm}
Next, we generate a pseudo-depth image $\bm{x}_\text{depth} \in \mathbb{R}^{3 \times W \times H}$ from $\bm{x}_\text{img}$ using a monocular depth estimation model (e.g., Depth Anything V2~\citep{oquab2023dinov2}) and we overlay the pseudo depth image with $\bm{x}_\text{img}$.
Similar to $f_{\text{vis}}(\bm{x}_\text{img})$, we input the overlaid image to DINOv2 and an additional CNN.
We introduce the CNN because DINOv2 is not specifically trained for overlaid images.
The above procedure is denoted by $f_{\text{depth}}(\bm{x}_\text{img})$.
Then, we integrate $\bm{h}_\text{inst}$ with a sum of $f_{\text{vis}}(\bm{x}_\text{img})$ and $f_{\text{depth}}(\bm{x}_\text{img})$ to model the relationship between linguistic and spatial features. 

\vspace{-2mm}
Moreover, to mitigate the generation of masks on inappropriate regions (e.g., the sky or areas occluded by objects), we integrate $f_{\text{depth}}(\bm{x}_\text{img})$ with a road representation.
We obtain a mask of the road $\bm{x}_\text{road} \in \mathbb{R}^{3 \times W \times H}$ using a semantic segmentation model (e.g., PIDNet~\citep{xu2023pidnet}) given $\bm{x}_\text{img}$.  
Subsequently, we input the mask to a CNN. The procedure is denoted by $f_{\text{road}}(\bm{x}_\text{img})$.
Finally, we obtain the following $\bm{h}_\text{mm}$ as the output of VLSiM:
\begin{align}
  \bm{h}_\text{mm} &= \bm{h}_\text{inst} \odot \left\{ f_{\text{vis}}\left( \bm{x}_\text{img} \right) + f_{\text{depth}}\left( \bm{x}_\text{img} \right) \right\} \odot \left\{ f_{\text{road}}\left( \bm{x}_\text{img} \right) + f_{\text{depth}}\left( \bm{x}_\text{img} \right) \right\},
\end{align}
where $\odot$ denotes the Hadamard product.

\vspace{-2mm}
% \textbf{ExPo.}
\textbf{Existence Aware Polygon Segmentation Module.}
% \vspace{-3mm}
This module predicts the existence of target regions and generates polygon-based segmentation masks by integrating four types of information:  $\bm{h}_{\text{inst}}$, $\bm{h}_{\text{mm}}$, $f_{\text{road}}\left(\bm{x}_\text{img}\right)$ and $\bm{h}_{\text{ldp}}$.
This module has two heads: a classification head that predicts the existence of target regions and a regression head that generates the segmentation masks of target regions.
The regression head leverages polygon-based segmentation that enables more efficient inference than existing pixel-based segmentation approaches~\citep{rufus2021grounding, tnrsmral24}, which is experimentally validated in Section~\ref{sec:ablation_study}.

\vspace{-2mm}
The module takes $\bm{h}_{\text{inst}}$, $\bm{h}_{\text{mm}}$, $f_{\text{road}}\left(\bm{x}_\text{img}\right)$ and $\bm{h}_{\text{ldp}}$ as inputs.
It integrates $\bm{h}_{\text{inst}}$ and $\bm{h}_{\text{ldp}}$ to obtain a multimodal representation, $\bm{h}_{\text{cmm}} = \left(\bm{h}_\text{ldp} \odot \bm{h}_\text{inst} \right) + \bm{h}_\text{mm}.$
% where another multimodal representation, $\bm{h}_{\text{mm}}$, is also added. 
Next, the regression head predicts the polygon vertices of target regions
$
\hat{\bm{c}}_i = \text{MLP}\left(\left[ \bm{h}_\text{cmm}; f_{\text{road}}\left(\bm{x}_\text{img}\right) \right]\right),
$
where $\text{MLP}(\cdot)$, $\hat{\bm{c}}_i \in \mathbb{R}^2 \ (i = 1, \dots, n_{\text{pt}})$, and $[\cdot; \cdot]$ denote an MLP, the coordinates of each vertex of the predicted polygon ordering in a clockwise manner starting from the top-left,  and concatenation, respectively.
Similarly, the classification head predicts $p(\hat{\bm{y}}_{\text{ex}})$ over the categories $\{\text{no-target}, \text{single-target}, \text{multi-target}\}$ based on the same input as the regression head.
Consequently, the output from the ExPo module is represented as $\hat{\bm{y}} = \{ p(\hat{\bm{y}}_{\text{ex}}), \hat{\bm{c}}_i \}$.

\vspace{-2mm}
We use the following loss function $\mathcal{L}$:
\vspace{-2mm}
\begin{align}
\mathcal{L} = \mathcal{L}_{\mathrm{CE}}\left(p(\hat{\bm{y}}_{\mathrm{ex}}), \bm{y}_{\mathrm{ex}}\right) + \lambda_{\mathrm{pt}} \sum_{i=1}^{n_{\mathrm{pt}}} \mathbb{I}\left[\bm{y}_{\mathrm{ex}} \in \{\text{single-target}, \text{multi-target}  \}\right] \cdot \ell_1 \left(\hat{\mathbf{c}}_i, \mathbf{c}_i\right),
\end{align}
where $\lambda_{\text{pt}}$, $\mathcal{L}_{\text{CE}}(\cdot, \cdot)$, $\ell_1$, and  $\mathbb{I}[\cdot]$ denote the loss weight, cross-entropy loss, L1 loss and indicator function, respectively.
\vspace{-3mm}
\section{
    Experiments
    \label{experimental Results}
}
\vspace{-3mm}
 \subsection{
    Experimental Setup 
    \label{experimental settings}
}
\vspace{-1mm}
\vspace{-1mm}
% \subsection{
%     Experimental Settings
%     \label{experimental-settings}
% }
\vspace{-1mm}
\textbf{Benchmark.}
% \subsection{
%     Benchmark
%     \label{Benchmark}
% }
% \vspace{-2mm}
To evaluate the performance of models on the three distinct types of cases in the GRNR task, we constructed the novel GRiN-Drive benchmark based on the Talk2Car-RegSeg~\citep{rufus2021grounding} and Refer-KITTI-V2~\citep{zhang2024bootstrapping} datasets.
To the best of our knowledge, aside from Talk2Car-RegSeg, most datasets do not provide front camera images that are annotated with instruction texts and segmentation masks of target regions.
In addition, the Talk2Car-RegSeg dataset alone is insufficient for the GRNR task because it lacks two critical components: (i) samples where no target region corresponds to the navigation instructions and (ii) samples that contain multiple landmarks in a single image.
To ensure that datasets cover no-target, single-target, and multi-target cases, we newly collected no-target and multi-target samples and constructed the GRiN-Drive benchmark.
Accordingly, the GRiN-Drive benchmark serves as a suitable benchmark for evaluating the performance of models in the three distinct types of cases in the GRNR task.

\vspace{-2mm}
The GRiN-Drive benchmark consists of 17,114 samples. 
Each sample consists of an image, navigation instruction and zero or more segmentation masks for each target region.
It has a vocabulary size, total number of words, and average sentence length of 2,699, 168,640, and 10.55 words, respectively.
% 5-6
The training, validation, and test sets contained 14,973, 1,413, and 758 samples, respectively.
% 5-7
We followed the original split~\citep{rufus2021grounding, wu2023referringRMOT}, as adopting a different dataset split would lead to an unfair comparison with baseline methods.
The GRiN-Drive test set achieves a statistical power greater than 0.999.
We used the training set for training models, validation set for adjusting their hyperparameters, and test set for evaluating the models.
The details of the construction of GRiN-Drive benchmark are explained in the supplementary materials.
 \vspace{-1mm}
 \textbf{Baselines.}
We used two groups of baseline methods: pixel-based methods (a method by~\citet{rufus2021grounding}, LAVT~\citep{yang2022lavt}, TNRSM~\citep{tnrsmral24},  and GSVA-Vicuna-7B~\citep{xia2024gsva}) and MLLMs (Gemini~\citep{reid2024gemini}, GPT-4o~\citep{gpt4o}, and Qwen2-VL~\citep{Qwen2-VL}).
 % 6-3
Each method was used as a baseline method for the following reasons.
We selected LAVT~\citep{yang2022lavt} and GSVA-Vicuna-7B~\citep{xia2024gsva} as a baseline method because of its successful application to pixel-wise referring image segmentation tasks.
Similarly, we used the method by Rufus et al. and TNRSM~\citep{tnrsmral24} because of their effective application in the RNR task, which is closely related to the Generalized Referring Navigable Regions (GRNR) task.
The details of the experimental settings of baseline methods are explained in the supplementary materials.
% The details of the reason for selection and the experimental settings of the baseline methods are described in the supplementary materials.

\vspace{-1mm}
% \subsection{
%     Evaluation Metrics
%     \label{evaluation_metrics}
% }
% \vspace{-3mm}
\textbf{Evaluation Metrics.} 
We used mean stuff Intersection over Union (msIoU), precision@$K$ ($\mathrm{P}@K$), accuracy (Acc.), and inference speed as evaluation metrics, with msIoU as the primary metric.
We propose the new metric msIoU to evaluate single-target, multi-target, and no-target samples without biases.
By contrast, most existing metrics yield biased evaluations: a correct no-target prediction yields an IoU of 1.0, whereas even accurate target segmentations yield scores lower than 1.0.
Therefore, the metric favors trivial solutions focusing on target-existence classification, which is weighted more than mask generation.
Indeed, in our task, even a trivial solution that always predicts ``no target'' can achieve a gIoU\citep{liu2023gres} score of 0.33, which is deceptively high, surpassing the human performance of msIoU 0.17 on target samples in the test set of the GRiN-Drive benchmark.

\vspace{-2mm}
To address this, msIoU assigns a score of 1 to samples with IoU exceeding a threshold $K$, while normalizing IoU scores based on $K$ for those below the threshold, as shown in Equation \ref{eq:msIoU}.
By further averaging $\mathrm{sIoU}@k$ where ($k=$1, \ldots, $1/K$), msIoU achieves a balanced evaluation for both target and no-target samples. The details of the remaining evaluation metrics are explained in the supplementary materials.
% Additional details on $\mathrm{sIoU}@k$ (k=1,2,3,4,5) are provided in the supplementary material (Sec XX).
msIoU is defined as follows:

\vspace{-5mm}
\begin{align}
\label{eq:msIoU}
% \text{msIoU} &= \frac{1}{M} \sum_{k=1}^{M=5} \text{sIoU@}k, \\
% \text{sIoU@}k &= \frac{1}{N} \sum_{i=1}^{N} \text{sIoU@}k_{i}, \\
\text{msIoU} &= \text{mean} \left( \frac{1}{N} \sum_{i=1}^{N} \text{sIoU@}k_{i} \right), \\
\text{sIoU@}k_i &=
\begin{cases}
\min\left(k \cdot \text{IoU}(\hat{y}_i, y_i), 1\right) & \text{if TP,} \\
\makebox[0pt][l]{\hspace{4.5em}1} & \text{if TN,} \\
\makebox[0pt][l]{\hspace{4.5em}0} & \text{if FP or FN,}
\end{cases}
\end{align} where $\hat{y}_i$ and $y_i$ denote the predicted and ground-truth masks for the $i$-th sample, respectively; $N$ denotes the number of samples, and $\text{IoU}(\cdot)$ represents the IoU between them. Additionally, TP, TN, FP and FN denote the numbers of true positive, true negative, false positive, and false negative samples, respectively, which represents the prediction regarding the existence of target regions. 
The details of the remaining evaluation metrics are explained in the supplementary materials.

% 6-6
\vspace{-3mm}
\subsection{Quantitative Results
\label{sec:quantitative}
}
\begin{table*}[t]
\centering
\renewcommand{\arraystretch}{1.25}
\caption{Quantitative comparison between GENNAV and the baseline methods on the test sets of the GRiN-Drive benchmark. The best score for each metric is in bold. The ``Type'' and ``Inf. speed'' columns specify the segmentation approach and inference speed of each method, respectively.}
\vspace{-1mm}
\setlength{\tabcolsep}{2pt}
\resizebox{0.85\textwidth}{!}{%
\begin{tabular}{lccccccc}
\Xhline{1pt}
\multirow{2}{*}{Method} & \multirow{2}{*}{Resolution} & \multirow{2}{*}{Type} & \multirow{2}{*}{msIoU [\%]$\uparrow$} & \multirow{2}{*}{$\mathrm{P}@0.1$ [\%]$\uparrow$} &  \multirow{2}{*}{$\mathrm{P}@0.2$ [\%]$\uparrow$} & \multirow{2}{*}{Acc. [\%]$\uparrow$} & Inf. speed \\
 & & & & & & & [ms/sample]$\downarrow$ \\
\hline
\citet{rufus2021grounding} & 224×224 & Pixel   & 35.44 {$\scriptscriptstyle \pm 3.12$} & 21.53 {$\scriptscriptstyle \pm 4.42$} & 17.52 {$\scriptscriptstyle \pm 2.41$} & 51.61 {$\scriptscriptstyle \pm 9.86$} & 39.87 {$\scriptscriptstyle \pm 1.10$} \\
LAVT~\citep{yang2022lavt} & 224×224 & Pixel   & 31.73 {$\scriptscriptstyle \pm 2.74$} & 30.28 {$\scriptscriptstyle \pm 3.16$} & 23.53 {$\scriptscriptstyle \pm 2.19$} & 40.95 {$\scriptscriptstyle \pm 1.56$} & \textbf{9.29 {$\scriptscriptstyle \pm 0.24$}} \\
TNRSM~\citep{tnrsmral24} & 224×224 & Pixel   & 37.90 {$\scriptscriptstyle \pm 0.82$} & 44.05 {$\scriptscriptstyle \pm 1.33$} & 34.09 {$\scriptscriptstyle \pm 0.58$} & 67.23 {$\scriptscriptstyle \pm 2.72$} & 502.69 {$\scriptscriptstyle \pm 0.22$} \\
\hline
\citet{rufus2021grounding} & 640×640 & Pixel   & 37.84 {$\scriptscriptstyle \pm 1.95$} & 11.48 {$\scriptscriptstyle \pm 4.42$} & 7.68 {$\scriptscriptstyle \pm 3.64$}  & 52.19 {$\scriptscriptstyle \pm 4.34$} & 65.00 {$\scriptscriptstyle \pm 2.11$} \\
LAVT~\citep{yang2022lavt}  & 640×640 & Pixel   & 34.09 {$\scriptscriptstyle \pm 1.53$} & 7.90 {$\scriptscriptstyle \pm 17.66$} & 6.79 {$\scriptscriptstyle \pm 15.17$} & 38.66 {$\scriptscriptstyle \pm 10.69$} & 11.51 {$\scriptscriptstyle \pm 0.33$} \\
TNRSM~\citep{tnrsmral24} & 640×640 & Pixel   & 22.84 {$\scriptscriptstyle \pm 8.77$} & 38.57 {$\scriptscriptstyle \pm 4.25$} & 19.45 {$\scriptscriptstyle \pm 7.53$} & 67.89 {$\scriptscriptstyle \pm 1.11$} & 503.68 {$\scriptscriptstyle \pm 0.20$} \\
\hline
GSVA-Vicuna-7B~\citep{xia2024gsva} & 1024×1024 & Pixel   & 32.94 {$\scriptscriptstyle \pm 0.04$} & 17.26 {$\scriptscriptstyle \pm 0.24$} & 9.36 {$\scriptscriptstyle \pm 0.14$} & 64.76 {$\scriptscriptstyle \pm 0.11$} & 306.73 {$\scriptscriptstyle \pm 7.11$} \\
\hline
Gemini~\citep{reid2024gemini} & 1600×900 & Bbox    & 6.98 {$\scriptscriptstyle \pm 0.58$} & 6.92 {$\scriptscriptstyle \pm 1.46$}  & 1.63 {$\scriptscriptstyle \pm 0.51$} & 46.33 {$\scriptscriptstyle \pm 0.05$} & 1793.68 {$\scriptscriptstyle \pm 37.96$} \\
GPT-4o~\citep{gpt4o} & 1600×900 & Bbox    & 23.41 {$\scriptscriptstyle \pm 5.72$} & 5.04 {$\scriptscriptstyle \pm 1.37$}  & 0.04 {$\scriptscriptstyle \pm 0.09$} & 62.44 {$\scriptscriptstyle \pm 9.35$} & 3525.68 {$\scriptscriptstyle \pm 701.04$} \\
Qwen2-VL~\citep{Qwen2-VL} & 1600×900 & Bbox    & 24.06 {$\scriptscriptstyle \pm 1.15$} & 3.85 {$\scriptscriptstyle \pm 0.95$}  & 1.64 {$\scriptscriptstyle \pm 0.52$} & 64.25 {$\scriptscriptstyle \pm 0.91$} & 1768.99 {$\scriptscriptstyle \pm 0.41$} \\
Qwen2-VL~\citep{Qwen2-VL} & 1600×900 & Polygon & 12.16 {$\scriptscriptstyle \pm 4.86$} & 0.08 {$\scriptscriptstyle \pm 0.18$}  & 1.88 {$\scriptscriptstyle \pm 0.74$} & 44.66 {$\scriptscriptstyle \pm 5.97$} & 1771.41 {$\scriptscriptstyle \pm 6.20$} \\
\hline
GENNAV (ours) & 640×640 & Polygon & \textbf{46.35 {$\scriptscriptstyle \pm 1.52$}} & \textbf{49.60 {$\scriptscriptstyle \pm 1.12$}} & \textbf{34.40 {$\scriptscriptstyle \pm 1.20$}} & \textbf{75.41 {$\scriptscriptstyle \pm 0.67$}} & 31.31 {$\scriptscriptstyle \pm 0.05$} \\
\hline
\hline
Human & 1600×900 & Polygon & 56.08 & 56.00 & 36.40 & 88.00 & - \\

\Xhline{1pt}
\end{tabular}
}
\vspace{-6mm}
\label{tab:quantitative}
\end{table*}
\vspace{-2mm}

Table \ref{tab:quantitative} shows the quantitative results of the baseline methods, GENNAV and human performance on the GRiN-Drive benchmark. 
The values in the table are the average and standard deviation over five trials.
The ``Type'' and ``Inf. speed'' columns specify the segmentation approach and inference speed of each method, respectively.
The detailed discussions of quantitative results and the subject experiment are provided in the supplementary material.

\vspace{-2mm}
Table \ref{tab:quantitative} indicates that GENNAV achieved highest msIoU of 46.35.
% The msIoU of GENNAV and the baseline methods are as follows, grouped by input resolution.
% In the 224×224 setting, a method by Rufus et al., LAVT, and TNRSM achieved msIoU scores of 35.44, 31.73, and 37.90, respectively.
% Under the 640×640 condition, GENNAV reached an msIoU of 46.35, whereas a method by Rufus et al., LAVT, and TNRSM resulted in scores of 37.84, 34.09, and 22.84, respectively.
% In the high-resolution (1600×900) setting, the MLLM baselines Gemini, GPT-4o, Qwen2-VL (bbox), and Qwen2-VL (polygon) yielded msIoU scores of 6.98, 23.41, 24.06, and 12.16, respectively.
These results show that GENNAV improved by 8.45 points over TNRSM (224×224), which achieved the highest msIoU among the baseline methods.
% Similarly, the $\mathrm{P}@0.1$ scores of the baseline methods by Rufus et al., LAVT, and TNRSM in the 224×224 setting were 21.53, 30.28, and 44.05, respectively.
%  At the 640×640 resolution, the $\mathrm{P}@0.1$ scores of GENNAV and the baseline methods by Rufus et al., LAVT, and TNRSM were 11.48, 7.90, and 38.57, respectively.
%  Furthermore, MLLM baselines — Gemini, GPT-4o, Qwen2-VL (bbox), and Qwen2-VL (polygon) — achieved $\mathrm{P}@0.1$ scores of 6.92, 5.04, 3.85, and 0.08, respectively.
% \vspace{-2mm}
% Similarly, GENNAV outperformed the baseline methods in terms of $\mathrm{P}@0.1$ and accuracy, demonstrating a 5.55, 7.61 point improvement over the best-performing baseline, TNRSM (224×224).
Similarly, GENNAV achieved $\mathrm{P}@0.1$ and accuracy of 49.60\% and 75.41\%, respectively, surpassing the baseline methods, including the MLLMs, with a 5.55 and 7.52 point improvement over the best-performing baseline, TNRSM (224×224).
The inference speed per sample for GENNAV was 31.31 ms.  
% For comparison, the inference speeds of the baseline methods were as follows. In the 224×224 setting, a method by Rufus et al., LAVT and TNRSM took 39.87 ms, 9.29 ms and 502.69 ms, respectively.
% Under the 640×640 resolution setting, a method by Rufus et al., LAVT, and TNRSM took 65.00 ms, 11.51 ms, and 503.68 ms, respectively.
% In contrast, MLLM-based methods exhibited significantly slower inference speeds: Gemini, GPT-4o, Qwen2-VL (bbox), and Qwen2-VL (polygon) took 1793.68 ms, 3525.68 ms, 1768.99 ms, and 1771.41 ms, respectively.  
Overall, GENNAV achieved faster inference speed than the baseline methods except for LAVT.
Although GENNAV was slower than LAVT, it achieved substantially higher performance on other metrics.
The improvement achieved by GENNAV in terms of msIoU, $\mathrm{P}@0.1$, and accuracy was statistically significant ($p < 0.05$).
% We provide the details of the subject experiment in the supplementary materials.

\vspace{-3mm}
\subsection{Qualitative Results}
\vspace{-2mm}

% 6-8
\begin{figure}[t]
    \centering
    \begin{minipage}{1.0\linewidth}
            \vspace{0pt}% to make [t] work properly
            % Row 1
            \begin{minipage}{0.03\linewidth}
                (i)
            \end{minipage}
            \begin{minipage}{0.2365\linewidth}
                \includegraphics[width=\linewidth]{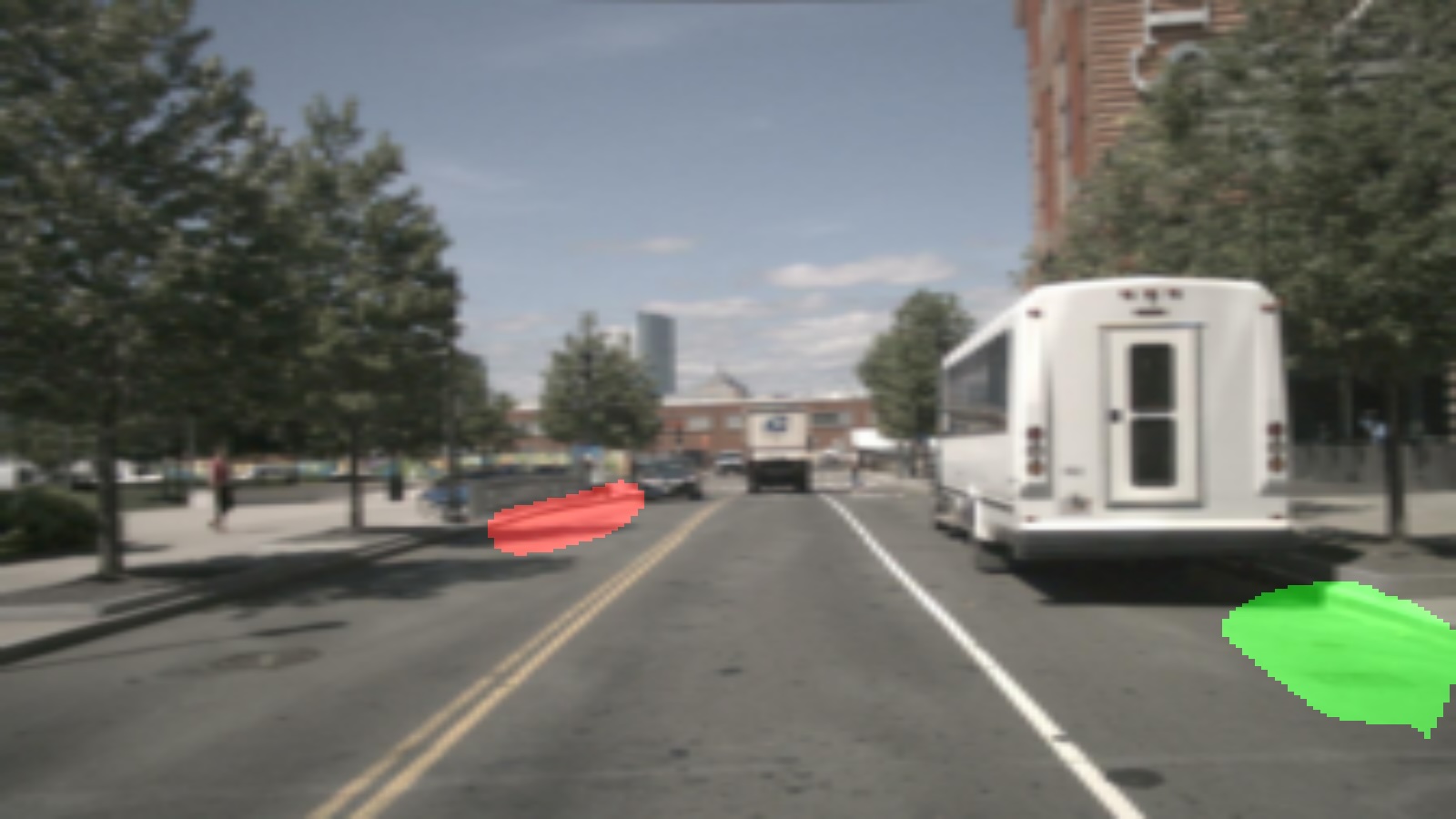}
            \end{minipage}
            \hfill
            \begin{minipage}{0.2365\linewidth}
                \includegraphics[width=\linewidth]{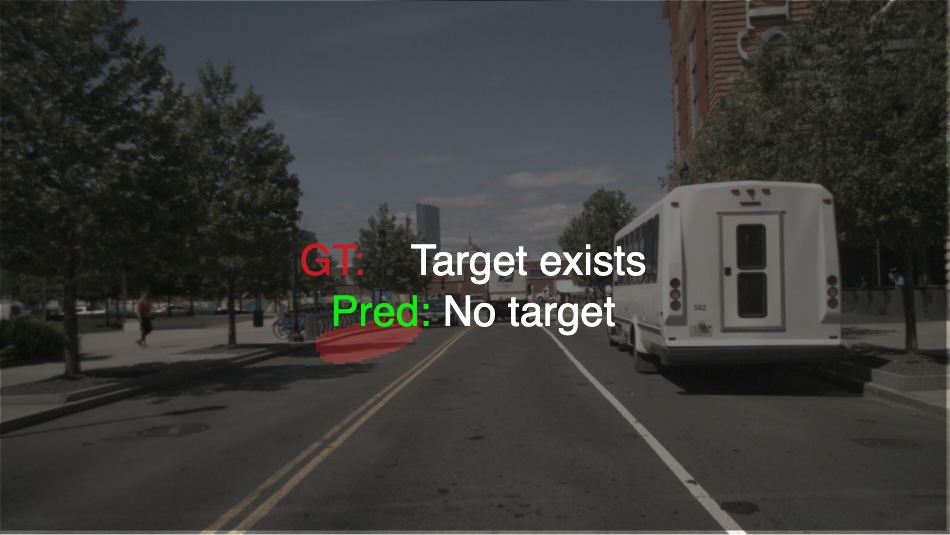}
            \end{minipage}
            \begin{minipage}{0.2365\linewidth}
                \includegraphics[width=\linewidth]{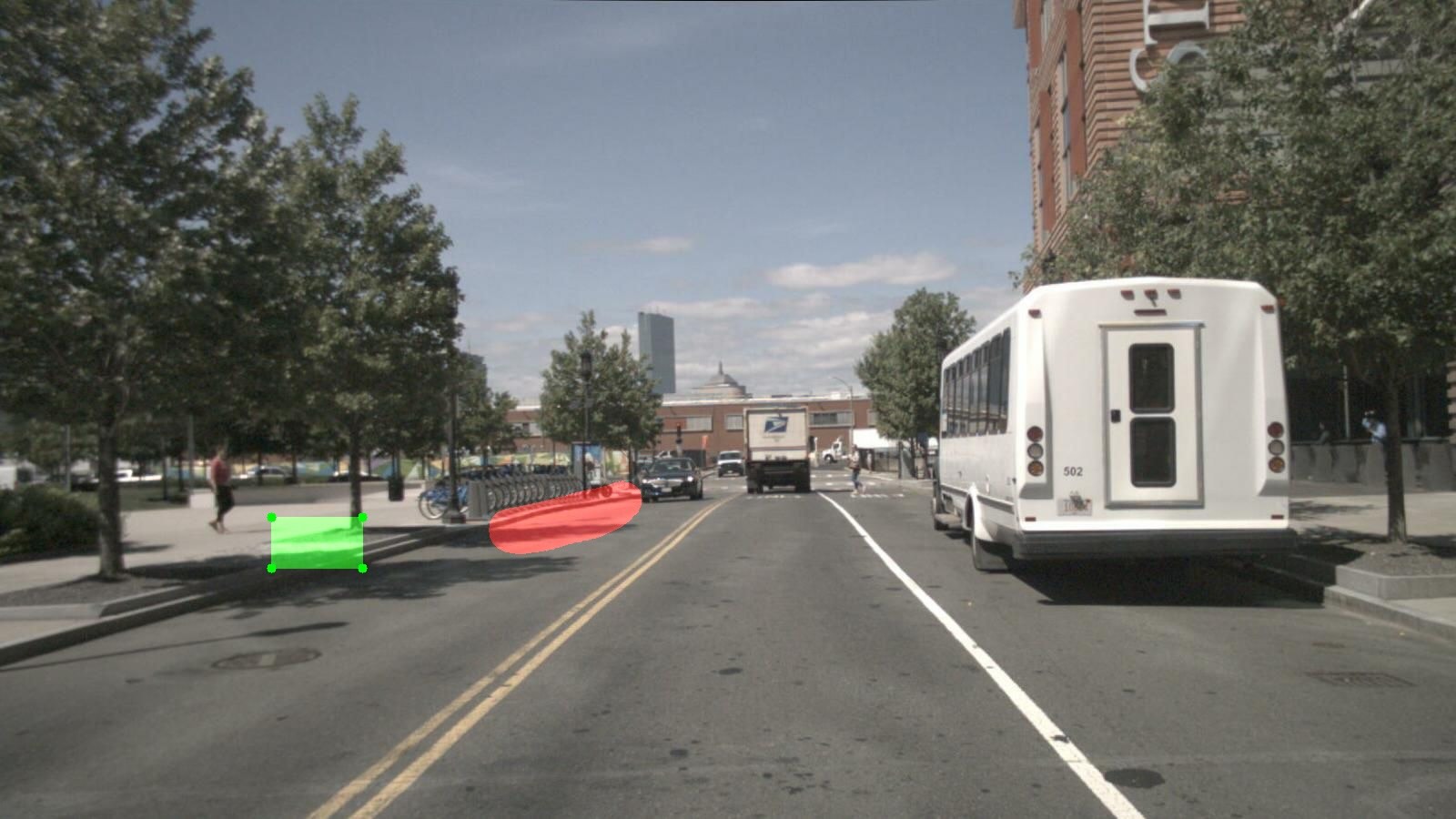}
            \end{minipage}
            \hfill
            \begin{minipage}{0.2365\linewidth}
                \includegraphics[width=\linewidth]{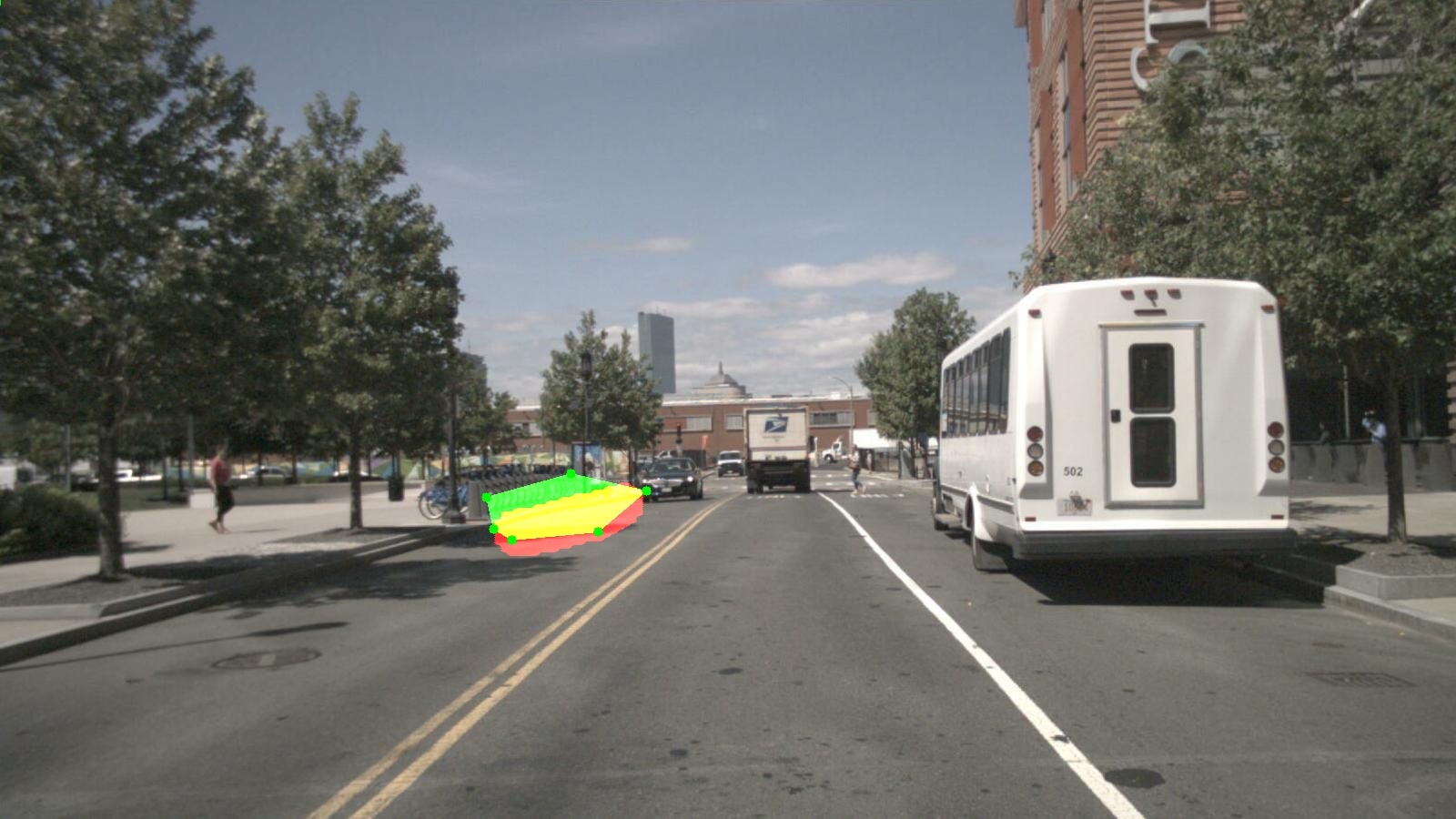}
            \end{minipage}    
            \begin{minipage}[t]{1.0\linewidth}
                % \vspace{3pt}
                \centering {$\bm{x}_\text{inst}$: ``Park my car next to the rack.''}
            \end{minipage}
            % Row 2
        \end{minipage}
    \begin{minipage}{1.0\linewidth}
            \vspace{0pt}% to make [t] work properly
            % Row 1
            \begin{minipage}{0.03\linewidth}
                (ii)
            \end{minipage}
            \begin{minipage}{0.2365\linewidth}
                \includegraphics[width=\linewidth]{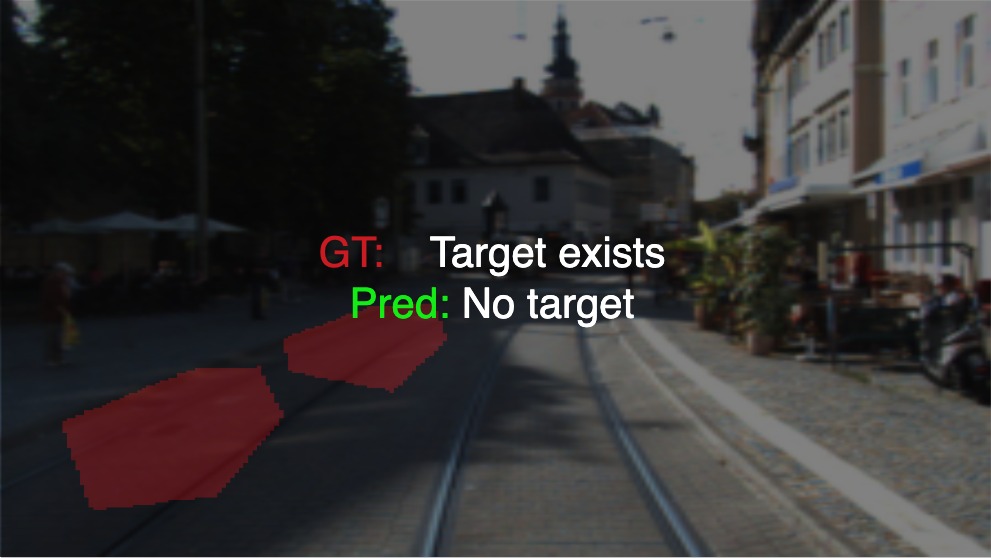}
            \end{minipage}
            \hfill
            \begin{minipage}{0.2365\linewidth}
                \includegraphics[width=\linewidth]{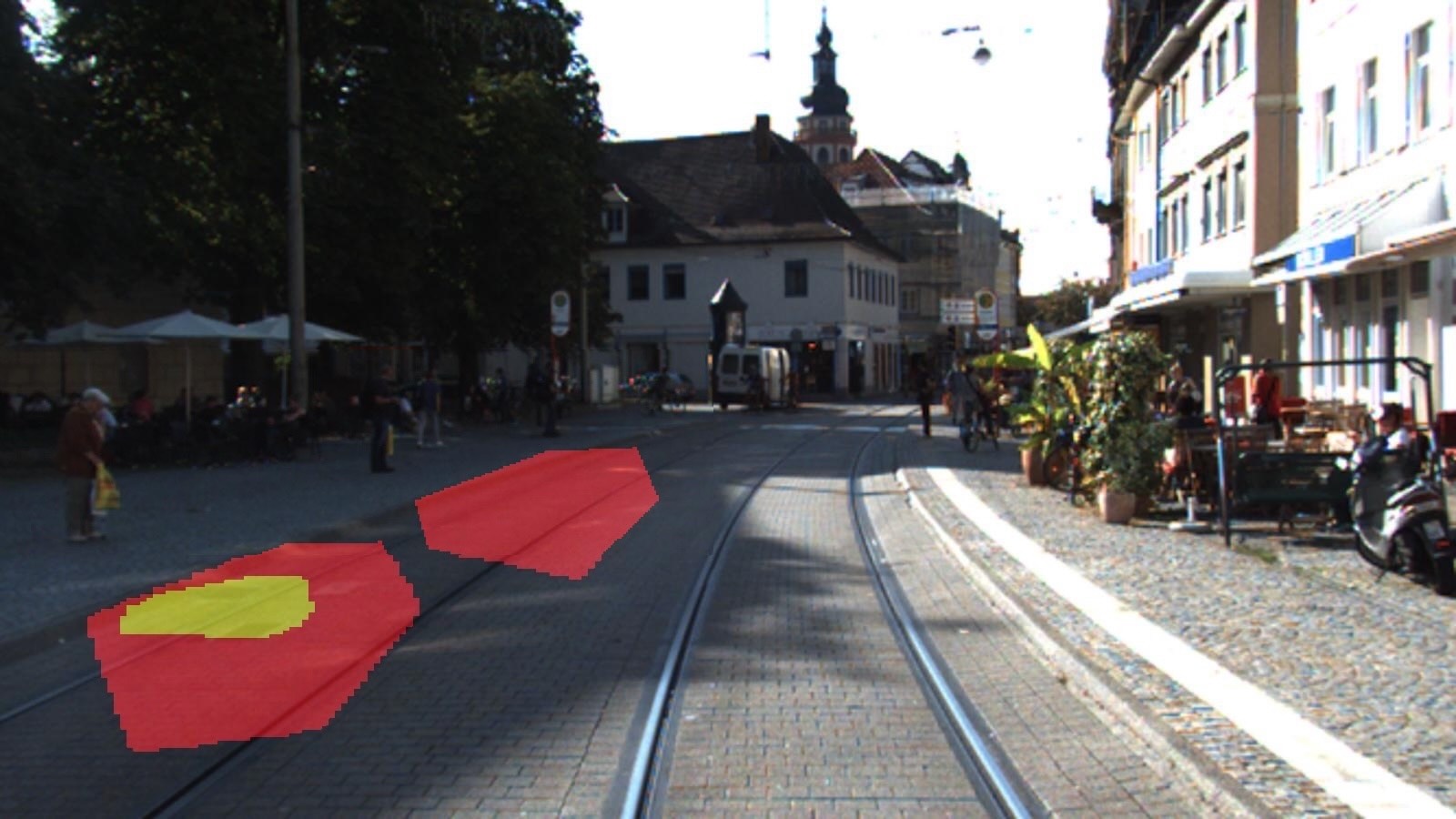}
            \end{minipage}
            \begin{minipage}{0.2365\linewidth}
                \includegraphics[width=\linewidth]{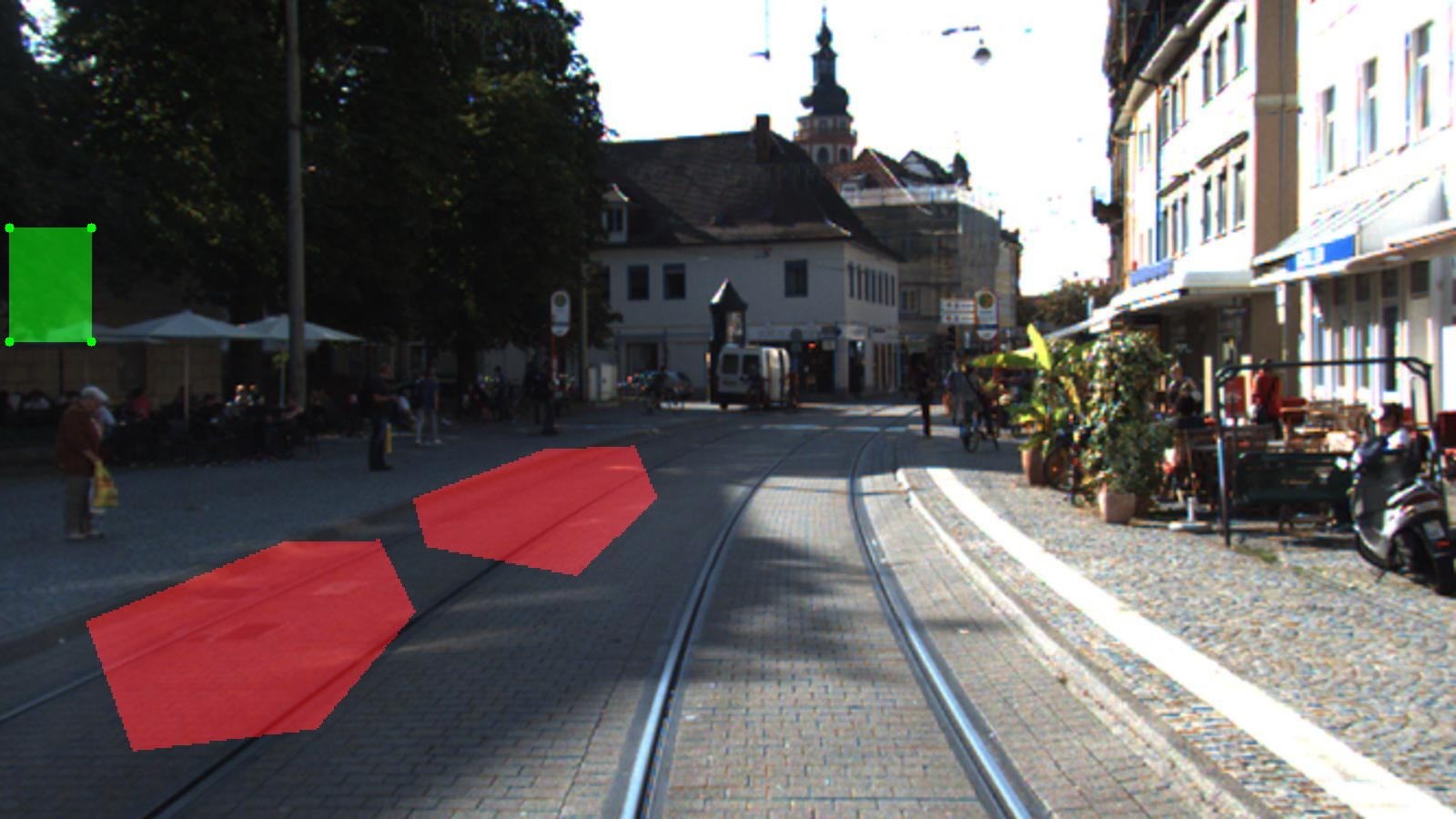}
            \end{minipage}
            \hfill
            \begin{minipage}{0.2365\linewidth}
                \includegraphics[width=\linewidth]{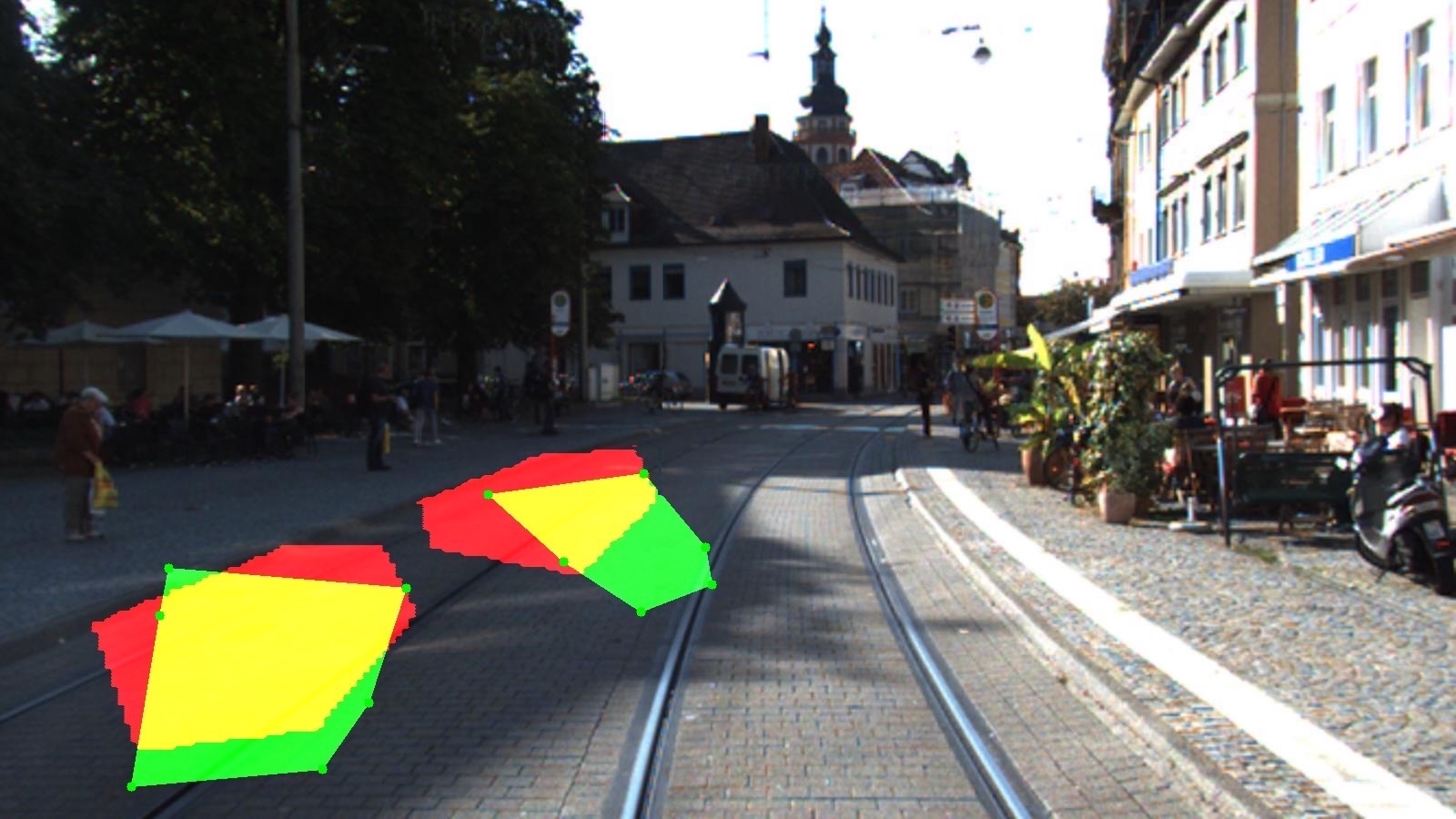}
            \end{minipage}    
            \begin{minipage}[t]{1.0\linewidth}
                \centering { $\bm{x}_\text{inst}$: ``Pull up right next to pedestrian on the left.''}
            \end{minipage}
            \begin{minipage}{0.03\linewidth}
                \quad
            \end{minipage}
            \begin{minipage}{0.2365\linewidth}
                \centering (a)
            \end{minipage}
            \begin{minipage}{0.2365\linewidth}
                \centering (b)
            \end{minipage}
            \begin{minipage}{0.2365\linewidth}
                \centering (c)
            \end{minipage}
            \begin{minipage}{0.2365\linewidth}
                \centering (d)
            \end{minipage} 
            \vspace{-3mm}
        \end{minipage}
    \caption{
     Qualitative results of GENNAV and baseline methods on the GRiN-Drive benchmark. Columns (a), (b), (c) and (d) show the prediction by LAVT, TNRSM, Qwen2-VL (bbox) and GENNAV, respectively.
     The green and red regions indicate the predicted and ground-truth regions, respectively; yellow indicates their overlap.
     % The green and red regions indicate the predicted and ground-truth regions, respectively, whereas the yellow region represents the overlap between the predicted and ground-truth regions.
    }
    \label{fig:qualitative-success}
    \vspace{-6mm}
\end{figure}

Fig.~\ref{fig:qualitative-success} shows the qualitative results of GENNAV and the baseline methods on the GRiN-Drive benchmark.
Figs.~\ref{fig:qualitative-success} (i) shows single-target cases and Figs.~\ref{fig:qualitative-success} (ii) shows multi-target cases. 
Further qualitative results, including no-target cases, are provided in the supplementary materials.
Figs.~\ref{fig:qualitative-success} (i) presents predictions given ``Park my car next to the rack'' as $\bm{x}_\text{inst}$. 
Masks should be generated for the region in front of the rack, because there is a bike rack located center-left in the image.
GENNAV correctly generated the region whereas TNRSM predicted a ``no-target'' and LAVT and Qwen2-VL (bbox) generated inappropriate masks. 
% Fig. \ref{fig:qualitative-success} (ii) illustrates an example, where the $\bm{x}_\text{inst}$ is, ``Pull over where that man is.’'
% In this example, the model is expected to generate masks on the road in front of the man on the right side of the image.
% The baseline methods LAVT and TNRSM generated the inappropriate masks on regions around the truck and Qwen2-VL (bbox) inappropriately predicted the central region of the road.
% By contrast, our method was able to generate an appropriate mask on the region next to the man. 
% Fig.~\ref{fig:qualitative-success} (ii) demonstrates a no-target case with $\bm{x}_\text{inst}$: ``Slow down and let that gold car pass in front of us.''
% Given that the gold car does not exist in the scene, the appropriate prediction should be a ``no-target.''
% GENNAV successfully predicted this sample as a ``no-target.'' 
% The baseline methods were unable to appropriately predict the absence of landmarks, which resulted in an inappropriate generation of masks on the region on the road.
In Figs.~\ref{fig:qualitative-success} (ii), $\bm{x}_\text{inst}$ was ``Pull up right next to pedestrian on the left.''
Therefore, the model is required to generate a mask for a region located next to each pedestrian positioned on the left side of the road.
GENNAV successfully generated the mask on the appropriate region for each of the two pedestrians.
However, none of the baseline methods were able to generate appropriate masks; for instance, TNRSM incorrectly predicted only a single region.

\vspace{-4mm}
\subsection{Ablation Study
\label{sec:ablation_study}
}
\vspace{-3mm}

\begin{table}[t]
    \centering
    \caption{
    Quantitative results of the ablation studies. The best values for each metric are highlighted in bold. The ``Type'' column specifies the segmentation approach of each method.
    }
    \resizebox{0.9\textwidth}{!}{%
    \begin{tabular}{lcccccccc}
        \toprule
        \multirow{2}{*}{Model} & \multirow{2}{*}{Type} & \multirow{2}{*}{LDPM} & \multicolumn{2}{c}{VLSiM} & \multirow{2}{*}{msIoU [\%]$\uparrow$} 
        & \multirow{2}{*}{$\mathrm{P}@0.1$ [\%]$\uparrow$} & \multirow{2}{*}{$\mathrm{P}@0.2$ [\%]$\uparrow$} & \multirow{2}{*}{Acc. [\%]$\uparrow$} \\
        & & & $f_{\text{depth}}$ & $f_{\text{road}}$ & & & & \\
        \midrule
        (i)   & Polygon &  & $\checkmark$ & $\checkmark$ & 43.05\,$\scriptscriptstyle\pm\,0.72$ & 40.75\,$\scriptscriptstyle\pm\,1.77$ & 24.48\,$\scriptscriptstyle\pm\,1.59$ & 74.62\,$\scriptscriptstyle\pm\,0.86$ \\
        (ii)  & Polygon & $\checkmark$ &  & $\checkmark$ & 44.13\,$\scriptscriptstyle\pm\,0.26$ & 48.14\,$\scriptscriptstyle\pm\,0.44$ & 31.63\,$\scriptscriptstyle\pm\,0.57$ & 74.01\,$\scriptscriptstyle\pm\,0.79$ \\
        (iii) & Polygon & $\checkmark$ & $\checkmark$ &  & 44.53\,$\scriptscriptstyle\pm\,0.78$ & 46.59\,$\scriptscriptstyle\pm\,2.56$ & 30.95\,$\scriptscriptstyle\pm\,1.41$ & 73.75\,$\scriptscriptstyle\pm\,0.49$ \\
        (iv)  & Pixel   & $\checkmark$ & $\checkmark$ & $\checkmark$ & 35.26\,$\scriptscriptstyle\pm\,1.21$ & 27.18\,$\scriptscriptstyle\pm\,2.10$ & 18.02\,$\scriptscriptstyle\pm\,1.67$ & 62.93\,$\scriptscriptstyle\pm\,2.03$ \\
        (v)   & Polygon & $\checkmark$ & $\checkmark$ & $\checkmark$ & \textbf{46.35}\,$\scriptscriptstyle\pm\,1.52$ & \textbf{49.60}\,$\scriptscriptstyle\pm\,1.12$ & \textbf{34.40}\,$\scriptscriptstyle\pm\,0.80$ & \textbf{75.41}\,$\scriptscriptstyle\pm\,0.67$ \\
        \bottomrule
    \end{tabular}
    }
    \label{tab:ablation}
    \vspace{-5mm}
\end{table}

We conducted three ablation studies to investigate the contribution of each module in GENNAV.
Table \ref{tab:ablation} shows the results of the ablation studies.
The table presents the average and standard deviation of the results of five trials. The highest score for each metric is in bold.

\vspace{-2mm}
\noindent\textbf{LDPM Ablation.}
We omitted LDPM to analyze its contributions. 
Table \ref{tab:ablation} shows that Model (i) achieved msIoU of 43.05, which was 3.3 points lower than that of Model (v), respectively. 
This indicates that LDPM enhanced the understanding of landmarks by partitioning patches based on landmark distributions.

\vspace{-2mm}
\noindent\textbf{VLSiM Ablation.}
To investigate the impact of the subnetworks $f_\text{depth}\left(\cdot\right)$ and $f_\text{road}\left(\cdot\right)$ on performance, we conducted ablation by removing each of them individually.
First, we compare Model (v) (full) and Model (ii) (without $f_\text{depth}\left(\cdot\right)$). 
Model (ii) showed decreases of 2.22 points in msIoU.
These results imply that $f_\text{depth}\left(\cdot\right)$ assisted the alignment between linguistic and spatial information by using generated pseudo-depth images.
Similarly, we removed $f_\text{road}\left(\cdot\right)$ to assess its contribution.
As shown in Table \ref{tab:ablation}, msIoU of Model (iii) was 1.82 points lower, respectively, than that of Model (v).
This indicates that $f_\text{road}\left(\cdot\right)$ facilitated the understanding of immovable regions and improved the model's mask generation capability.

\vspace{-2mm}
\noindent\textbf{Decoder Ablation.}
To analyze the impact of the decoder structure, we compare our polygon-based approach with a pixel-based method.
Model (iv) underperformed Model (v) by 11.09 points in terms of msIoU.
Notably, Model (iv), which employs pixel-based prediction, exhibits a decrease in inference speed to 47.92 ms/sample compared to polygon-based Model (v) 31.31 ms/sample.
This demonstrates that our polygon-based approach successfully reduced the computational cost.

\vspace{-4mm}
\section{Real-World Experiments}
\vspace{-2mm}

\vspace{-1mm}
% \textbf{Experimental Settings.} 
\subsection{Experimental Setup}
\vspace{-1mm}
We validated GENNAV in zero-shot, real-world experiments.
In real-world experiments, we used four automobiles operated across five geographically distinct urban areas, which resulted in 20 video recordings.
Each vehicle was equipped with smartphone cameras (both iPhone and Android devices), which allowed us to evaluate GENNAV under varying hardware setups.
Video data were collected under a wide range of traffic and environmental conditions (e.g.,  low-light settings, clear weather, and snow-covered roads), which covered diverse real-world scenarios.
The model executed the GRNR task based on navigation instructions in environments that had not been observed during training.
The experiments consisted of 120 episodes across single-target, no-target and multi-target scenarios, with 40 episodes per scenario.
The experiment involved single-target, no-target and multi-target scenarios, each of which consisted of 40 scenarios.
% We used GPT-4o to generate navigation instructions. 
The details for the assignment of instructions and data selection are shown in the supplementary materials.
% {\color{blue}Due to the usage of GPT-4o for instruction generation in our experiment, we considered its use as a baseline method to constitute data leakage.
% Therefore, we excluded GPT-4o from the quantitative comparisons.}

\vspace{-3mm} 
% \textbf{Quantitative Results.} 
\subsection{Quantitative Results}
\vspace{-3mm}
Table~\ref{tab:quantitative-physical} shows the quantitative comparison between GENNAV and baseline methods in the real-world experiment.
The values in the table are the average and standard deviation over five trials.
We show the best baseline method for each resolution. The remaining results and detailed discussions are provided in the supplementary material.
GENNAV reached an msIoU of 34.32, whereas a method by TNRSM (224×224), LAVT(640×640), and Qwen2-VL (bbox) resulted in scores of 23.11, 30.44, and 20.48, respectively.
% \begin{table*}[t]
% \centering
% \setlength{\tabcolsep}{2pt}
% \renewcommand{\arraystretch}{1.25}
% \caption{Quantitative comparison between the proposed method and baseline methods in the real-world experiment. The best score for each metric is in bold.}
% \begin{tabular}{lcccccc}
% \hline
% Method & msIoU [\%]$\uparrow$ & $\mathrm{P}@0.1$ [\%]$\uparrow$ & Acc. [\%]$\uparrow$ \\
% \hline
% TNRSM \citep{tnrsmral24} (224×224) & 23.11 {$\scriptscriptstyle \pm 5.88$} & 24.25 {$\scriptscriptstyle \pm 4.11$} & 56.83 {$\scriptscriptstyle \pm 5.08$} \\
% LAVT \citep{yang2022lavt} (640×640) & 30.44 {$\scriptscriptstyle \pm 1.63$} & 11.73 {$\scriptscriptstyle \pm 10.69$} & 32.67 {$\scriptscriptstyle \pm 19.74$} \\
% Qwen2-VL~\citep{Qwen2-VL} (bbox)    & 20.48 {$\scriptscriptstyle \pm 2.33$} & 5.75 {$\scriptscriptstyle \pm 2.59$} & 66.33 {$\scriptscriptstyle \pm 3.36$} \\
% \hline
% \textbf{GENNAV (ours)} & \textbf{34.32 {$\scriptscriptstyle \pm 2.97$}} & \textbf{28.75 {$\scriptscriptstyle \pm 5.08$}} & \textbf{67.50 {$\scriptscriptstyle \pm 2.95$}} \\
% \hline
% \end{tabular}
% \label{tab:quantitative-physical}
% \end{table*}

\begin{wraptable}{r}{0.6\textwidth}
\centering
\vspace{-8mm}
\setlength{\tabcolsep}{2pt}
\renewcommand{\arraystretch}{1.25}
\caption{Quantitative comparison between the proposed method and baseline methods in the real-world experiment. The best score for each metric is in bold.}
\vspace{-1mm}
\resizebox{0.55\textwidth}{!}{% ここを変更
\begin{tabular}{lccc}
\Xhline{1pt}
Method & msIoU [\%]$\uparrow$ & $\mathrm{P}@0.1$ [\%]$\uparrow$ & Acc. [\%]$\uparrow$ \\
\hline
TNRSM \citep{tnrsmral24} (224×224) & 23.11 {$\scriptscriptstyle \pm 5.88$} & 24.25 {$\scriptscriptstyle \pm 4.11$} & 56.83 {$\scriptscriptstyle \pm 5.08$} \\
LAVT \citep{yang2022lavt} (640×640) & 30.44 {$\scriptscriptstyle \pm 1.63$} & 11.73 {$\scriptscriptstyle \pm 10.69$} & 32.67 {$\scriptscriptstyle \pm 19.74$} \\
Qwen2-VL~\citep{Qwen2-VL} (bbox)    & 20.48 {$\scriptscriptstyle \pm 2.33$} & 5.75 {$\scriptscriptstyle \pm 2.59$} & 66.33 {$\scriptscriptstyle \pm 3.36$} \\
\hline
GENNAV (ours) & \textbf{34.32 {$\scriptscriptstyle \pm 2.97$}} & \textbf{28.75 {$\scriptscriptstyle \pm 5.08$}} & \textbf{67.50 {$\scriptscriptstyle \pm 2.95$}} \\
\Xhline{1pt}
\end{tabular}
}
\vspace{-7mm}
\label{tab:quantitative-physical}
\end{wraptable}

\vspace{-2mm}
These results show that GENNAV achieved the highest msIoU among the baseline methods.
Overall, GENNAV achieved higher $\mathrm{P}@0.1$ (28.75\%) and accuracy (67.50\%) than the baseline methods in the real-world experiments conducted in a zero-shot manner.  

\vspace{-3mm}
% \textbf{Qualitative Results.} 
\subsection{Qualitative Results}
\begin{wrapfigure}{r}{0.65\textwidth}
    \centering
    \vspace{-4mm} % 必要に応じて調整
    \begin{minipage}[t]{0.32\textwidth}
        \includegraphics[width=\linewidth]{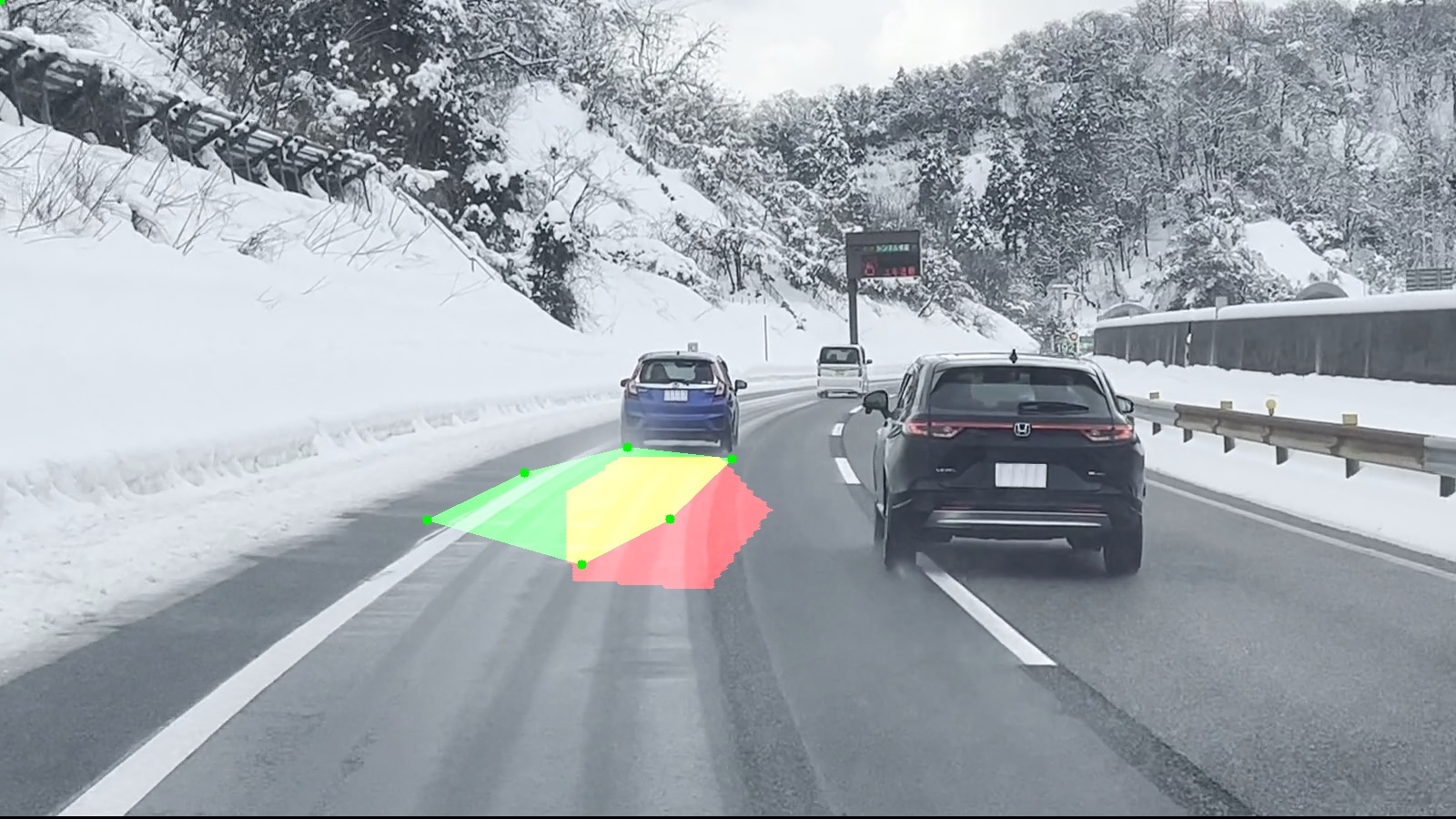}
        \vspace{2pt}
        \raggedright {\small (i) $\bm{x}_\text{inst}$: ``Please go near the blue car traveling in the same direction.''}
    \end{minipage}%
    \hfill
    \begin{minipage}[t]{0.32\textwidth}
        \includegraphics[width=\linewidth]{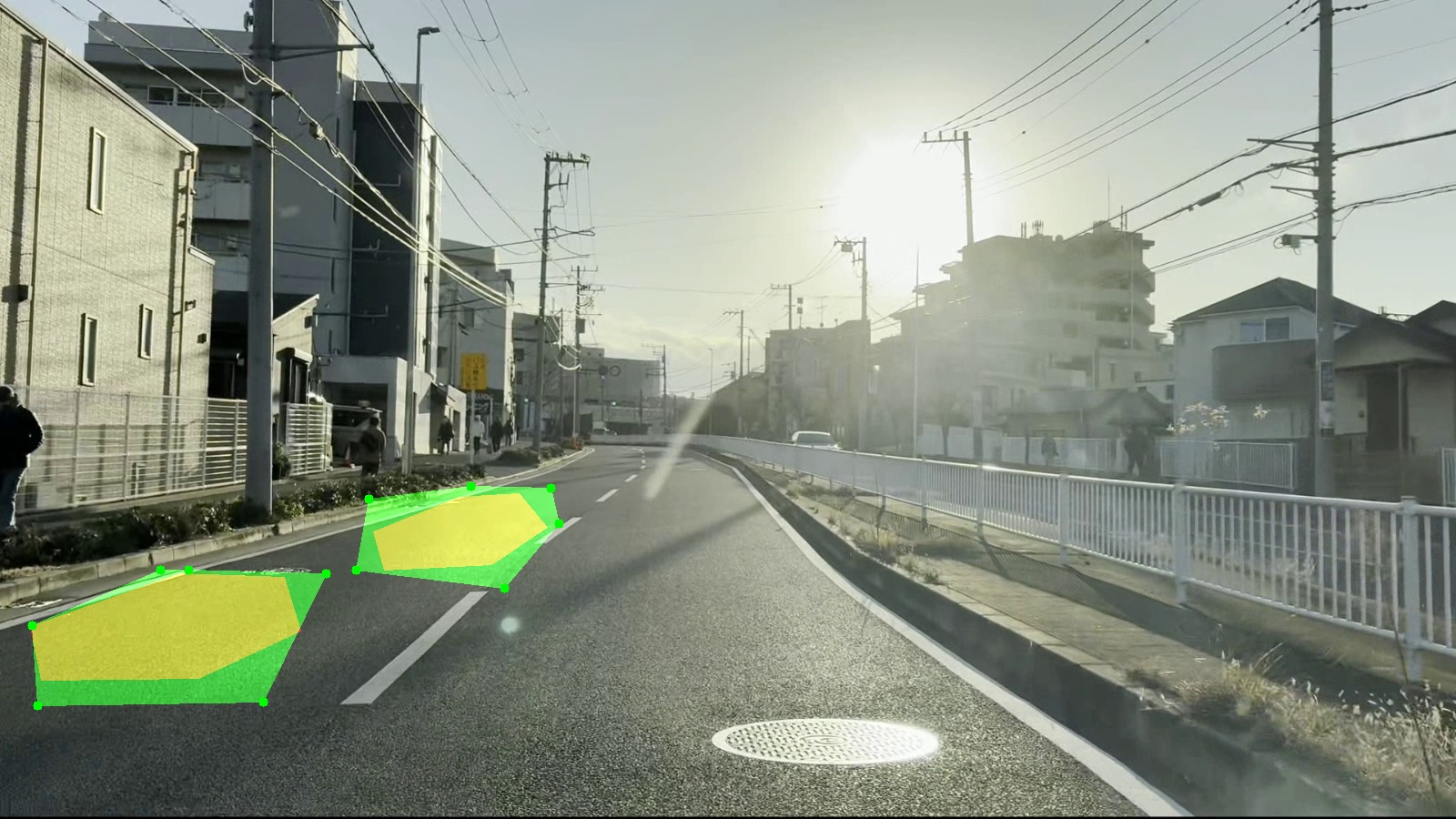}
        \vspace{2pt}
        \raggedright {\small (ii) $\bm{x}_\text{inst}$: ``Stop to the right of the pedestrian on the left side.''}
    \end{minipage}
    \vspace{-2mm}
    \caption{
      Qualitative results of GENNAV in the real world experiment. 
      The color of regions are the same as in Fig.~\ref{fig:model}
    }
    \vspace{-7mm} % 必要に応じて調整
    \label{fig:realworld-qualitative-success}
\end{wrapfigure}

\vspace{-2mm}
Fig.~\ref{fig:realworld-qualitative-success} shows successful cases of GENNAV in the real-world experiments.
Fig. \ref{fig:realworld-qualitative-success} (i) shows a single-target case and Fig.~\ref{fig:realworld-qualitative-success} (ii) shows a multi-target case.
Fig. \ref{fig:realworld-qualitative-success} (i) presents an example where $\bm{x}_\text{inst}$ was ``Please go near the blue car traveling in the same direction.''
GENNAV appropriately generated the region behind the blue car traveling in the same direction.

\vspace{-2mm}
Fig. \ref{fig:realworld-qualitative-success} (ii) illustrates an example of the $\bm{x}_\text{inst}$, ``Stop to the right of the pedestrian on the left side.''
The models were required to generate a mask for a region to the right of each pedestrian on the left side.
The proposed model successfully generated the mask appropriately for each of the two pedestrians.

\vspace{-4mm}
\section{Conclusion}
\vspace{-3mm}
% 7-1
In this paper, we focused on the GRNR task and proposed GENNAV, which predicts whether target regions exist and generates segmentation masks in cases that involve stuff-type target regions.
% which predicts both the existence and location of target regions based on a natural language instruction and a front camera image taken from a moving mobility. 
% 7-2
Our contributions can be summarized as follows.
We introduced GENNAV, which consists of three novel modules.
(i) ExPo, which predicts the existence of target regions and generates polygon-based segmentation masks.
(ii) LDPM, which  models fine-grained visual representations related to potential landmarks by patchifying images based on the spatial distribution of landmarks.
(iii) VLSiM, which models the relationship between language and spatial information.
We constructed the GRiN-Drive benchmark, which includes three distinct types of samples: single-target, no-target, and multi-target scenarios.
Our proposed method GENNAV outperformed the baseline methods on the GRiN-Drive benchmark.
In real-world experiments, GENNAV outperformed the baseline methods, including the MLLMs, under a zero-shot transfer setting. These results indicate that GENNAV can be effectively integrated into real-world systems.
% \begin{itemize}
%     \setlength{\parskip}{0mm} % 段落間
%     \setlength{\itemsep}{0.2mm} % 項目間
%     \item We introduced the ExPo, which predicts the existence of target regions and generates polygon-based segmentation masks by integrating four types of information: linguistic features, estimated road surface features, fine-grained visual representations and spatially grounded multimodal representations.
%     \item We incorporated the LDPM, which  models fine-grained visual representations related to potential landmarks by patchifying images based on the spatial distribution of landmarks.
%     \item We introduced the {\color{red}VLSiM}, which models the relationship between language and spatial information by integrating features derived from an image overlaid with a pseudo-depth image and a segmentation mask of the road region, and aligning the resulting representation with linguistic features.
%     \item In this study, we constructed the GRiN-Drive benchmark, which includes three distinct types of samples: single-target, no-target, and multi-target scenarios.
%     \item GENNAV outperformed all baseline methods on the GRiN-Drive benchmark.
%     \item In real-world experiments, GENNAV outperformed all baseline methods, including MLLMs, under a zero-shot transfer setting. These results indicate that GENNAV can be effectively integrated into real-world systems.
% \end{itemize}

\clearpage
\vspace{-4mm}
\section*{Limitations}
\vspace{-3mm}
Although we obtained promising results, the proposed method has several limitations.
First, it exhibits limited semantic understanding of landmarks, which can lead to an inappropriate understanding of the scene. 
To address this, one potential solution is to overlay masks generated by more robust semantic segmentation models (e.g., Grounded SAM~\citep{ren2024grounded}, Detic~\citep{zhou2022detecting}, Grounding DINO~\citep{liu2023grounding} and Open-Vocabulary SAM~\citep{yuan2024ovsam}) onto the input images before processing them with LDPM. 
Second, the current method lacks temporal modeling, which may lead to inconsistent predictions when applied to video-based GRES tasks. 
Incorporating information from adjacent frames could improve the coherence of predictions across time.
In future work, we may explore temporal modeling strategies to maintain consistency across frames and improve overall performance.

\bibliography{reference}  % .bib

\begin{thebibliography}{59}
\providecommand{\natexlab}[1]{#1}
\providecommand{\url}[1]{\texttt{#1}}
\expandafter\ifx\csname urlstyle\endcsname\relax
  \providecommand{\doi}[1]{doi: #1}\else
  \providecommand{\doi}{doi: \begingroup \urlstyle{rm}\Url}\fi

\bibitem[Rufus et~al.(2021)Rufus, Jain, Nair, Gandhi, and Krishna]{rufus2021grounding}
N.~Rufus, K.~Jain, K.~Nair, V.~Gandhi, and M.~Krishna.
\newblock {Grounding Linguistic Commands to Navigable Regions}.
\newblock In \emph{IROS}, pages 8593--8600, 2021.

\bibitem[Hosomi et~al.(2024)Hosomi, Hatanaka, Iioka, Yang, Kuyo, Misu, Yamada, and Sugiura]{tnrsmral24}
N.~Hosomi, S.~Hatanaka, Y.~Iioka, W.~Yang, K.~Kuyo, T.~Misu, K.~Yamada, and K.~Sugiura.
\newblock {Trimodal Navigable Region Segmentation Model: Grounding Navigation Instructions in Urban Areas}.
\newblock \emph{IEEE RA-L}, 9\penalty0 (5):\penalty0 4162--4169, 2024.

\bibitem[Kirillov et~al.(2019)Kirillov, He, Girshick, Rother, and Doll{\'a}r]{kirillov2019panoptic}
A.~Kirillov, K.~He, R.~Girshick, C.~Rother, and P.~Doll{\'a}r.
\newblock {Panoptic Segmentation}.
\newblock In \emph{CVPR}, pages 9404--9413, 2019.

\bibitem[Liu et~al.(2023)Liu, Ding, and Jiang]{liu2023gres}
C.~Liu, H.~Ding, and X.~Jiang.
\newblock {GRES: Generalized Referring Expression Segmentation}.
\newblock In \emph{CVPR}, pages 23592--23601, 2023.

\bibitem[Xia et~al.(2024)Xia, Han, Han, Pan, Song, and Huang]{xia2024gsva}
Z.~Xia, D.~Han, Y.~Han, X.~Pan, S.~Song, and G.~Huang.
\newblock {GSVA: Generalized Segmentation via Multimodal Large Language Models}.
\newblock In \emph{CVPR}, pages 3858--3869, 2024.

\bibitem[Nishimura et~al.(2024)Nishimura, Kuyo, Kambara, and Sugiura]{nishimura24iros}
T.~Nishimura, K.~Kuyo, M.~Kambara, and K.~Sugiura.
\newblock {Object Segmentation from Open-Vocabulary Manipulation Instructions Based on Optimal Transport Polygon Matching with Multimodal Foundation Models}.
\newblock In \emph{IROS}, pages 9549--9556, 2024.

\bibitem[Liu et~al.(2023)Liu, Ding, Cai, Zhang, Satzoda, Mahadevan, and Manmatha]{liu2023polyformer}
J.~Liu, H.~Ding, Z.~Cai, Y.~Zhang, R.~Satzoda, V.~Mahadevan, and R.~Manmatha.
\newblock {PolyFormer: Referring Image Segmentation as Sequential Polygon Generation}.
\newblock In \emph{CVPR}, pages 18653--18663, 2023.

\bibitem[Cui et~al.(2024)Cui, Ma, Cao, Ye, Zhou, Liang, Chen, Lu, Yang, Liao, et~al.]{cui2024survey}
C.~Cui, Y.~Ma, X.~Cao, W.~Ye, Y.~Zhou, K.~Liang, J.~Chen, J.~Lu, Z.~Yang, K.-D. Liao, et~al.
\newblock {A Survey on Multimodal Large Language Models for Autonomous Driving}.
\newblock In \emph{WACV}, pages 958--979, 2024.

\bibitem[Liu et~al.(2024)Liu, Yurtsever, Fossaert, Zhou, Zimmer, Cui, Zagar, and Knoll]{datasetsurveyMingyu}
M.~Liu, E.~Yurtsever, J.~Fossaert, X.~Zhou, W.~Zimmer, Y.~Cui, B.~Zagar, and A.~Knoll.
\newblock {A Survey on Autonomous Driving Datasets: Statistics, Annotation Quality, and a Future Outlook}.
\newblock \emph{IEEE T-IV}, pages 1--29, 2024.

\bibitem[Huang et~al.(2022)Huang, Shi, Li, Li, Huang, and Li]{huang2022multi}
K.~Huang, B.~Shi, X.~Li, X.~Li, S.~Huang, and Y.~Li.
\newblock {Multi-modal Sensor Fusion for Auto Driving Perception: A Survey}.
\newblock \emph{arXiv preprint arXiv:2202.02703}, 2022.

\bibitem[Zhou et~al.(2024)Zhou, Liu, Yurtsever, Zagar, Zimmer, Cao, and Knoll]{zhou2024vision}
X.~Zhou, M.~Liu, E.~Yurtsever, B.~Zagar, W.~Zimmer, H.~Cao, and A.~Knoll.
\newblock {Vision Language Models in Autonomous Driving: A Survey and Outlook}.
\newblock \emph{IEEE T-IV}, pages 1--20, 2024.

\bibitem[Du et~al.(2024)Du, Lei, Zhao, and Su]{ikun_Du_2024_CVPR}
Y.~Du, C.~Lei, Z.~Zhao, and F.~Su.
\newblock {iKUN: Speak to Trackers without Retraining}.
\newblock In \emph{CVPR}, pages 19135--19144, 2024.

\bibitem[Nguyen et~al.(2023)Nguyen, Quach, Kitani, and Luu]{NEURIPS2023_typetotrack}
P.~Nguyen, K.~Quach, K.~Kitani, and K.~Luu.
\newblock {Type-to-Track: Retrieve Any Object via Prompt-based Tracking}.
\newblock In \emph{NeurIPS}, volume~36, pages 3205--3219, 2023.

\bibitem[Deruyttere et~al.(2019)Deruyttere, Vandenhende, Grujicic, Gool, and Moens]{deruyttere-etal-2019-talk2car}
T.~Deruyttere, S.~Vandenhende, D.~Grujicic, L.~Gool, and M.~Moens.
\newblock {Talk2{C}ar: Taking Control of Your Self-Driving Car}.
\newblock In \emph{EMNLP}, pages 2088--2098, 2019.

\bibitem[Wu et~al.(2023)Wu, Han, Wang, Dong, Zhang, and Shen]{wu2023referringRMOT}
D.~Wu, W.~Han, T.~Wang, X.~Dong, X.~Zhang, and J.~Shen.
\newblock {Referring Multi-Object Tracking}.
\newblock In \emph{CVPR}, pages 14633--14642, 2023.

\bibitem[Zhang et~al.(2024)Zhang, Wu, Han, and Dong]{zhang2024bootstrapping}
Y.~Zhang, D.~Wu, W.~Han, and X.~Dong.
\newblock {Bootstrapping Referring Multi-Object Tracking}.
\newblock \emph{arXiv preprint arXiv:2406.05039}, 2024.

\bibitem[Kamath et~al.(2021)Kamath, Singh, LeCun, Synnaeve, Misra, and Carion]{kamath2021mdetr}
A.~Kamath, M.~Singh, Y.~LeCun, G.~Synnaeve, I.~Misra, and N.~Carion.
\newblock {MDETR-Modulated Detection for End-to-end Multi-modal Understanding}.
\newblock In \emph{ICCV}, pages 1780--1790, 2021.

\bibitem[Zeng et~al.(2022)Zeng, Dong, Zhang, Wang, Zhang, and Wei]{zeng2022motr}
F.~Zeng, B.~Dong, Y.~Zhang, T.~Wang, X.~Zhang, and Y.~Wei.
\newblock {MOTR: End-to-End Multiple-Object Tracking with Transformer}.
\newblock In \emph{ECCV}, pages 659--675, 2022.

\bibitem[Zhang et~al.(2021)Zhang, Wang, Wang, Zeng, and Liu]{zhang2021fairmot}
Y.~Zhang, C.~Wang, X.~Wang, W.~Zeng, and W.~Liu.
\newblock {FairMOT: On the Fairness of Detection and Re-Identification in Multiple Object Tracking}.
\newblock \emph{IJCV}, 129:\penalty0 3069--3087, 2021.

\bibitem[Zhu et~al.(2022)Zhu, Zhou, Shen, Luo, Pan, Lin, Chen, Cao, Sun, and Ji]{zhu2022seqtr}
C.~Zhu, Y.~Zhou, Y.~Shen, G.~Luo, X.~Pan, M.~Lin, C.~Chen, L.~Cao, X.~Sun, and R.~Ji.
\newblock {SeqTR: A Simple yet Universal Network for Visual Grounding}.
\newblock In \emph{ECCV}, pages 598--615, 2022.

\bibitem[Cheng et~al.(2024)Cheng, Li, Jin, Li, Ji, Yuan, Liu, and Chen]{cheng2024parallel}
Z.~Cheng, K.~Li, P.~Jin, S.~Li, X.~Ji, L.~Yuan, C.~Liu, and J.~Chen.
\newblock {Parallel Vertex Diffusion for Unified Visual Grounding}.
\newblock In \emph{AAAI}, pages 1326--1334, 2024.

\bibitem[Yang et~al.(2022)Yang, Wang, Tang, Chen, Zhao, and Torr]{yang2022lavt}
Z.~Yang, J.~Wang, Y.~Tang, K.~Chen, H.~Zhao, and P.~H. Torr.
\newblock {LAVT: Language-Aware Vision Transformer for Referring Image Segmentation}.
\newblock In \emph{CVPR}, pages 18155--18165, 2022.

\bibitem[Lai et~al.(2024)Lai, Tian, Chen, Li, Yuan, Liu, and Jia]{lai2024lisa}
X.~Lai, Z.~Tian, Y.~Chen, Y.~Li, Y.~Yuan, S.~Liu, and J.~Jia.
\newblock {LISA: Reasoning Segmentation via Large Language Model}.
\newblock In \emph{CVPR}, pages 9579--9589, 2024.

\bibitem[Cheng et~al.(2015)Cheng, Zheng, Lin, Vineet, Sturgess, Crook, Mitra, and Torr]{imagespirit}
M.~Cheng, S.~Zheng, W.~Lin, V.~Vineet, P.~Sturgess, N.~Crook, N.~Mitra, and P.~Torr.
\newblock {ImageSpirit: Verbal Guided Image Parsing}.
\newblock \emph{ACM Trans. Graph.}, 34\penalty0 (1), 2015.

\bibitem[Hu et~al.(2023)Hu, Wang, Shao, Xie, Li, Han, and Luo]{hu2023beyond}
Y.~Hu, Q.~Wang, W.~Shao, E.~Xie, Z.~Li, J.~Han, and P.~Luo.
\newblock {Beyond One-to-One: Rethinking the Referring Image Segmentation}.
\newblock In \emph{ICCV}, pages 4067--4077, 2023.

\bibitem[Luo et~al.(2024)Luo, Wu, Cheng, Liu, Xiao, Wang, Zhang, and Yang]{luo2024cohd}
Z.~Luo, Y.~Wu, T.~Cheng, Y.~Liu, Y.~Xiao, H.~Wang, X.~Zhang, and Y.~Yang.
\newblock {CoHD: A Counting-Aware Hierarchical Decoding Framework for Generalized Referring Expression Segmentation}.
\newblock \emph{arXiv preprint arXiv:2405.15658}, 2024.

\bibitem[Li et~al.(2024)Li, Zhao, Bai, and Su]{li2024bring}
W.~Li, Z.~Zhao, H.~Bai, and F.~Su.
\newblock {Bring Adaptive Binding Prototypes to Generalized Referring Expression Segmentation}.
\newblock \emph{arXiv preprint arXiv:2405.15169}, 2024.

\bibitem[Hosomi et~al.(2025)Hosomi, Iioka, Hatanaka, Misu, Yamada, Tsukamoto, Kobayashi, and Sugiura]{hosomitrip2025}
N.~Hosomi, Y.~Iioka, S.~Hatanaka, T.~Misu, K.~Yamada, N.~Tsukamoto, S.~Kobayashi, and K.~Sugiura.
\newblock {Multimodal Target Localization With Landmark-Aware Positioning for Urban Mobility}.
\newblock \emph{IEEE RA-L}, 10\penalty0 (1):\penalty0 716--723, 2025.

\bibitem[Anderson et~al.(2018)Anderson, Wu, Teney, Bruce, Johnson, S{\"u}nderhauf, Reid, Gould, and Hengel]{anderson2018vision}
P.~Anderson, Q.~Wu, D.~Teney, J.~Bruce, M.~Johnson, N.~S{\"u}nderhauf, I.~Reid, S.~Gould, and A.~Hengel.
\newblock {Vision-and-Language Navigation: Interpreting Visually-Grounded Navigation Instructions in Real Environments}.
\newblock In \emph{CVPR}, pages 3674--3683, 2018.

\bibitem[Zhu et~al.(2021)Zhu, Liang, Zhu, Yu, Chang, and Liang]{zhu2021soon}
F.~Zhu, X.~Liang, Y.~Zhu, Q.~Yu, X.~Chang, and X.~Liang.
\newblock {SOON: Scenario Oriented Object Navigation with Graph-Based Exploration}.
\newblock In \emph{CVPR}, pages 12689--12699, 2021.

\bibitem[Qi et~al.(2020)Qi, Wu, Anderson, Wang, Wang, Shen, and Hengel]{qi2020reverie}
Y.~Qi, Q.~Wu, P.~Anderson, X.~Wang, Y.~Wang, C.~Shen, and A.~Hengel.
\newblock {REVERIE: Remote Embodied Visual Referring Expression in Real Indoor Environments}.
\newblock In \emph{CVPR}, pages 9982--9991, 2020.

\bibitem[Zhang et~al.(2024)Zhang, Wang, Xu, Zhou, Hong, Fang, Wu, Zhang, and Wang]{zhang2024navid}
J.~Zhang, K.~Wang, R.~Xu, G.~Zhou, Y.~Hong, X.~Fang, Q.~Wu, Z.~Zhang, and H.~Wang.
\newblock {NaVid: Video-based VLM Plans the Next Step for Vision-and-Language Navigation}.
\newblock In \emph{RSS}, 2024.

\bibitem[Krantz et~al.(2020)Krantz, Wijmans, Majumdar, Batra, and Lee]{krantz2020beyond}
J.~Krantz, E.~Wijmans, A.~Majumdar, D.~Batra, and S.~Lee.
\newblock {Beyond the Nav-Graph: Vision-and-Language Navigation in Continuous Environments}.
\newblock In \emph{ECCV}, pages 104--120, 2020.

\bibitem[Deruyttere et~al.(2022)Deruyttere, Grujicic, Blaschko, and Moens]{deruyttere2022talk2car}
T.~Deruyttere, D.~Grujicic, M.~B. Blaschko, and M.-F. Moens.
\newblock {Talk2car: Predicting Physical Trajectories for Natural Language Commands}.
\newblock \emph{IEEE Access}, 10:\penalty0 123809--123834, 2022.

\bibitem[Omama et~al.(2023)Omama, Inani, Paul, and et~al.]{omama2023alt}
M.~Omama, P.~Inani, P.~Paul, and et~al.
\newblock {ALT-Pilot: Autonomous navigation with Language augmented Topometric maps}.
\newblock \emph{arXiv preprint arXiv:2310.02324}, 2023.

\bibitem[Shah et~al.(2022)Shah, Osi{\'n}ski, Ichter, and Levine]{shah2022lmnav}
D.~Shah, B.~Osi{\'n}ski, B.~Ichter, and S.~Levine.
\newblock {LM-Nav: Robotic Navigation with Large Pre-Trained Models of Language, Vision, and Action}.
\newblock In \emph{CoRL}, pages 492--504, 2022.

\bibitem[Jain et~al.(2023)Jain, Chhangani, Tiwari, Krishna, and Gandhi]{jain2023ground}
K.~Jain, V.~Chhangani, A.~Tiwari, K.~M. Krishna, and V.~Gandhi.
\newblock {Ground then Navigate: Language-guided Navigation in Dynamic Scenes}.
\newblock In \emph{ICRA}, pages 4113--4120, 2023.

\bibitem[Xiang et~al.(2020)Xiang, Wang, and Wang]{xiang-etal-2020-learning}
J.~Xiang, X.~Wang, and Y.~Wang.
\newblock {Learning to Stop: A Simple yet Effective Approach to Urban Vision-Language Navigation}.
\newblock In \emph{EMNLP}, pages 699--707, 2020.

\bibitem[Schumann and Riezler(2022)]{schumann-riezler-2022-analyzing}
R.~Schumann and S.~Riezler.
\newblock {Analyzing Generalization of Vision and Language Navigation to Unseen Outdoor Areas}.
\newblock In \emph{ACL}, pages 7519--7532, 2022.

\bibitem[Dosovitskiy et~al.(2021)Dosovitskiy, Beyer, Kolesnikov, Weissenborn, Zhai, Unterthiner, et~al.]{dosovitskiy2021an}
A.~Dosovitskiy, L.~Beyer, A.~Kolesnikov, D.~Weissenborn, X.~Zhai, T.~Unterthiner, et~al.
\newblock {An Image is Worth 16x16 Words: Transformers for Image Recognition at Scale}.
\newblock In \emph{ICLR}, pages 12888--12900, 2021.

\bibitem[Yu et~al.(2020)Yu, Chen, Wang, Xian, Chen, Liu, Madhavan, and Darrell]{yu2020bdd100k}
F.~Yu, H.~Chen, X.~Wang, W.~Xian, Y.~Chen, F.~Liu, V.~Madhavan, and T.~Darrell.
\newblock {BDD100K: A Diverse Driving Dataset for Heterogeneous Multitask Learning}.
\newblock In \emph{CVPR}, pages 2636--2645, 2020.

\bibitem[Cordts et~al.(2016)Cordts, Omran, Ramos, Rehfeld, Enzweiler, Benenson, Franke, Roth, and Schiele]{cordts2016cityscapes}
M.~Cordts, M.~Omran, S.~Ramos, T.~Rehfeld, M.~Enzweiler, R.~Benenson, U.~Franke, S.~Roth, and B.~Schiele.
\newblock {The Cityscapes Dataset for Semantic Urban Scene Understanding}.
\newblock In \emph{CVPR}, pages 3213--3223, 2016.

\bibitem[Geiger et~al.(2012)Geiger, Lenz, and Urtasun]{kittiGeiger2012CVPR}
A.~Geiger, P.~Lenz, and R.~Urtasun.
\newblock {Are we ready for Autonomous Driving? The KITTI Vision Benchmark Suite}.
\newblock In \emph{CVPR}, pages 3354--3361, 2012.

\bibitem[Caesar et~al.(2020)Caesar, Bankiti, Lang, Vora, Liong, Xu, Krishnan, Pan, Baldan, and Beijbom]{caesar2020nuscenes}
H.~Caesar, V.~Bankiti, A.~Lang, S.~Vora, V.~Liong, Q.~Xu, A.~Krishnan, Y.~Pan, G.~Baldan, and O.~Beijbom.
\newblock {nuScenes: A Multimodal Dataset for Autonomous Driving}.
\newblock In \emph{CVPR}, pages 11621--11631, 2020.

\bibitem[Yang et~al.(2024)Yang, Kang, Huang, Zhao, Xu, Feng, and Zhao]{depth_anything_v2}
L.~Yang, B.~Kang, Z.~Huang, Z.~Zhao, X.~Xu, J.~Feng, and H.~Zhao.
\newblock {Depth Anything V2}.
\newblock \emph{arXiv:2406.09414}, 2024.

\bibitem[Solatorio(2024)]{solatorio2024gistembed}
A.~Solatorio.
\newblock {GISTEmbed: Guided In-sample Selection of Training Negatives for Text Embedding Fine-tuning}.
\newblock \emph{arXiv preprint arXiv:2402.16829}, 2024.

\bibitem[Oquab et~al.(2023)Oquab, Darcet, Moutakanni, Vo, Szafraniec, Khalidov, Fernandez, Haziza, Massa, El-Nouby, Howes, Huang, Xu, Sharma, Li, Galuba, et~al.]{oquab2023dinov2}
M.~Oquab, T.~Darcet, T.~Moutakanni, H.~Vo, M.~Szafraniec, V.~Khalidov, P.~Fernandez, D.~Haziza, F.~Massa, A.~El-Nouby, R.~Howes, Y.~Huang, H.~Xu, V.~Sharma, S.-W. Li, W.~Galuba, et~al.
\newblock {DINOv2: Learning Robust Visual Features without Supervision}, 2023.

\bibitem[Xu et~al.(2023)Xu, Xiong, and Bhattacharyya]{xu2023pidnet}
J.~Xu, Z.~Xiong, and S.~Bhattacharyya.
\newblock {PIDNet: A Real-Time Semantic Segmentation Network Inspired by PID Controllers}.
\newblock In \emph{CVPR}, pages 19529--19539, 2023.

\bibitem[Reid et~al.(2024)Reid, Savinov, Teplyashin, Lepikhin, Lillicrap, et~al.]{reid2024gemini}
M.~Reid, N.~Savinov, D.~Teplyashin, D.~Lepikhin, T.~Lillicrap, et~al.
\newblock {Gemini 1.5: Unlocking Multimodal Understanding Across Millions of Tokens of Context}.
\newblock \emph{arXiv preprint arXiv:2403.05530}, 2024.

\bibitem[Achiam et~al.(2023)Achiam, Adler, Agarwal, et~al.]{gpt4o}
J.~Achiam, S.~Adler, S.~Agarwal, et~al.
\newblock {GPT-4 Technical Report}.
\newblock \emph{arXiv preprint arXiv:2303.08774}, 2023.

\bibitem[Wang et~al.(2024)Wang, Bai, Tan, Wang, Fan, Bai, Chen, Liu, Wang, Ge, Fan, Dang, Du, Ren, Men, et~al.]{Qwen2-VL}
P.~Wang, S.~Bai, S.~Tan, S.~Wang, Z.~Fan, J.~Bai, K.~Chen, X.~Liu, J.~Wang, W.~Ge, Y.~Fan, K.~Dang, M.~Du, X.~Ren, R.~Men, et~al.
\newblock {Qwen2-VL: Enhancing Vision-Language Model's Perception of the World at Any Resolution}.
\newblock \emph{arXiv preprint arXiv:2409.12191}, 2024.

\bibitem[Ren et~al.(2024)Ren, Liu, Zeng, Lin, Li, Cao, Chen, Huang, Chen, Yan, Zeng, et~al.]{ren2024grounded}
T.~Ren, S.~Liu, A.~Zeng, J.~Lin, K.~Li, H.~Cao, J.~Chen, X.~Huang, Y.~Chen, F.~Yan, Z.~Zeng, et~al.
\newblock {Grounded SAM: Assembling Open-World Models for Diverse Visual Tasks}.
\newblock \emph{arXiv preprint arXiv:2401.14159}, 2024.

\bibitem[Zhou et~al.(2022)Zhou, Girdhar, Joulin, Kr{\"a}henb{\"u}hl, and Misra]{zhou2022detecting}
X.~Zhou, R.~Girdhar, A.~Joulin, P.~Kr{\"a}henb{\"u}hl, and I.~Misra.
\newblock {Detecting Twenty-thousand Classes using Image-level Supervision}.
\newblock In \emph{ECCV}, pages 350--368, 2022.

\bibitem[Liu et~al.(2023)Liu, Zeng, Ren, Li, Zhang, Yang, Li, Yang, Su, Zhu, et~al.]{liu2023grounding}
S.~Liu, Z.~Zeng, T.~Ren, F.~Li, H.~Zhang, J.~Yang, C.~Li, J.~Yang, H.~Su, J.~Zhu, et~al.
\newblock {Grounding DINO: Marrying DINO with Grounded Pre-Training for Open-Set Object Detection}.
\newblock \emph{arXiv preprint arXiv:2303.05499}, 2023.

\bibitem[Yuan et~al.(2024)Yuan, Li, Zhou, Li, Chen, and Loy]{yuan2024ovsam}
H.~Yuan, X.~Li, C.~Zhou, Y.~Li, K.~Chen, and C.~Loy.
\newblock {Open-Vocabulary SAM: Segment and Recognize Twenty-thousand Classes Interactively}.
\newblock In \emph{ECCV}, pages 419--437, 2024.

\bibitem[Mao et~al.(2016)Mao, Huang, Toshev, Camburu, Yuille, and Murphy]{mao2016generation}
J.~Mao, J.~Huang, A.~Toshev, O.~Camburu, A.~Yuille, and K.~Murphy.
\newblock {Generation and Comprehension of Unambiguous Object Descriptions}.
\newblock In \emph{CVPR}, pages 11--20, 2016.

\bibitem[Fiedler et~al.(2019)Fiedler, Bestmann, and Hendrich]{imagetagger2018}
N.~Fiedler, M.~Bestmann, and N.~Hendrich.
\newblock {ImageTagger: An Open Source Online Platform for Collaborative Image Labeling}.
\newblock In \emph{RoboCup}, pages 162--169, 2019.

\bibitem[Liu et~al.(2021)Liu, Lin, Cao, Hu, Wei, Zhang, Lin, and Guo]{liu2021swin}
Z.~Liu, Y.~Lin, Y.~Cao, H.~Hu, Y.~Wei, Z.~Zhang, S.~Lin, and B.~Guo.
\newblock {Swin Transformer: Hierarchical Vision Transformer Using Shifted Windows}.
\newblock In \emph{ICCV}, pages 10012--10022, 2021.

\bibitem[Kumar et~al.(2021)Kumar, Kashiyama, Maeda, and Sekimoto]{vehicleorientation}
A.~Kumar, T.~Kashiyama, H.~Maeda, and Y.~Sekimoto.
\newblock {Citywide Reconstruction of Cross-Sectional Traffic Flow from Moving Camera Videos}.
\newblock In \emph{Big Data}, pages 1670--1678, 2021.

\end{thebibliography}

\clearpage
{ \huge \textbf{Appendix}}
\setcounter{section}{0}   % セクション番号をリセット
\renewcommand{\thesection}{\Alph{section}}  % A, B, C … にする

\setcounter{figure}{4}

\setcounter{table}{3} % A5 にしたければ 4 にセット（LaTeX は次の番号が使われる）

\setcounter{equation}{4}
% \vspace{-2mm}
% \section{Additional Related Work}

\vspace{-2mm}
\section{Task Details}
\vspace{-4mm}

\begin{figure}[h]
  \centering
    \centering
    \includegraphics[width=\linewidth]{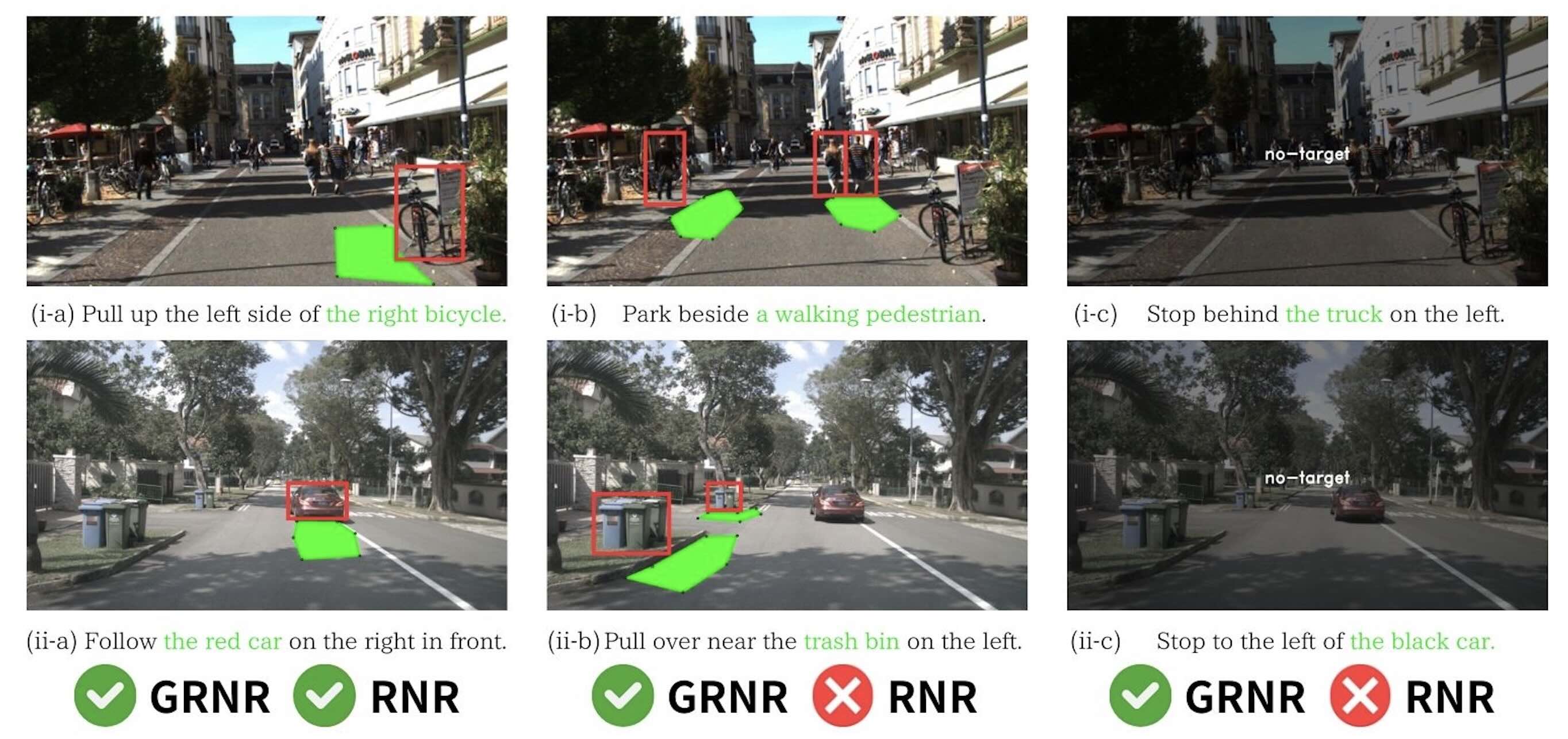}
  \caption{Typical examples of the GRNR task. Left: single target. Center: multi-target. Right: no-target. The goal is to generate zero or more segmentation masks (shown in green). Unlike the RNR, the GRNR accommodates instructions that specify an arbitrary number of landmarks, including cases where multiple target regions exist or no target region exists.}
  \label{fig:sample}
  % \vspace{-6mm}
\end{figure}
Fig. \ref{fig:sample} shows typical scenes from the Generalized Referring Navigable Regions (GRNR) task.
The bounding boxes in the figure indicate the landmarks referenced in the instructions.
In Fig.~\ref{fig:sample}~(i-a), given the instruction ``Pull up the left side of the right bicycle,'' the model is required to identify the target region and generate the mask depicted in green.
Similarly, in Fig.~\ref{fig:sample}~(i-b), because the instruction is ``Park beside a walking pedestrian'' and multiple pedestrians exist in the image, the model should generate a mask for each of the pedestrians.
Finally, in Fig.~\ref{fig:sample}~(i-c), the instruction ``Stop behind the truck on the left'' is provided but no truck exists, therefore the model should determine that there is no target region in the given image.
Unlike the RNR task, the GRNR task involves instructions that specify any number of target regions, including multi-target and no-target instructions.

The input to this task is a front camera image and a navigation instruction, and the output is a binary indicator that represents whether at least one target region is present in the image, accompanied by a set of segmentation masks (with no masks generated if no target region is present).
% Furthermore, incorporating trajectory prediction into this task would significantly increase its complexity, requiring additional trajectory annotations and a substantial increase in the amount of necessary data.

% \vspace{-2mm}
% \section{Details of Modules}
% \vspace{-4mm}

% \vspace{-2mm}
% \section{Experimental Settings}

% \vspace{-2mm}
\section{GRiN-Drive Details}
\begin{table}[ht]
    \centering
    \caption{
    Comparison of datasets.
    }
    % \vspace{-4mm}
    \resizebox{\linewidth}{!}{%
    \begin{tabular}{lcccccc}
        \toprule
        % \multirow{2}{*}{Dataset} & \multirow{2}{*}{Images} & \multirow{2}{*}{Region type} & \multicolumn{3}{c}{Number of targets} & \multirow{2}{*}{Expression type} \\
        % \cmidrule(lr){4-6}
        % & & & No & Single & Multi & \\
        Dataset & Images & Region type & No & Single & Multi & Expression type \\
        % \cmidrule(lr){4-6}
        % & & & No & Single & Multi & \\
        \midrule
        Refcocog~\citep{mao2016generation}         & 26,711 & Thing & -- & \checkmark & -- & Phrase \\
        gRefcoco~\citep{liu2023gres}        & 19,992 & Thing & \checkmark & \checkmark & \checkmark & Phrase \\
        % Talk2car         & 9,217  & Thing & -- & \checkmark & -- & Instruction \\
        Talk2carRegSeg~\citep{rufus2021grounding}   & 10,016 & Stuff & -- & \checkmark & -- & Instruction \\
        Reffer-Ksitti-V2~\citep{wu2023referringRMOT}     & 6,650  & Thing & \checkmark & \checkmark & \checkmark & Phrase \\
        GRiN-Drive (ours)       & 17,114 & Stuff & \checkmark & \checkmark & \checkmark & Instruction \\
        \bottomrule
    \end{tabular}
    }
    \label{tab:datasets}
    % \vspace{-4.5mm}
\end{table}
% % 5-3
\begin{figure*}[h]
    \centering
    \includegraphics[width=\linewidth]{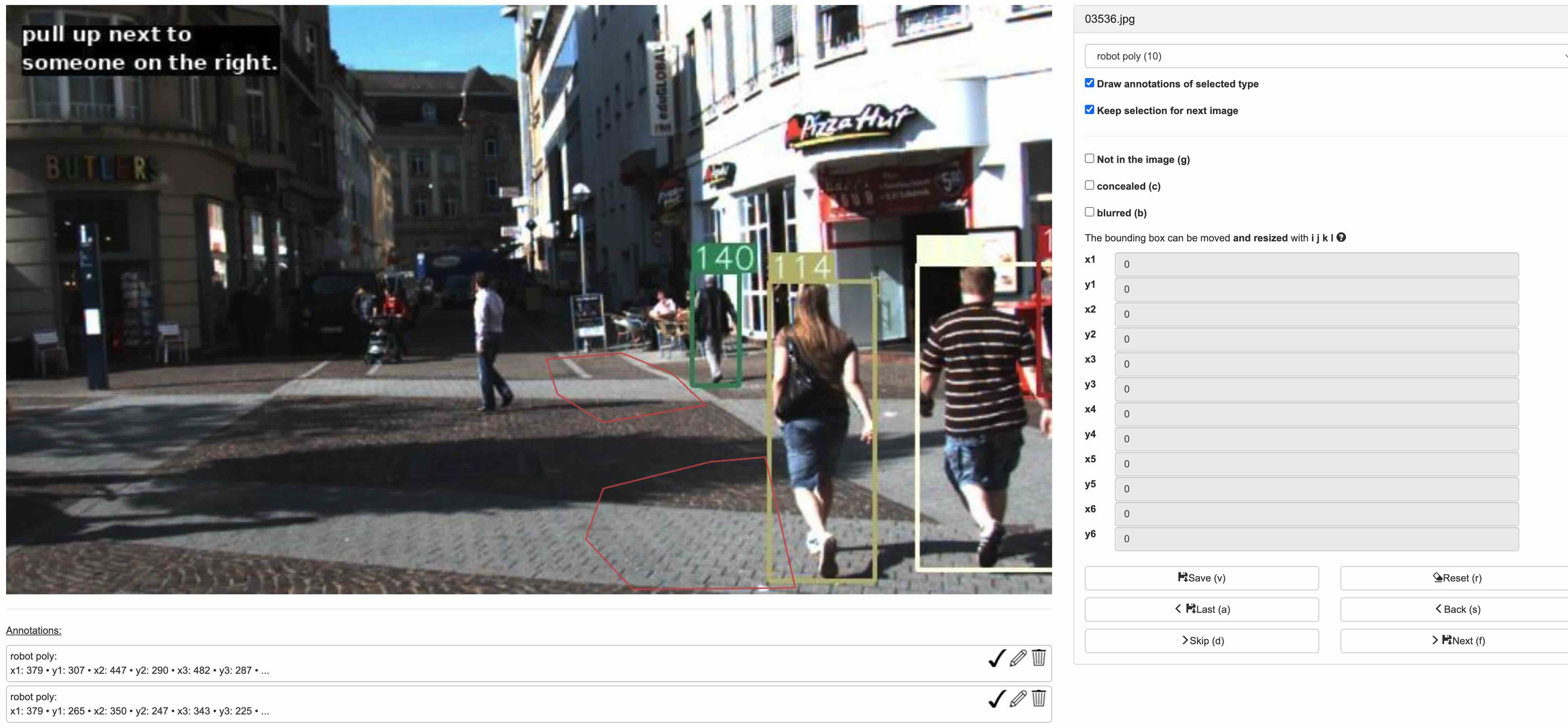}
    \caption{Annotation interface for multi-target samples in the GRiN-Drive benchmark. Annotators were instructed to provide polygons for an arbitrary number of target regions in the given image, corresponding to the navigation instruction.}
    \label{fig:annotation}
    \vspace{-2mm}
\end{figure*}

Most existing benchmarks lack coverage of no-target cases, which fundamentally limits their practicality.
In contrast, as summarised in Table~\ref{tab:datasets}, the GRiN-Drive benchmark comprehensively covers single-, no-, and multi-target scenarios, serving as a more representative benchmark of real-world scenarios.
To evaluate the performance of models on the three distinct types of cases in the GRNR task, we constructed the novel GRiN-Drive benchmark based on the Talk2Car-RegSeg~\citep{rufus2021grounding} and Refer-KITTI-V2~\citep{zhang2024bootstrapping} datasets.
The images from Talk2Car-RegSeg have a resolution of 1600×900, whereas those from Refer-KITTI-V2 were originally captured at 1280×384.
To standardize the aspect ratio to 16:9, the Refer-KITTI-V2 images were cropped to 656×369, with the cropping centered on the lower part of the image.
The segmentation masks of the multi-target samples in the GRiN-Drive benchmark were provided by 244 annotators, with an average of 29.07 samples per annotator.
We constructed the GRiN-Drive benchmark following the procedure described below.

% \noindent\textbf{Single / No-target Samples.}
\subsection{Single / No-target Samples}
We constructed single-target and no-target samples based on the Talk2Car-RegSeg dataset.
Given that it contains exclusively single-target samples, we created no-target samples using the following steps:
(i) We randomly selected samples from the dataset and swapped their instructions. 
(ii) We used GPT-4o~\cite{gpt4o} to assess whether the landmark specified in each swapped instruction existed in the corresponding image.
(iii) If GPT-4o indicated that the landmark was still present, we replaced the instruction again and repeated the assessment.
(iv) Finally, we manually inspected the samples classified as containing no target region.

\subsection{Multi-target Samples}
% \noindent\textbf{\noindent\textbf{Multi-target Samples.}}
To create multi-target samples, we leveraged the Refer-KITTI-V2~\citep{zhang2024bootstrapping} dataset, which was designed for Referring Multi-Object Tracking tasks in urban driving scenes.
We did not use the Talk2Car-RegSeg dataset to create multi-target samples because creating sample with a many-to-many relationship between landmarks and target regions using the Talk2Car-RegSeg dataset is challenging.
This is mainly because the test set of the Talk2Car-RegSeg dataset includes samples that contain no landmarks in navigation instruction. 
Moreover, because these samples lack positional annotations for objects other than landmarks, the construction of a multi-landmark test set from the Talk2Car-RegSeg dataset would be impractical because of the high annotation costs.
Therefore, we used the Refer-KITTI-V2 dataset as an alternative because it explicitly annotates multiple landmarks within a single image. 

To create the multi-target samples for the GRiN-Drive benchmark, we first extracted relevant images from Refer-KITTI-V2.
We selected frames from Refer-KITTI-V2 according to the following procedure: For each video, we identified frames containing multiple landmarks corresponding to noun phrases. 
Some landmarks may have been only partially visible, because we cropped wide-format images from the original dataset for our analysis.
To ensure the validity of landmarks, we only considered landmarks whose bounding boxes had more than half of their area within the image boundaries, thereby avoiding cases where objects were significantly truncated. To prevent redundancy and maintain diversity in the dataset, we excluded 20 subsequent frames adjacent to each selected frame for the same noun phrase instance.

Next, we generated navigation instructions for the extracted images based on Talk2Car-RegSeg instructions as follows: First, we leveraged MLLMs to create instruction templates from the original instructions in the Talk2Car-RegSeg dataset by replacing noun phrases that specify landmarks with <landmark> tags. For example, given a Talk2Car-RegSeg instruction such as ``Stop next to the pedestrian on the right,'' the MLLM generated the following template: ``Stop next to <landmark>pedestrian on the right</landmark>.'' Second, we converted the noun phrases from Refer-KITTI-V2 into their singular forms using MLLMs and verified the conversions manually. Third, we randomly replaced the <landmark> slots in the templates with singularized noun phrases to create the final navigation instructions.

Fig.~\ref{fig:annotation} shows the annotation interface used in this study.
Using the generated instruction-image pairs we asked annotators to identify the target regions for each sample.
% During the annotation process, the bounding boxes associated with noun phrases, originally annotated in Refer-KITTI-V2, were displayed as references to aid annotators.
To assist annotators during the annotation process, the bounding boxes associated with noun phrases were displayed as visual references.
For instructions referring to multiple landmarks, the annotators specified a target region that corresponds to each landmark separately using polygons with a minimum of six vertices.
We used SoSci Survey\footnote{https://www.soscisurvey.de/} to manage the annotation procedure and Imagetagger~\citep{imagetagger2018} to collect the annotation of target regions.
Finally, we manually removed inappropriate samples, such as cases with a significant overlap between target regions and landmarks or samples with unusually short response times, to enhance dataset quality.

\vspace{-4mm}
\section{
    Implementation Details
    \label{implementation_details}
}

\subsection{GENNAV}
% \begin{table}[t]
%     \normalsize
%     \centering
%     \caption{Experimental settings of GENNAV}
%     \label{tab:exp_settings}
%     \begin{tabular}{ll}
%         \toprule
%         Epoch & $100$ \\
%         \hline
%         Batch size & $384$ \\
%         \hline
%         Learning rate & $1\times10^{-4}$ \\
%         \hline
%         Optimizer & AdamW ($\beta_1=0.9$, $\beta_2=0.98$) \\
%         \hline
%         Loss weights & $\lambda_{\text{pt}}$ = $3$ \\
%         \bottomrule
%     \end{tabular}
%     \vspace{-5mm}
% \end{table}

\begin{wraptable}{r}{0.5\textwidth}
  \centering
  \caption{Experimental settings of GENNAV}
  \label{tab:exp_settings}
  \vspace{-2mm}
  \begin{tabular}{ll}
    \toprule
    Epoch & $100$ \\
    \hline
    Batch size & $384$ \\
    \hline
    Learning rate & $1\times10^{-4}$ \\
    \hline
    Optimizer & AdamW ($\beta_1=0.9$, $\beta_2=0.98$) \\
    \hline
    Loss weights & $\lambda_{\text{pt}} = 3$ \\
    \bottomrule
  \end{tabular}
  \vspace{-4mm}
\end{wraptable}

Table \ref{tab:exp_settings} shows the experimental settings for GENNAV.
In the Landmark Distribution Patch Module, we divided each region of 560×315 into patches and resized them to 224×224. 
For all other modules, we resized the original images to 224×224 to reduce the computational cost. 
% The details of the experimental setup of GENNAV  are explained in the supplementary materials.
To ensure a fair comparison, we evaluated the baseline method by Rufus et al., LAVT, and TNRSM at both 224×224 and an additional resolution of 640×640, which is equivalent to the resolution used in Landmark Distribution Patchification Module.
% 5-9

Our model had approximately 67.9M trainable parameters and 4.20T multiply-add operations.
% 5-10
We trained our model on a GeForce RTX 4090 with 24 GB of GPU memory and an Intel Core i9-13900KF with 64 GB of RAM.
% 5-11
The learning rate was warmed up during the first 5 epochs and then decayed by a factor of 0.1 after the 75th epoch.
The training time for the proposed model was approximately 3 hours, and the inference time was approximately 31.3 milliseconds.
% 5-12
We calculated the msIoU on the validation set every three epochs.
To evaluate performance on the test set, we used the model that achieved the highest msIoU on the validation set.

\subsection{Baselines}

We used two groups of baseline methods: pixel-based methods (a method by~\citet{rufus2021grounding}, LAVT~\citep{yang2022lavt}, TNRSM~\citep{tnrsmral24}, and GSVA-Vicuna-7B~\citep{xia2024gsva}) and MLLMs (Gemini~\citep{reid2024gemini}, GPT-4o~\citep{gpt4o}, and Qwen2-VL~\citep{Qwen2-VL}).
We selected Gemini, GPT-4o and Qwen2-VL because they are representative multimodal LLMs that have been pre-trained on large-scale datasets and have demonstrated outstanding performance on various vision-and-language tasks~\citep{reid2024gemini, gpt4o, Qwen2-VL}.
In our comparative experiments, we used eight baseline methods with the following experimental settings.

% \vspace{-2mm}

\textbf{Pixel-based methods (Rufus et al., LAVT, TNRSM, and GSVA-Vicuna-7B).}
We fine-tuned each model following the hyperparameter settings described in its respective paper.
If no mask is generated for any pixel, the model is considered to have predicted a no-target.

% \subsection{MLLMs (Gemini, GPT-4o and Qwen2-VL)}
\noindent\textbf{MLLMs (Gemini, GPT-4o and Qwen2-VL)}.
We conducted zero-shot evaluations under bounding box settings.
The bounding box-based outputs are commonly recommended for object detection tasks for representative MLLMs, such as Gemini, GPT-4o, and Qwen2-VL.
We conducted five experiments by varying the temperature parameter.
To predict the existence of a target region and its corresponding bounding box, we used the following prompt: 
\textit{``Given an image and a movement instruction, analyze if there is a target region for the movement. If it exists, identify the region by specifying a bounding box with two points in 2D coordinates. The top-left and bottom-right corners of the box that surrounds the target area. If there are multiple options, list them all. Return the response in the following JSON format: \{``has\_target'': boolean, ``bbox'': [[[x1, y1], [x2, y2]], ... [[x1, y1], [x2, y2]]] or null\}. The coordinates should be in pixels relative to the top-left corner of the image. If there is no target region, set `bbox' to null''.}
To select the above prompt, we conducted preliminary experiments using over ten prompts and selected the one yielding the best results.
% \noindent\textbf{Polygon-based method (Qwen2-VL)}.
To compare performance under settings similar to GENNAV, we also conducted an experiment with polygon-based method using Qwen2-VL.
We used the following prompt:
\textit{``Given an image and a movement instruction, analyze if there is a target region for the movement. If it exists, identify the region by specifying a 6 points polygon with 12 points in 2D coordinates. If there are multiple options, list them all. Return the response in the following JSON format: `\{``has\_target'': boolean, ``polygon'': [[[x1, y1], [x2, y2], [x3, y3], [x4, y4], [x5, y5], [x6, y6]], ... [[x1, y1], [x2, y2], [x3, y3], [x4, y4], [x5, y5], [x6, y6]]] or null\}'. The coordinates should be in pixels relative to the top-left corner of the image. If there is no target region, set `polygon' to null.''}

\vspace{-2mm}
\section{Evaluation Metrics Details}
\vspace{-4mm}
We employed $\mathrm{P}@K$ because it is a standard metric for GRNR and RNR tasks.
$\mathrm{P}@K$ counts the percentage of samples with IoU higher than the threshold $K$. $\mathrm{P}@K$ is defined as follows:
\begin{align}
\begin{split}
\mathrm{P}@K &= \frac{N_K}{N}, \\
N_K &= \sum_{i=1}^{N} \mathbbm{1} \left[\text{IoU}(\hat{y}_i, y_i) > K\right].
\end{split}
\end{align}
We set the threshold to 0.1 and 0.2 for $\mathrm{P}@K$.
This is because setting the IoU threshold to 0.1 and 0.2 is considered appropriate in the GRNR task where human agreement substantially deviates from 1.0.
As shown in Table~1, even at low thresholds such as $\mathrm{P}@0.1$ and $\mathrm{P}@0.2$, human performance was only 56.00\% and 36.40\%, respectively. 
This indicates that even humans struggle to predict consistent target regions.
Moreover, we observe that even relatively simple models, when trained on data from a specific (in-domain) distribution, can sometimes surpass human performance. 
Such cases raise concerns about whether $\mathrm{P}@K$ remains a valid and informative metric, particularly when model performance exceeds that of humans.
In fact, the baseline methods already outperform human performance at $\mathrm{P}@K$ with $K$ $\geqq$ 0.3.
Given these observations, we consider that $\mathrm{P}@K$ at thresholds higher than 0.3 should be excluded from our evaluation because it no longer reflects meaningful or reliable distinctions in model capability.

We also evaluated the accuracy of predicting the existence of target regions.
Accuracy is defined as follows:
\begin{align}
\mathrm{Accuracy} = \frac{\mathrm{TP} + \mathrm{TN}}{\mathrm{TP} + \mathrm{TN} + \mathrm{FP} + \mathrm{FN}}.
\end{align}

\vspace{-2mm}
\section{Additional Results}
\vspace{-3mm}
\subsection{Additional Quantitative Results}
% 6-15
Table 1 shows the quantitative results of the baseline methods, GENNAV and the human performance on the GRiN-Drive benchmark. 
The values in the table are the average and standard deviation over five trials.
The ``Type'' and ``Inf. Speed'' columns specify the segmentation approach and inference speed of each method, respectively.

The msIoU of GENNAV and the baseline methods are as follows, grouped by input resolution.
In the 224×224 setting, the method by Rufus et al., LAVT, and TNRSM achieved msIoU scores of 35.44, 31.73, and 37.90, respectively.
Under the 640×640 condition, GENNAV achieved an msIoU of 46.35, whereas the method by Rufus et al., LAVT, and TNRSM resulted in scores of 37.84, 34.09, and 22.84, respectively.
In the high-resolution (1024×1024 and 1600×900) setting, GSVA and the MLLM baselines Gemini, GPT-4o, Qwen2-VL (bbox), and Qwen2-VL (polygon) yielded msIoU scores of 32.94, 6.98, 23.41, 24.06, and 12.16, respectively.
These results show that GENNAV improved by 8.45 points over TNRSM (224×224), which achieved the highest msIoU among the methods.

Similarly, the $\mathrm{P}@0.1$ scores of the baseline methods by Rufus et al., LAVT, and TNRSM in the 224×224 setting were 21.53, 30.28, and 44.05, respectively.
 At the 640×640 resolution, the $\mathrm{P}@0.1$ scores of GENNAV and the baseline methods by Rufus et al., LAVT, and TNRSM were 11.48, 7.90, and 38.57, respectively.
Moreover, at the 1024×1024 resolution, the $\mathrm{P}@0.1$ score of GSVA-Vicuna-7b was 17.26.
 Furthermore, the MLLM baselines, that is, Gemini, GPT-4o, Qwen2-VL (bbox), and Qwen2-VL (polygon), achieved $\mathrm{P}@0.1$ scores of 6.92, 5.04, 3.85, and 0.08, respectively.
These results demonstrate that GENNAV outperformed the baseline methods in terms of $\mathrm{P}@0.1$, demonstrating a 5.55 point improvement over the best-performing baseline, TNRSM (224×224).
Moreover, GENNAV achieved an accuracy of 75.41\%, surpassing the baseline methods, including the MLLMs, with a 7.52 point improvement over the best-performing baseline, TNRSM (224×224).

Accordingly, the results can be summarized as follows:
GENNAV outperformed the baseline methods in terms of msIoU, $\mathrm{P}@0.1$, $\mathrm{P}@0.2$ and accuracy. Moreover, the improvement achieved by GENNAV in terms of msIoU, $\mathrm{P}@0.1$, and accuracy was statistically significant ($p < 0.05$).
 Using high-resolution images with the baseline methods (the method by Rufus et al., LAVT, and TNRSM) did not lead to improved performance in terms of msIoU, $\mathrm{P}@K$, and accuracy. 
This limitation arose mainly because these methods used backbones that were fine-tuned on a resolution of 224×224, such as the Swin Transformer~\citep{liu2021swin}. 
Consequently, they were not well-suited for processing high-resolution inputs, which hindered potential performance gains.
Furthermore, the use of high-resolution images substantially increased the computational cost, which resulted in longer inference speeds across all methods compared to inputs with a resolution of 224×224. 

Finally, we compared the inference speeds of all methods.
The inference speed per sample for GENNAV was 31.31 ms.  
For comparison, the inference speeds of the baseline methods were as follows. In the 224×224 setting, the method by Rufus et al., LAVT and TNRSM took 39.87 ms, 9.29 ms and 502.69 ms, respectively.
Under the 640×640 resolution setting, the method by Rufus et al., LAVT, and TNRSM took 65.00 ms, 11.51 ms, and 503.68 ms, respectively.
By contrast, MLLM-based methods exhibited significantly slower inference speeds: Gemini, GPT-4o, Qwen2-VL (bbox), and Qwen2-VL (polygon) took 1793.68 ms, 3525.68 ms, 1768.99 ms, and 1771.41 ms, respectively.  
Overall, GENNAV achieved faster inference speed than the baseline methods except LAVT.  
Although GENNAV was slower than LAVT, it achieved substantially higher performance on other metrics.

\begin{table}[t]
    \centering
    \setlength{\tabcolsep}{10pt}
    \renewcommand{\arraystretch}{1.5}
    \vspace{-5mm}
    \caption{
    Confusion matrix of target region existence for our method on the GRIN-Drive benchmark.}
    \begin{tabular}{|c|c|>{\centering\arraybackslash}p{3cm}|>{\centering\arraybackslash}p{3cm}|}
        \hline
        \multicolumn{2}{|c|}{} & \multicolumn{2}{c|}{Predicted target region existence} \\
        \cline{3-4}
        \multicolumn{2}{|c|}{} & Positive & Negative \\
        \hline
        \multirow{2}{*}{\shortstack{Ground truth  \\target region existence}} 
        & Positive & 359 & 143 \\
        \cline{2-4}
        & Negative & 81 & 175 \\
        \hline
    \end{tabular}
    \vspace{-3mm}
    \label{tab:confusion-matrix}
\end{table}

We conducted a subject experiment to evaluate human performance on the GRiN-Drive test set.
In total, we recruited 11 participants (aged 21-24) who all had technical backgrounds.
First, we asked the participants  to determine whether any target region specified by the navigation instruction was present in the image.
If one or more target regions were identified by the annotators, they provided these regions by drawing polygons. 
The human performance in terms of  msIoU, $\mathrm{P}@0.1$, $\mathrm{P}@0.2$, and accuracy were 56.08, 56.00, 36.40, and 88.00, respectively, as shown in Table 1.

Table \ref{tab:confusion-matrix} presents the confusion matrix for our method GENNAV using GRiN-Drive test set. 
There were 359, 175, 81, and 143 samples corresponding to TP, TN, FP, and FN, respectively.

\subsection{Additional Qualitative Results}
\begin{figure}[t]
    \centering
    \begin{minipage}{1.0\linewidth}
            \vspace{0pt}% to make [t] work properly
            \begin{minipage}{0.03\linewidth}
                (i)
            \end{minipage}
            \begin{minipage}{0.2365\linewidth}
                \includegraphics[width=\linewidth]{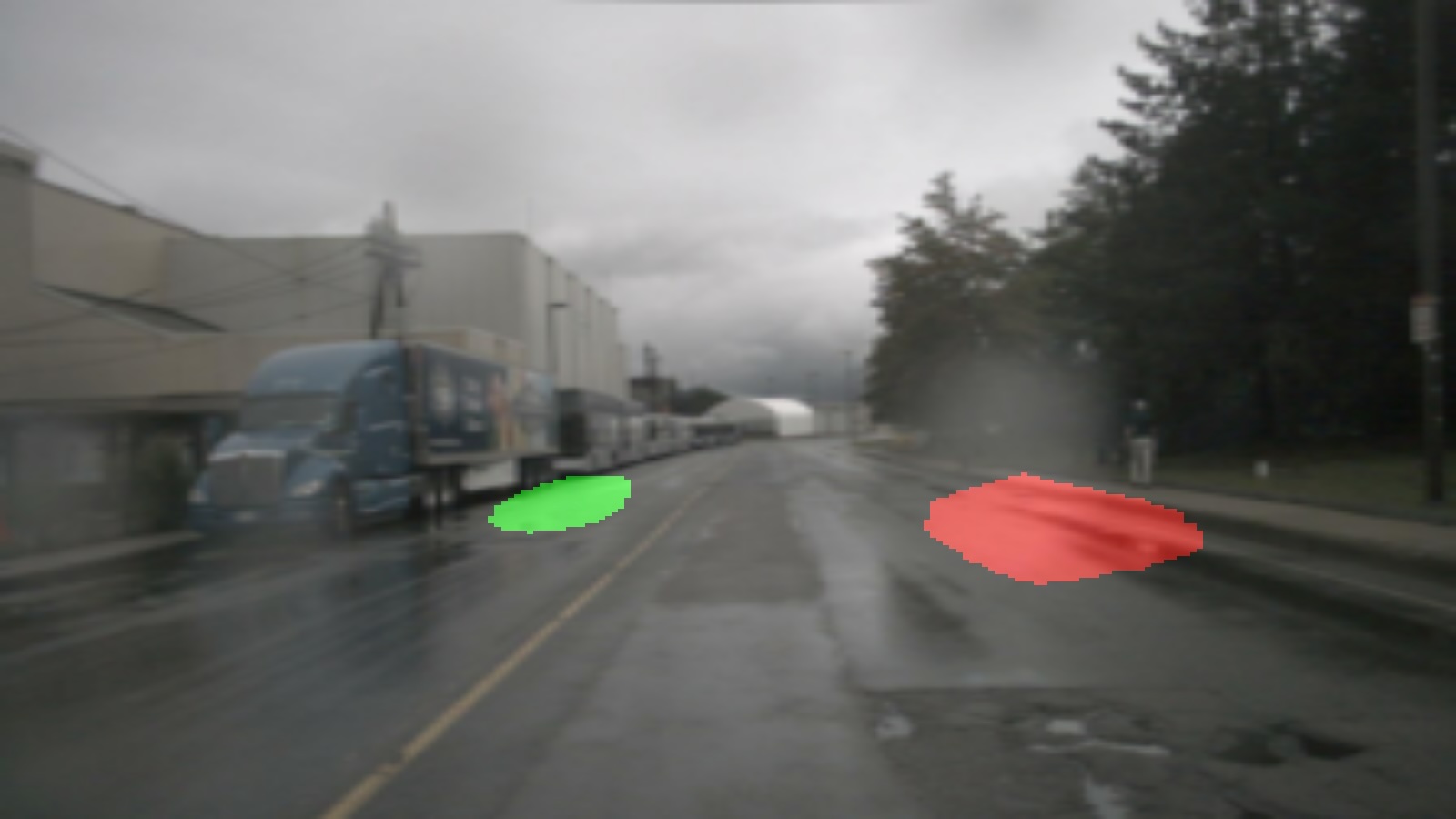}
            \end{minipage}
            \hfill
            \begin{minipage}{0.2365\linewidth}
                \includegraphics[width=\linewidth]{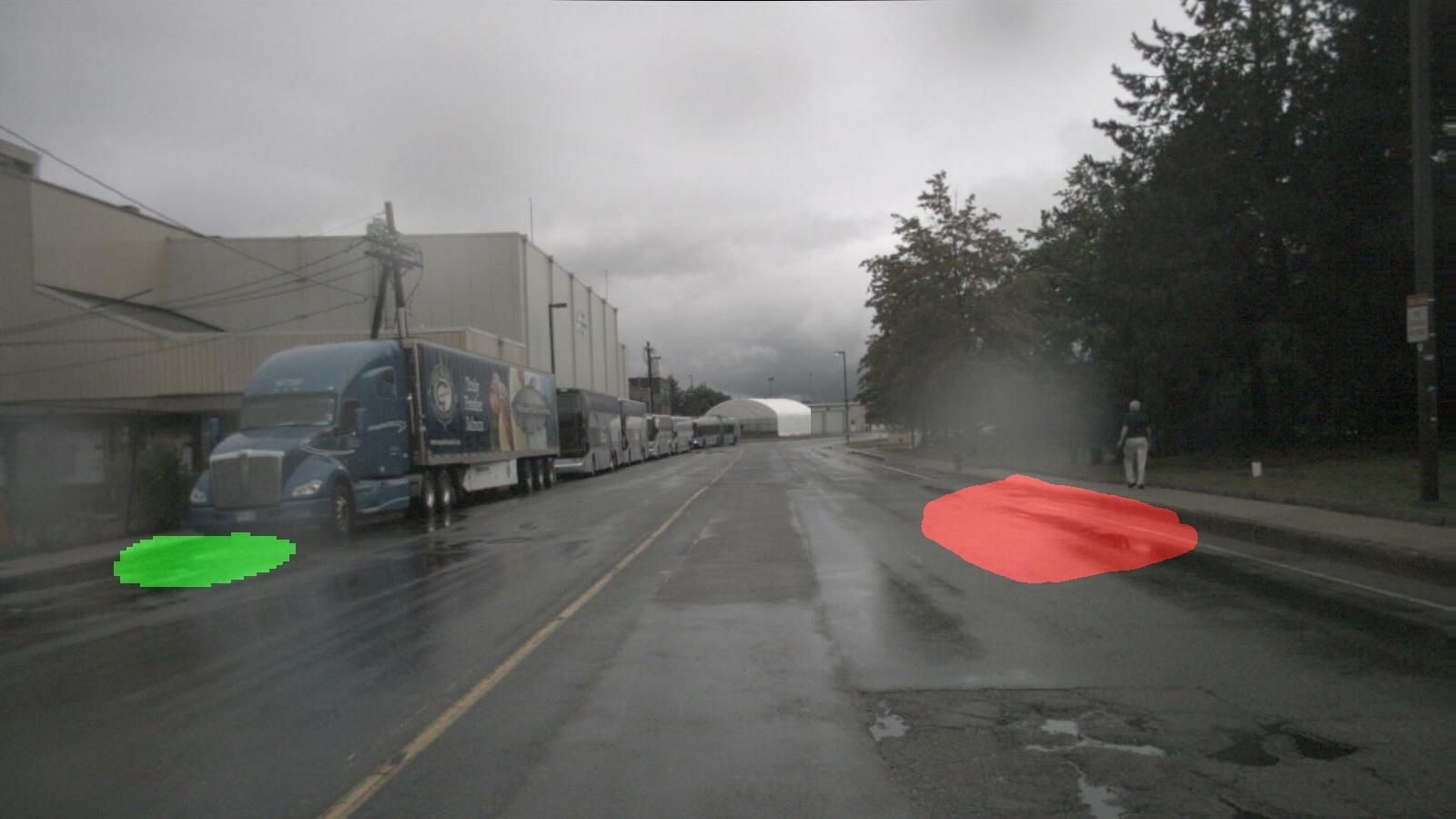}
            \end{minipage}
            \begin{minipage}{0.2365\linewidth}
                \includegraphics[width=\linewidth]{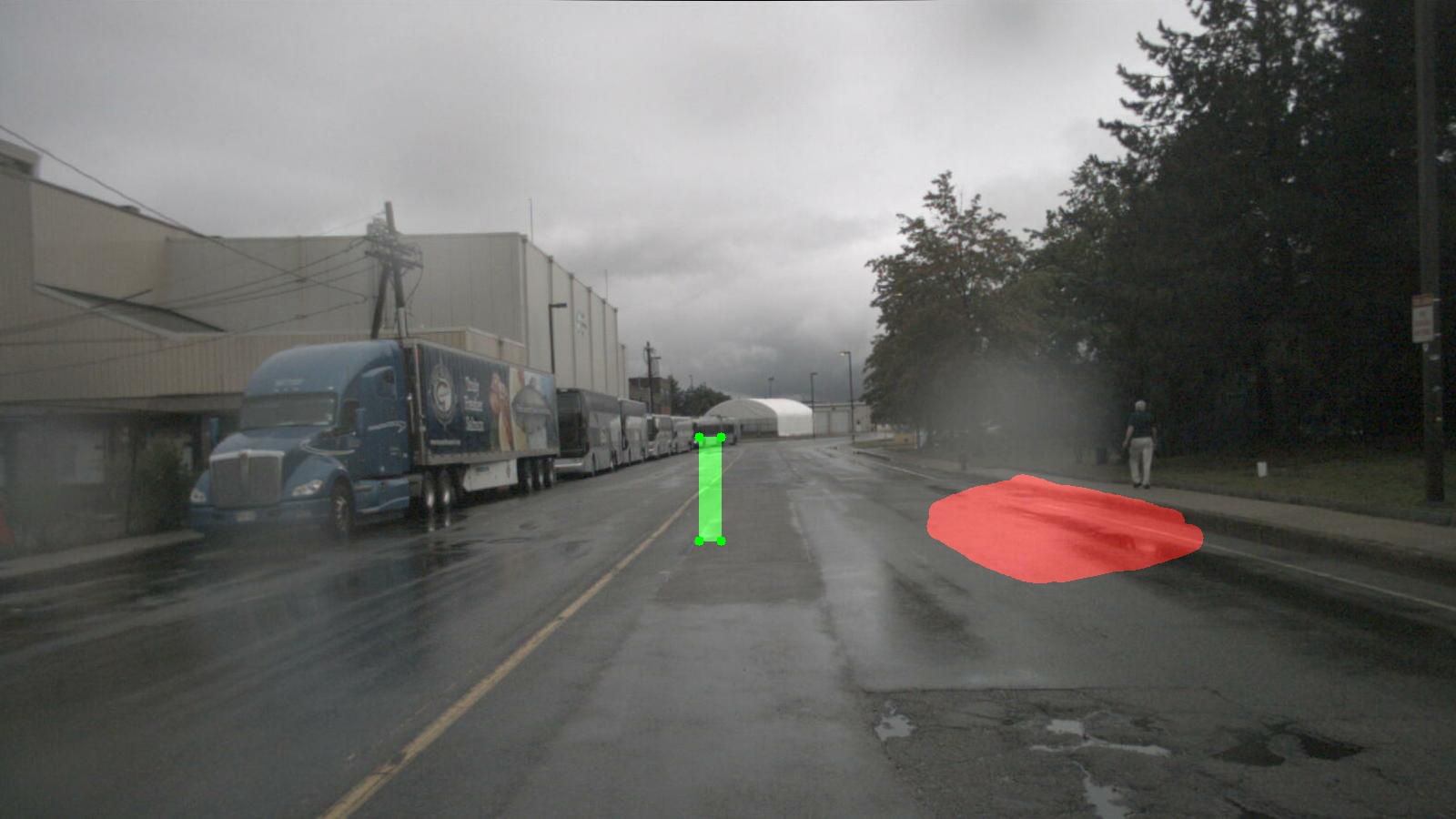}
            \end{minipage}
            \hfill
            \begin{minipage}{0.2365\linewidth}
                \includegraphics[width=\linewidth]{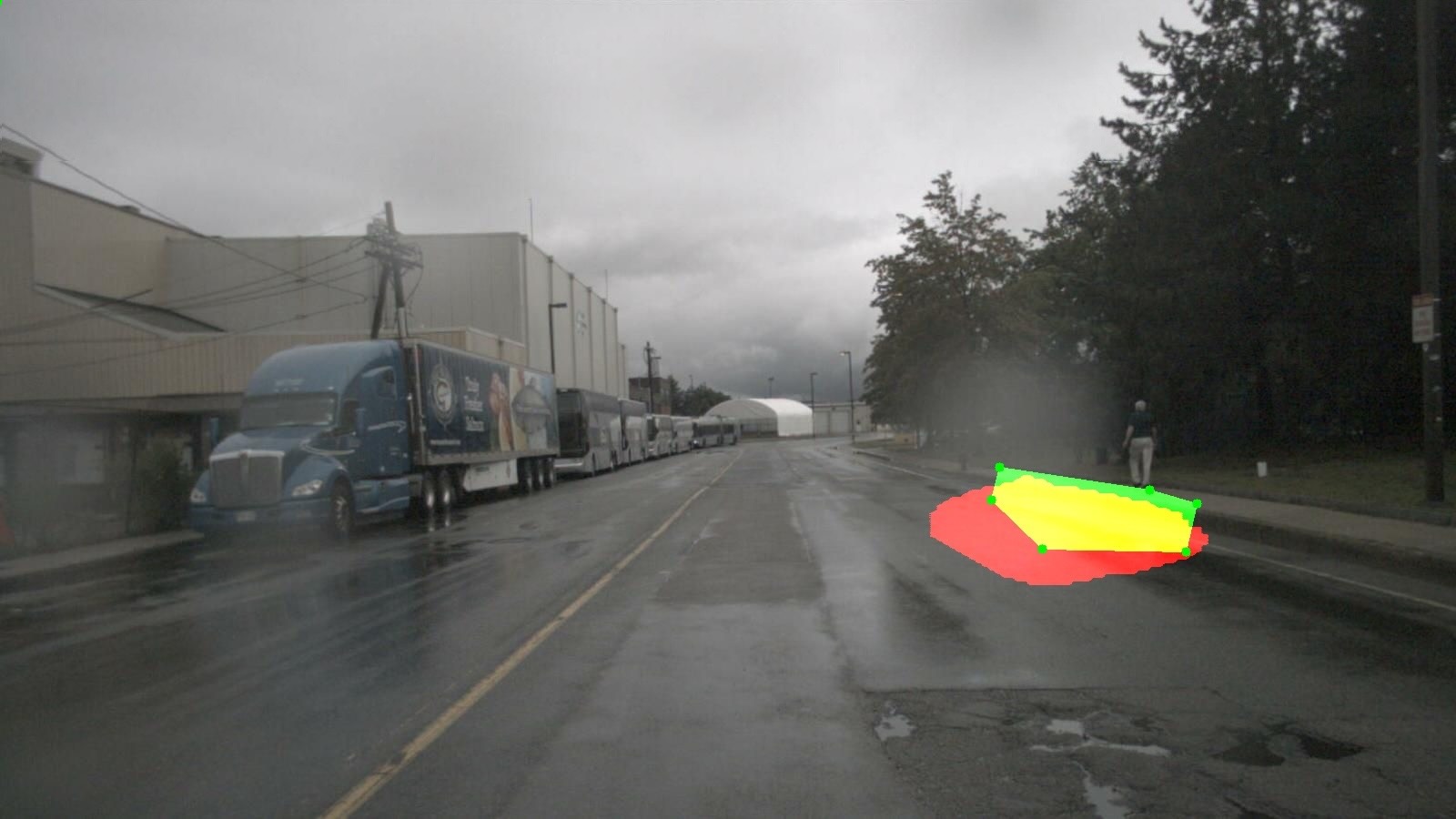}
            \end{minipage}    
            \begin{minipage}[t]{1.0\linewidth}
                \centering {
                $\bm{x}_\text{inst}$: ``Pull over where that man is.''}
            \end{minipage}
            % Row 2
        \end{minipage}
    \begin{minipage}{1.0\linewidth}
            \vspace{0pt}% to make [t] work properly
            % Row 1
            \begin{minipage}{0.03\linewidth}
                (ii)
            \end{minipage}
            \begin{minipage}{0.2365\linewidth}
                \includegraphics[width=\linewidth]{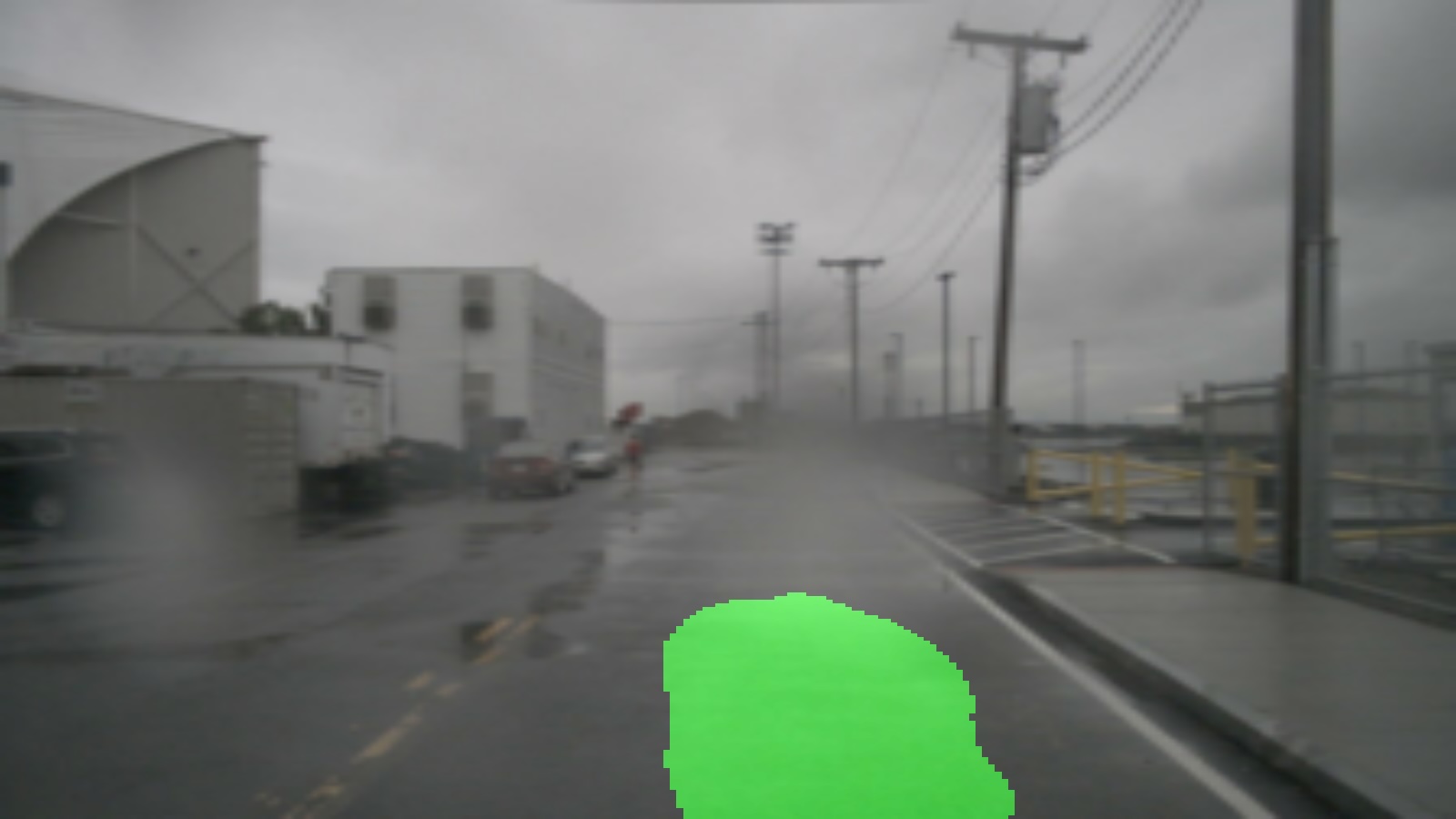}
            \end{minipage}
            \hfill
            \begin{minipage}{0.2365\linewidth}
                \includegraphics[width=\linewidth]{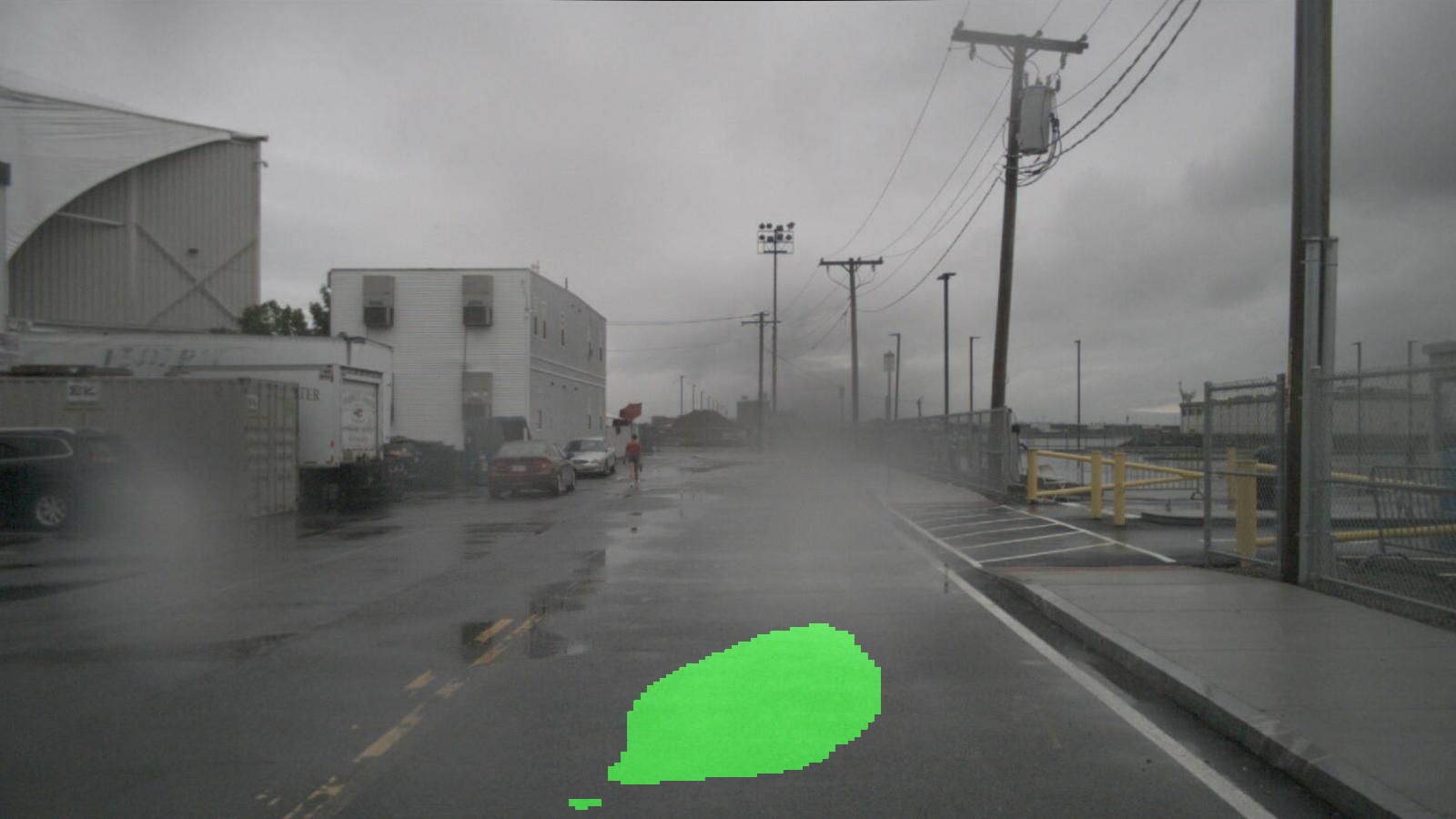}
            \end{minipage}
            \begin{minipage}{0.2365\linewidth}
                \includegraphics[width=\linewidth]{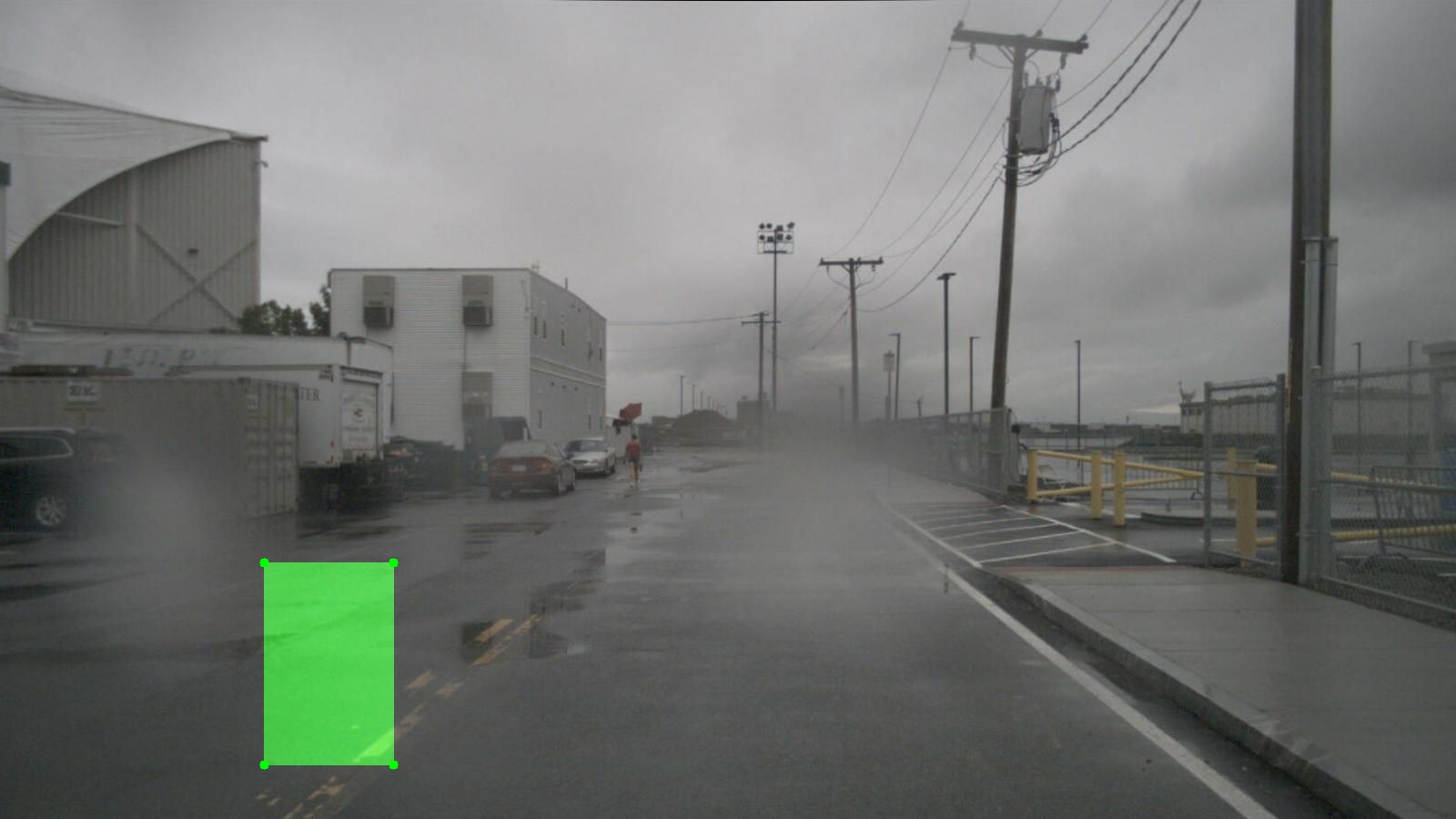}
            \end{minipage}
            \hfill
            \begin{minipage}{0.2365\linewidth}
                \includegraphics[width=\linewidth]{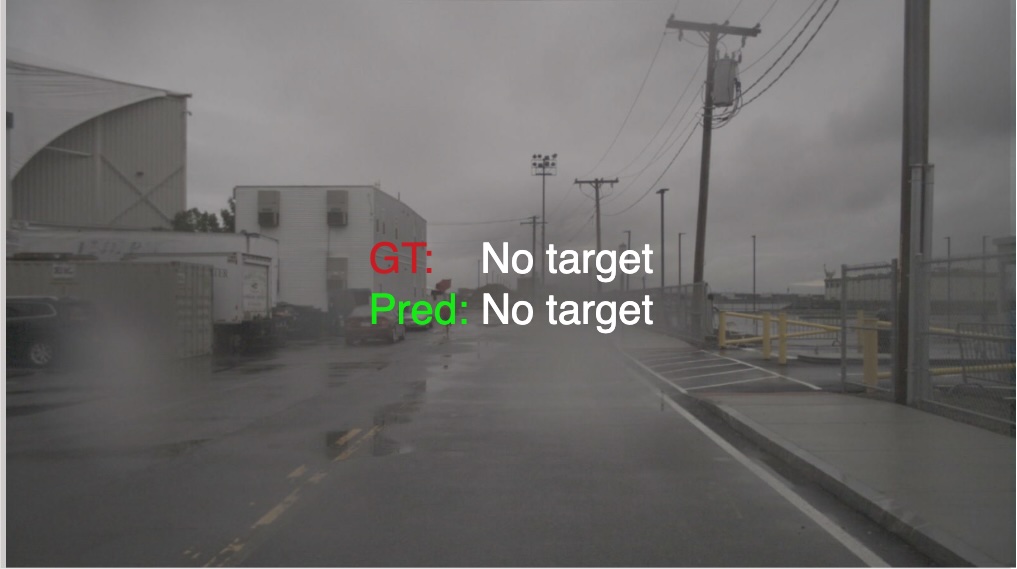}
            \end{minipage}    
            \begin{minipage}[t]{1.0\linewidth}
                \centering { $\bm{x}_\text{inst}$: ``Slow down and let that gold car pass in front of us.''}
            \end{minipage}
            % Row 2
        \end{minipage}
    \caption{
     Qualitative results of the proposed method and baseline methods on the GRiN-Drive benchmark. Columns (a), (b), (c), and (d) show the predictions by LAVT, TNRSM, Qwen2-VL (bbox) and GENNAV, respectively. The green and red regions indicate the predicted and ground-truth regions, respectively, while the yellow region represents the overlap between the predicted and ground-truth regions.
    }
    \label{fig:qualitative-success-supp}
\end{figure}

\begin{figure}[t]
    \centering
    \begin{minipage}{1.0\linewidth}
            \vspace{0pt}
            % Row 1
            \begin{minipage}{0.49\linewidth}
                \includegraphics[width=\linewidth]{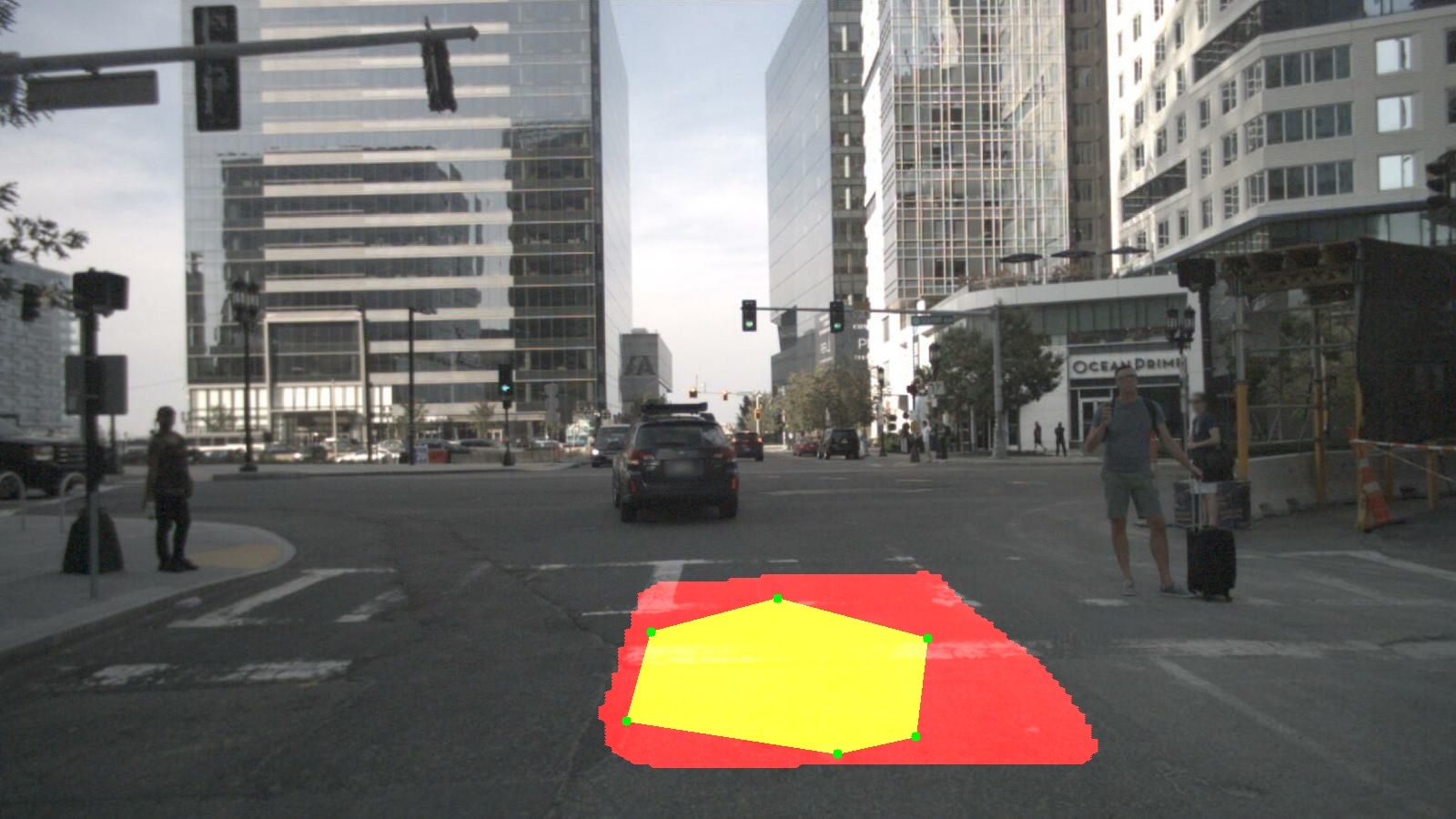}
            \end{minipage}
            \hfill
            \begin{minipage}{0.49\linewidth}
                \includegraphics[width=\linewidth]{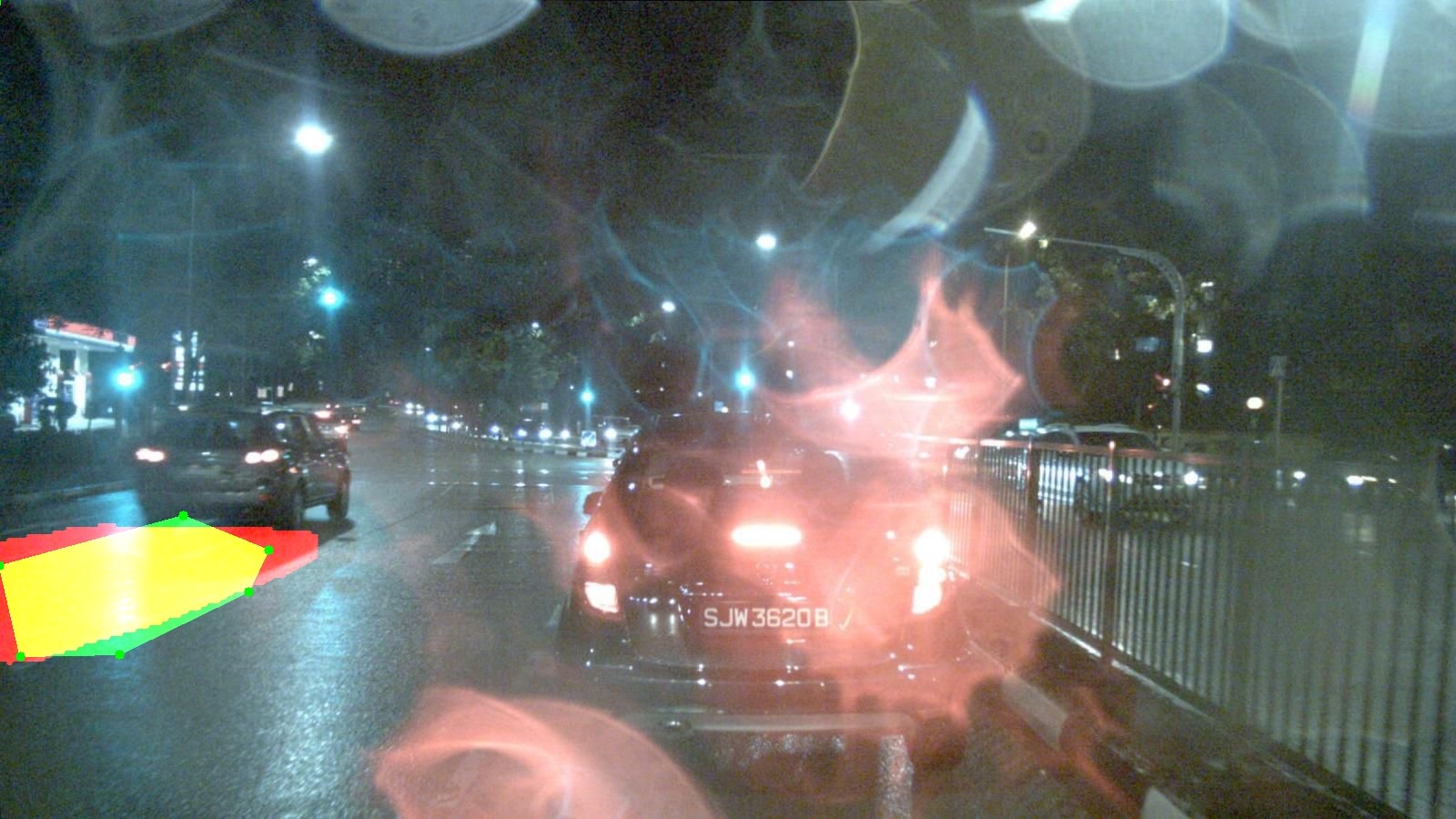}
            \end{minipage}
            \begin{minipage}[t]{0.5\linewidth}
                \vspace{3pt}
                \centering {(i)~$\bm{x}_\text{inst}$: ``Slow down near that person.''}
            \end{minipage}
            \begin{minipage}[t]{0.5\linewidth}
                \vspace{3pt}
                \centering {(ii)~$\bm{x}_\text{inst}$: ``Switch lanes and follow that gray car two lanes to the left.''}
            \end{minipage}
            % Row 2
             \begin{minipage}{0.49\linewidth}
                \includegraphics[width=\linewidth]{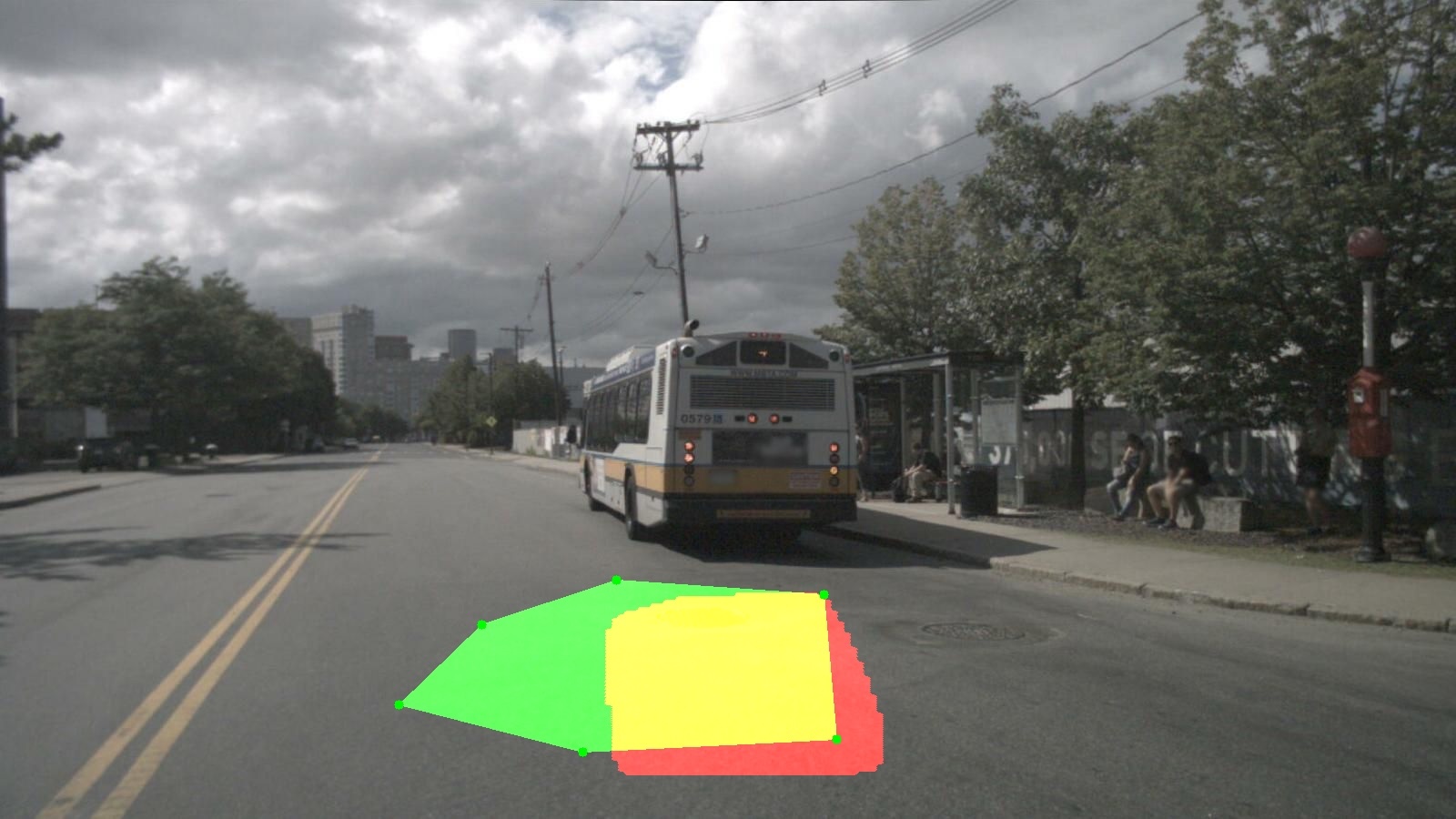}
            \end{minipage}
            \hfill
            \begin{minipage}{0.49\linewidth}
                \includegraphics[width=\linewidth]{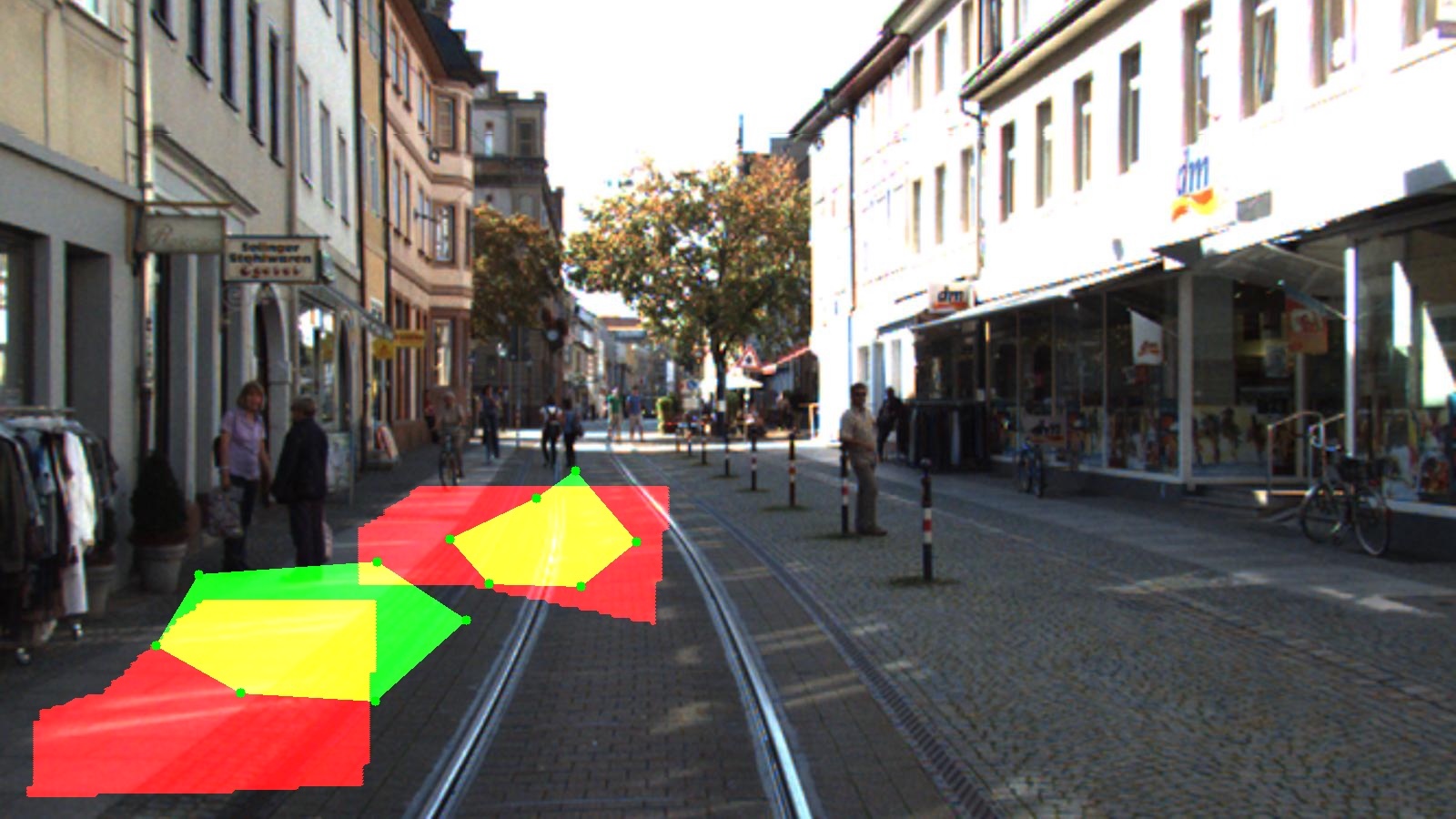}
            \end{minipage}
            \begin{minipage}[t]{0.5\linewidth}
                \vspace{3pt}
                \centering {(iii)~$\bm{x}_\text{inst}$: ``Slow down a bit, that bus might pull out into traffic.''}
            \end{minipage}
            \begin{minipage}[t]{0.5\linewidth}
                \vspace{3pt}
                \centering {(iv)~$\bm{x}_\text{inst}$: ``Stop near those to the left.''}
            \end{minipage}

             \begin{minipage}{0.49\linewidth}
                \includegraphics[width=\linewidth]{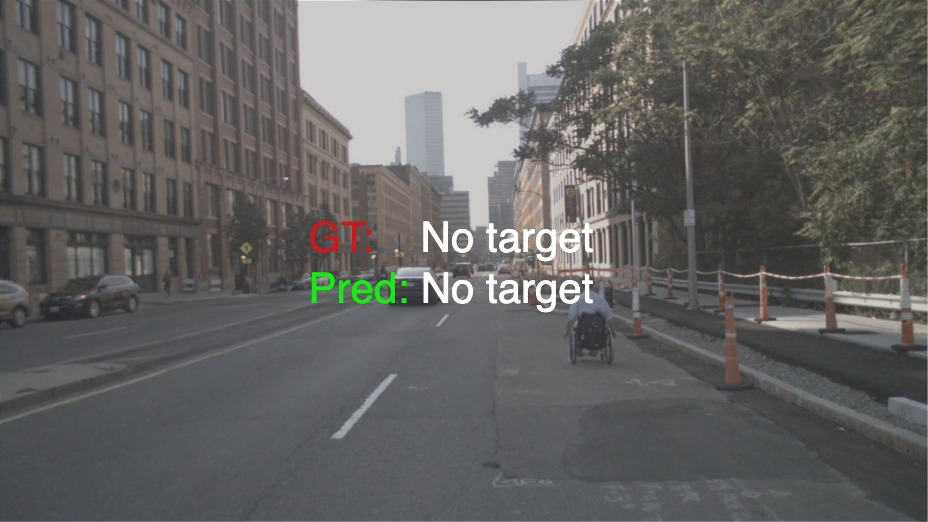}
            \end{minipage}
            \hfill
            \begin{minipage}{0.49\linewidth}
                \includegraphics[width=\linewidth]{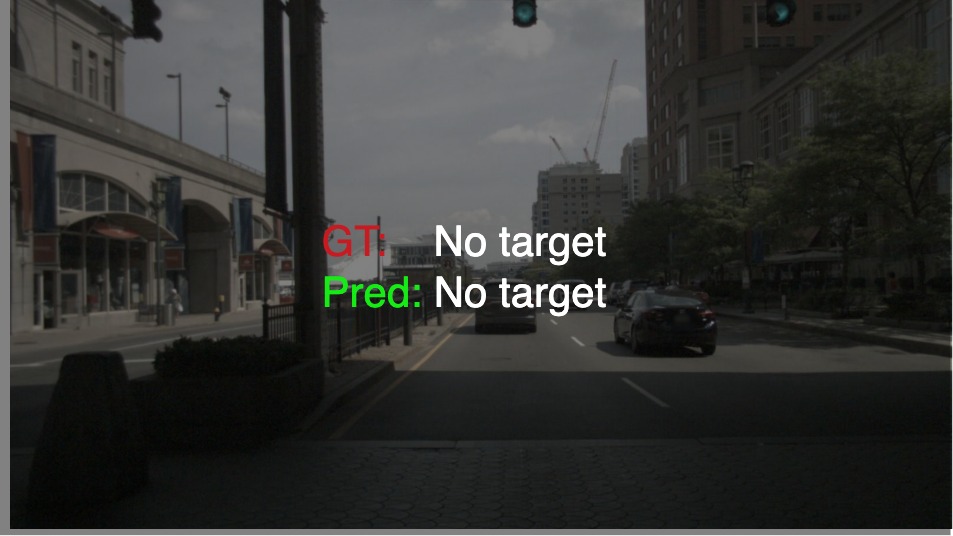}
            \end{minipage}
            \begin{minipage}[t]{0.5\linewidth}
                \vspace{3pt}
                \centering {(v)~$\bm{x}_\text{inst}$: ``Park behind that truck that is way up ahead.''}
            \end{minipage}
            \begin{minipage}[t]{0.5\linewidth}
                \vspace{3pt}
                \centering {(vi)~$\bm{x}_\text{inst}$: ``Pull up in front of that gate left of the green trash bin.''}
            \end{minipage}
            % Row 2
             \begin{minipage}{0.49\linewidth}
                \includegraphics[width=\linewidth]{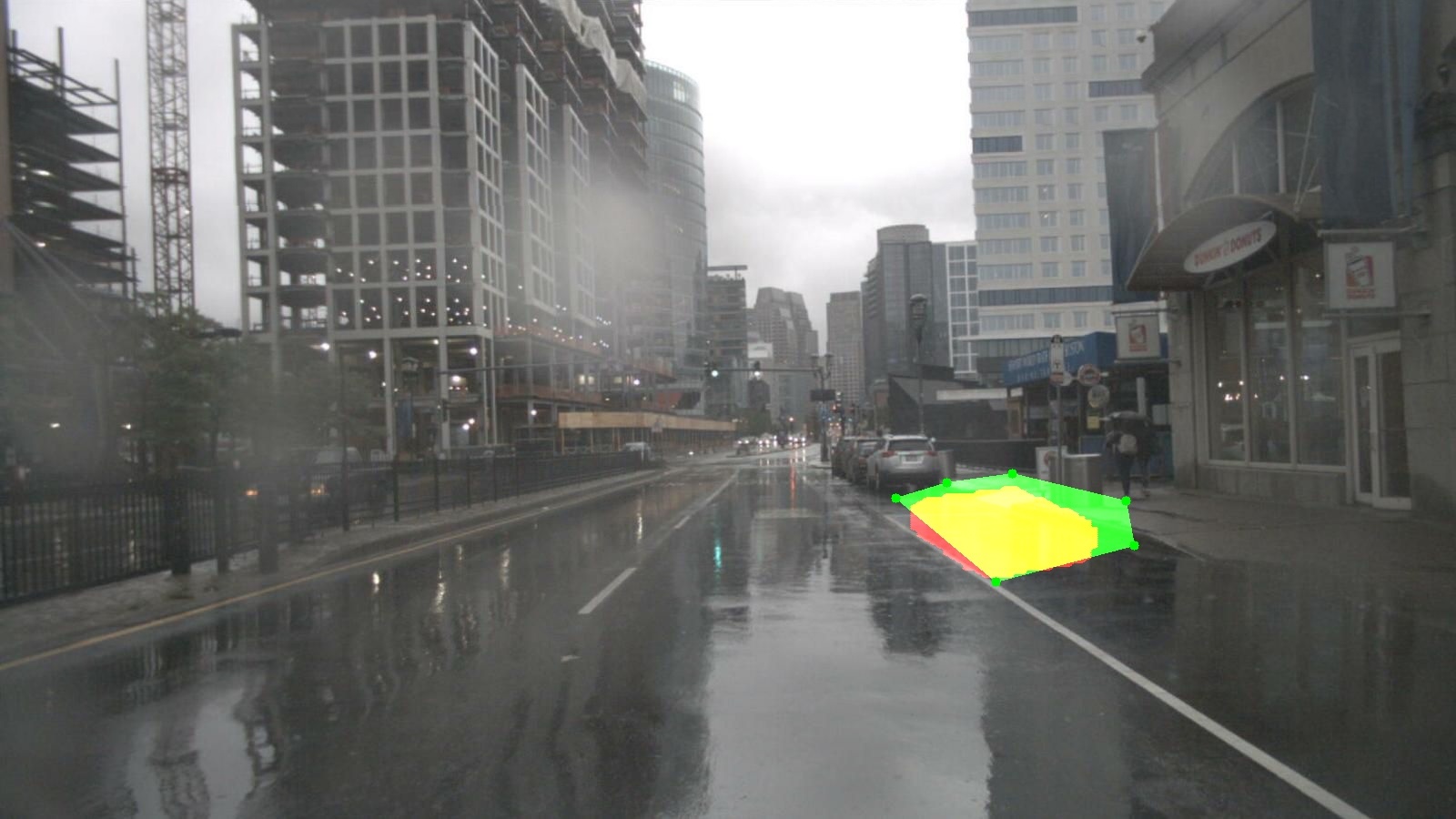}
            \end{minipage}
            \hfill
            \begin{minipage}{0.49\linewidth}
                \includegraphics[width=\linewidth]{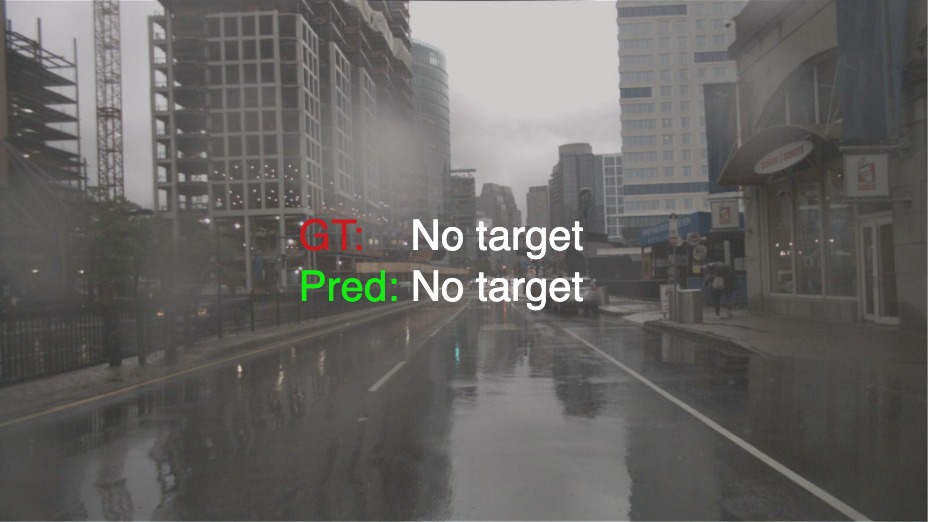}
            \end{minipage}
            \begin{minipage}[t]{0.5\linewidth}
                \vspace{3pt}
                \centering {(vii)~$\bm{x}_\text{inst}$: ``Pull up to that girl on the right closer to the road.''}
            \end{minipage}
            \begin{minipage}[t]{0.5\linewidth}
                \vspace{3pt}
                \centering {(viii)~$\bm{x}_\text{inst}$: ``Park behind that truck on the right side of the road.''}
            \end{minipage}
            % Row 2
        \end{minipage}

    \caption{
      Qualitative results of the proposed method in the GRiN-Drive benchmark. The green and red regions indicate the predicted and ground-truth regions, respectively, while the yellow region represents the overlap between the predicted and ground-truth regions.
    }
    \vspace{-4mm}
    \label{fig:supp-qualitative}
\end{figure}

Figs.~\ref{fig:supp-qualitative} (i) illustrates an example, where the $\bm{x}_\text{inst}$ is, ``Pull over where that man is.’'
In this example, the model is expected to generate masks on the road in front of the man on the right side of the image.
The baseline methods LAVT and TNRSM generated the inappropriate masks on regions around the truck and Qwen2-VL (bbox) inappropriately predicted the central region of the road.
By contrast, our method was able to generate an appropriate mask on the region next to the man. 
Figs.~\ref{fig:supp-qualitative} (ii) demonstrates a no-target case with $\bm{x}_\text{inst}$: ``Slow down and let that gold car pass in front of us.''
The appropriate prediction should be a ``no-target'', because the gold car does not exist in the scene.
GENNAV successfully predicted this sample as a ``no-target.'' 
The baseline methods were unable to appropriately predict the absence of landmarks, which resulted in mask generation in inappropriate regions of the road

\begin{figure}[t]
    \centering
    \begin{minipage}{1.0\linewidth}
            \vspace{0pt}
            \begin{minipage}{0.244\linewidth}
                \includegraphics[width=\linewidth]{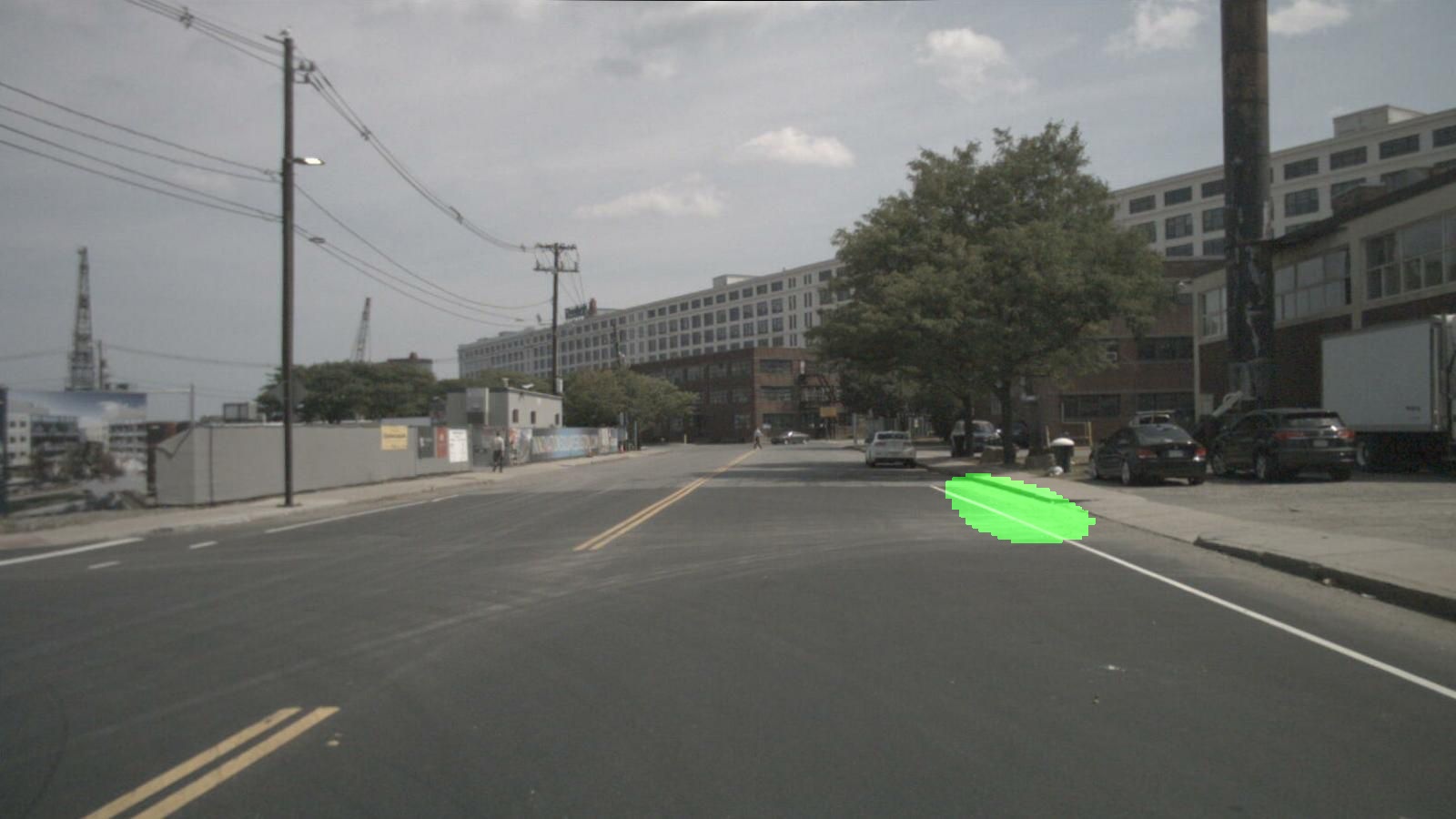}
                \caption*{(a)}
            \end{minipage}
            \hfill
            \begin{minipage}{0.244\linewidth}
                \includegraphics[width=\linewidth]{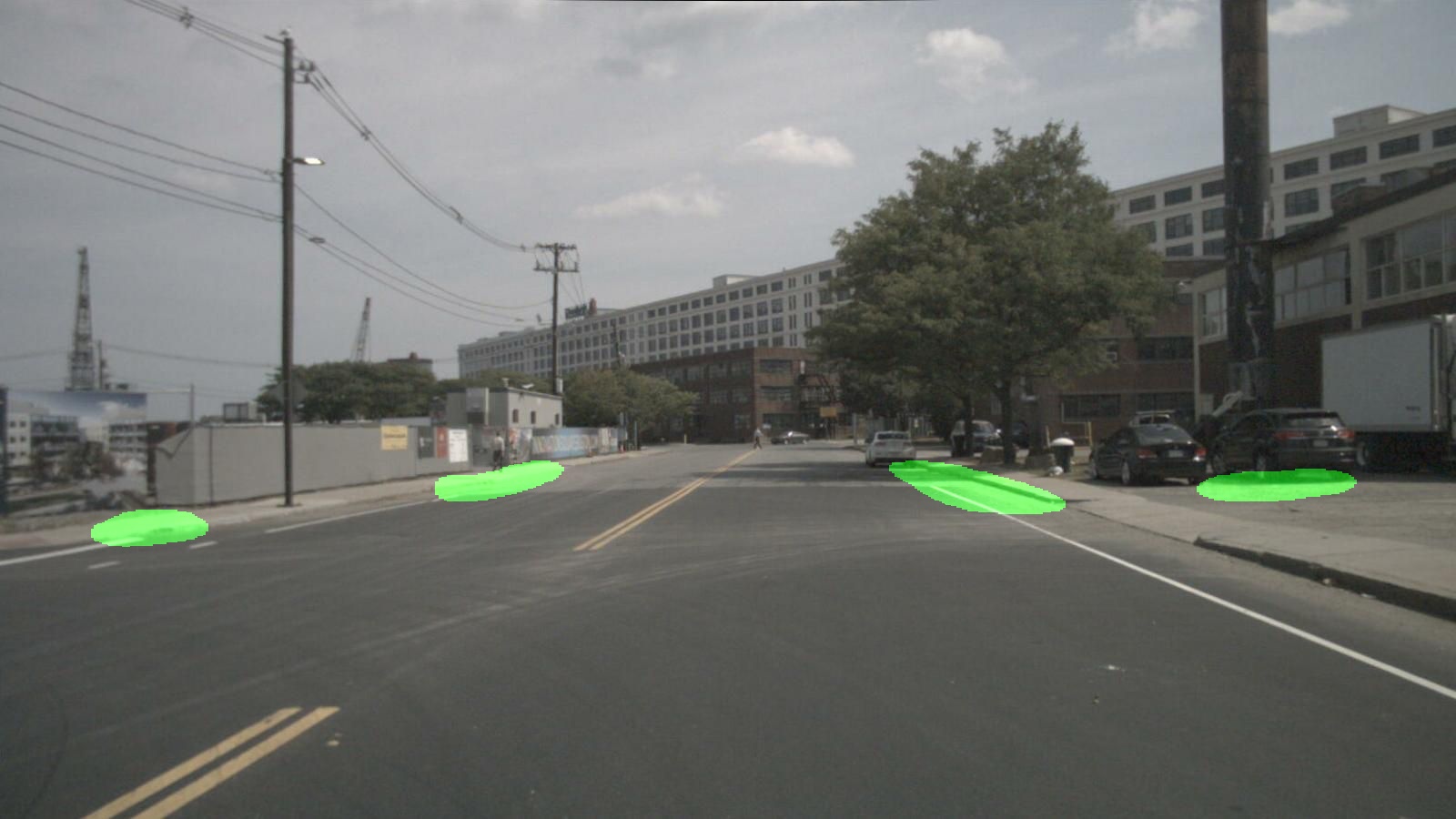}
                \caption*{(b)}
            \end{minipage}
            \begin{minipage}{0.244\linewidth}
                \includegraphics[width=\linewidth]{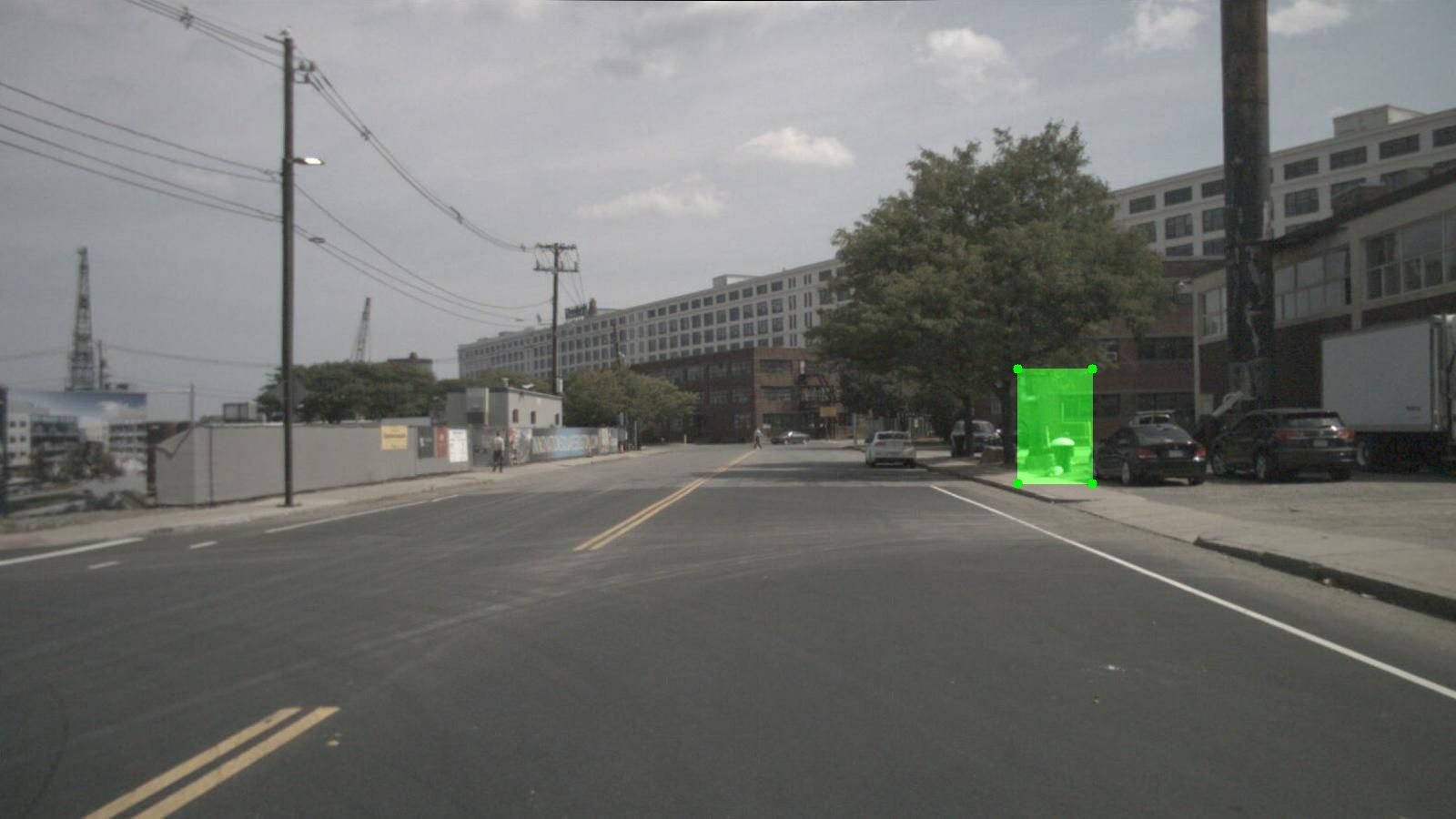}
                \caption*{(c)}
            \end{minipage}
            \hfill
            \begin{minipage}{0.244\linewidth}
                \includegraphics[width=\linewidth]{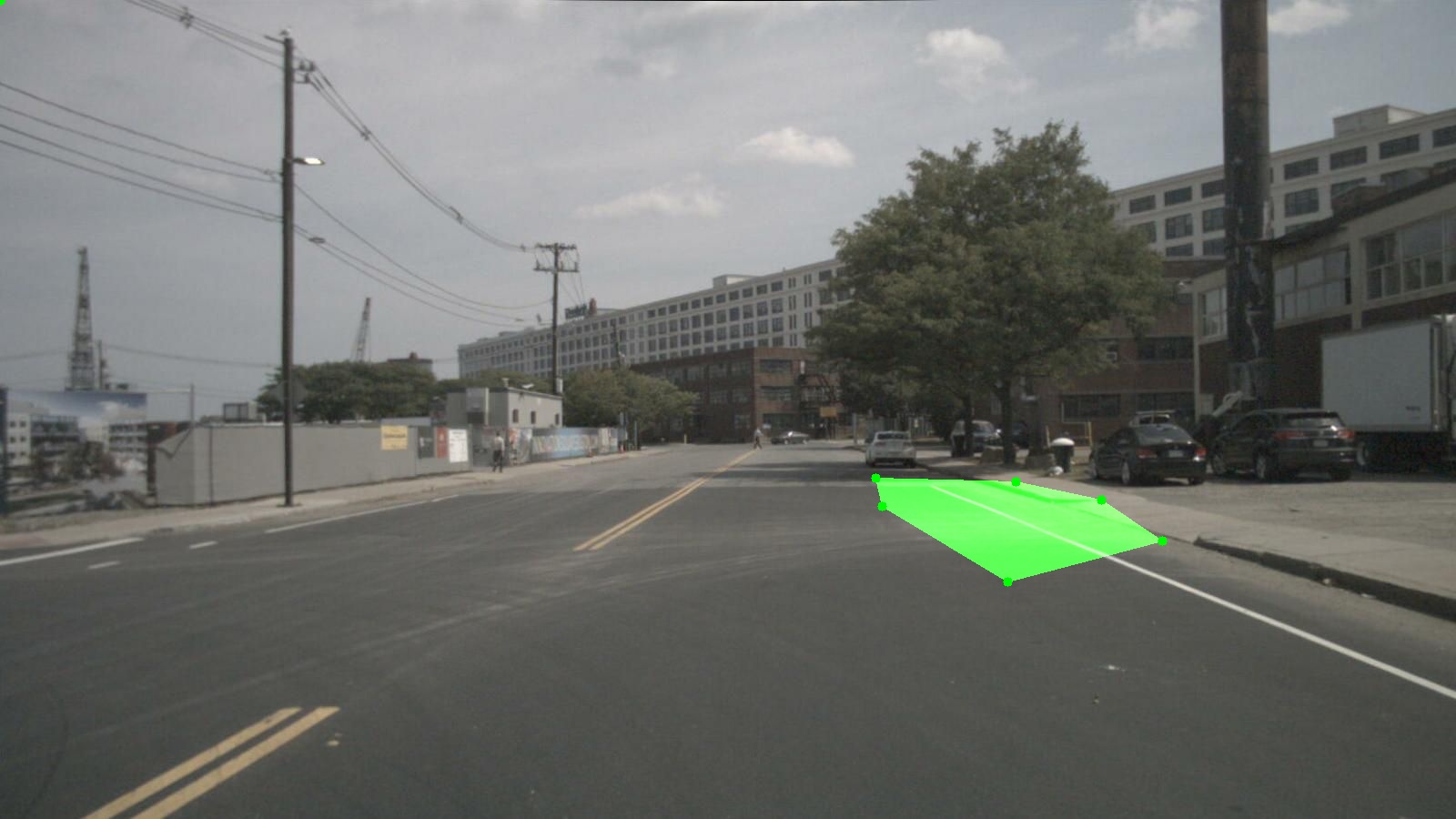}
                \caption*{(d)}
            \end{minipage}    
            \begin{minipage}[t]{1.0\linewidth}
                \centering { $\bm{x}_\text{inst}$: ``Pull up next to the pedestrian on our right close to the tree.''}
            \end{minipage}
            % Row 2
        \end{minipage}
    \caption{
     Qualitative results of a failed case. Columns (a), (b), (c), and (d) show the predictions made by LAVT~\citep{yang2022lavt}, TNRSM~\citep{tnrsmral24}, Qwen2-VL~\citep{Qwen2-VL} (bbox), and GENNAV, respectively. The green regions indicate the predicted segmentation.
    }
    \label{fig:qualitative-fail}
\end{figure}

% 6-9
 Fig. \ref{fig:qualitative-fail} shows the qualitative result of a failed case. In the figure, Columns (a), (b), (c), and (d)  show the prediction made by LAVT, TNRSM, Qwen2-VL (bbox) and GENNAV, respectively.
The green regions indicate the predicted segmentation.
Fig.~\ref{fig:qualitative-fail} presents a failed example in a no-target sample. 
In this example, the instruction was ``Pull up next to the pedestrian on our right close to the tree.'' 
All methods including GENNAV predicted the region around the tree.
However, because there are no pedestrians around the tree, predicting a ``no-target'' is appropriate.
GENNAV was influenced by the word ``tree,'' thereby leading to the incorrect prediction of the surrounding region.

Fig.~\ref{fig:supp-qualitative} provide additional success examples of GENNAV on the GRiN-Drive benchmark. 
As shown in Fig.~\ref{fig:supp-qualitative}~(vii) and Fig.~\ref{fig:supp-qualitative}~(viii), GENNAV is capable of generating different predictions for the same image when provided with different instructions.

\clearpage
% \begin{figure}[t]
%     \centering
%     \begin{minipage}{1.0\linewidth}
%             \vspace{0pt}% to make [t] work properly
%             % Row 1
%             \begin{minipage}{0.47\linewidth}
%                 \includegraphics[width=\linewidth]{fig/6-13/rvc-1.jpg}
%                 \caption*{(a)}
%             \end{minipage}
%             \hfill
%             \begin{minipage}{0.47\linewidth}
%                 \includegraphics[width=\linewidth]{fig/6-13/rvc-2.jpg}
%                 \caption*{(b)}
%             \end{minipage}
%             \begin{minipage}[t]{0.5\linewidth}
%                 \vspace{3pt}
%                 \centering { $\bm{x}_\text{inst}$: ``Park on the grass in front of the traffic cone.''}
%             \end{minipage}
%             \begin{minipage}[t]{0.5\linewidth}
%                 \vspace{3pt}
%                 \centering { $\bm{x}_\text{inst}$: ``Stay behind that car in front of us.''}
%             \end{minipage}
%             % Row 2
%         \end{minipage}
%     \caption{
%      Qualitative analysis of failure cases of the proposed method categorized as reduced visibility conditions. Regarding limited or poor visibility conditions, such as those caused by inadequate lighting, reflections from rain, or cloudy weather, errors may arise.
%     }
%     \label{fig:RVC}
%     \vspace{-5mm}
% \end{figure}

\begin{figure}[t]
    \centering
    \begin{minipage}{1.0\linewidth}
            \vspace{0pt}% to make [t] work properly
            % Row 1
            \begin{minipage}{0.47\linewidth}
                \includegraphics[width=\linewidth]{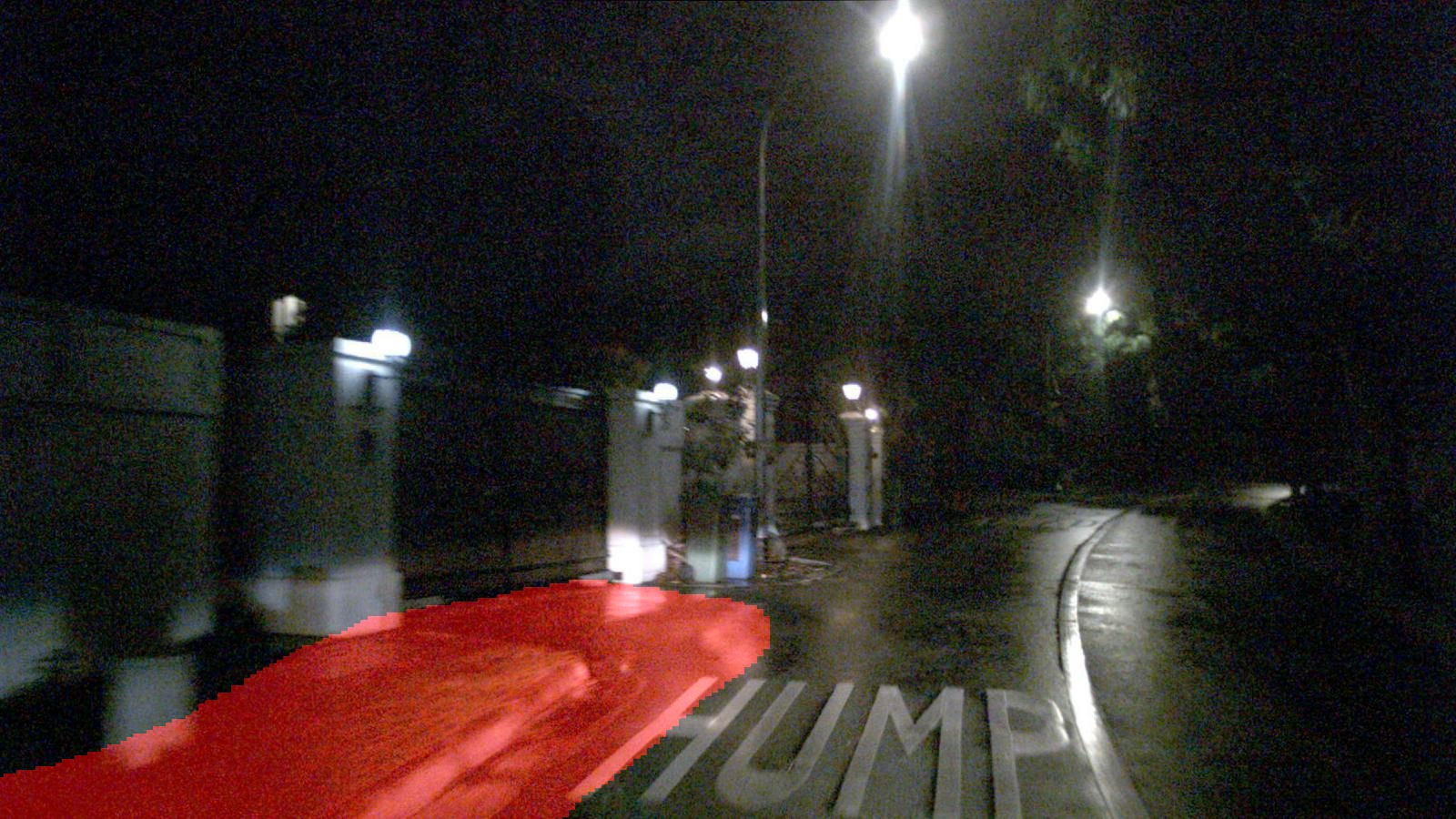}
                \caption*{(a)}
            \end{minipage}
            \hfill
            \begin{minipage}{0.47\linewidth}
                \includegraphics[width=\linewidth]{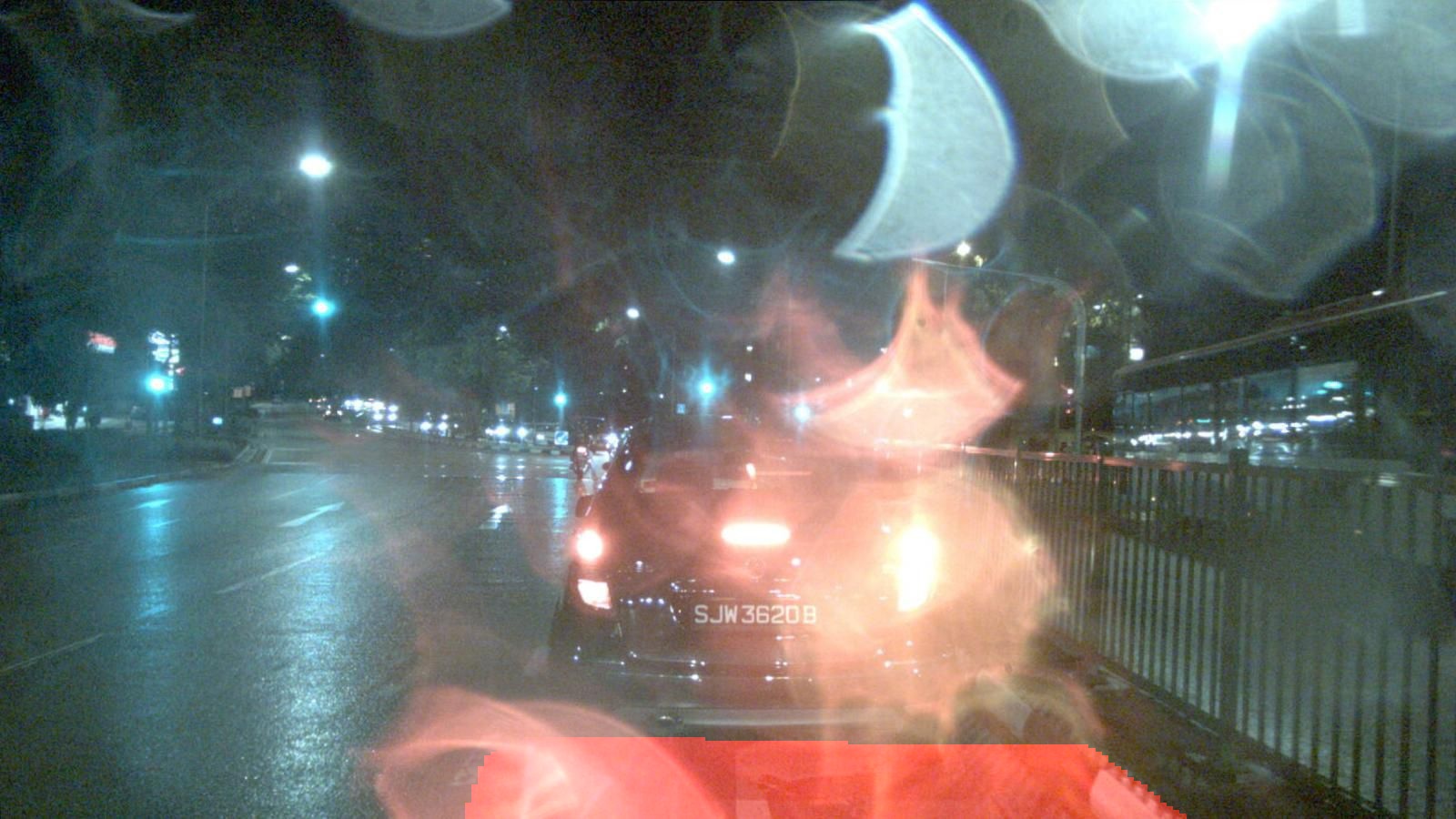}
                \caption*{(b)}
            \end{minipage}
            \begin{minipage}[t]{0.5\linewidth}
                \vspace{3pt}
                \centering { $\bm{x}_\text{inst}$: ``Park on the grass in front of the traffic cone.''}
            \end{minipage}
            \begin{minipage}[t]{0.5\linewidth}
                \vspace{3pt}
                \centering { $\bm{x}_\text{inst}$: ``Stay behind that car in front of us.''}
            \end{minipage}
    \end{minipage}
    \caption{
     Qualitative analysis of failure cases of the proposed method categorized as Reduced Visibility Conditions. Regarding limited or poor visibility conditions, such as those caused by inadequate lighting, reflections from rain, or cloudy weather, errors may arise.
    }
    \label{fig:RVC}
    \vspace{-5mm}
\end{figure}

\begin{wrapfigure}{r}{0.5\textwidth}
  \centering
  \vspace{-8mm}
  \includegraphics[width=\linewidth]{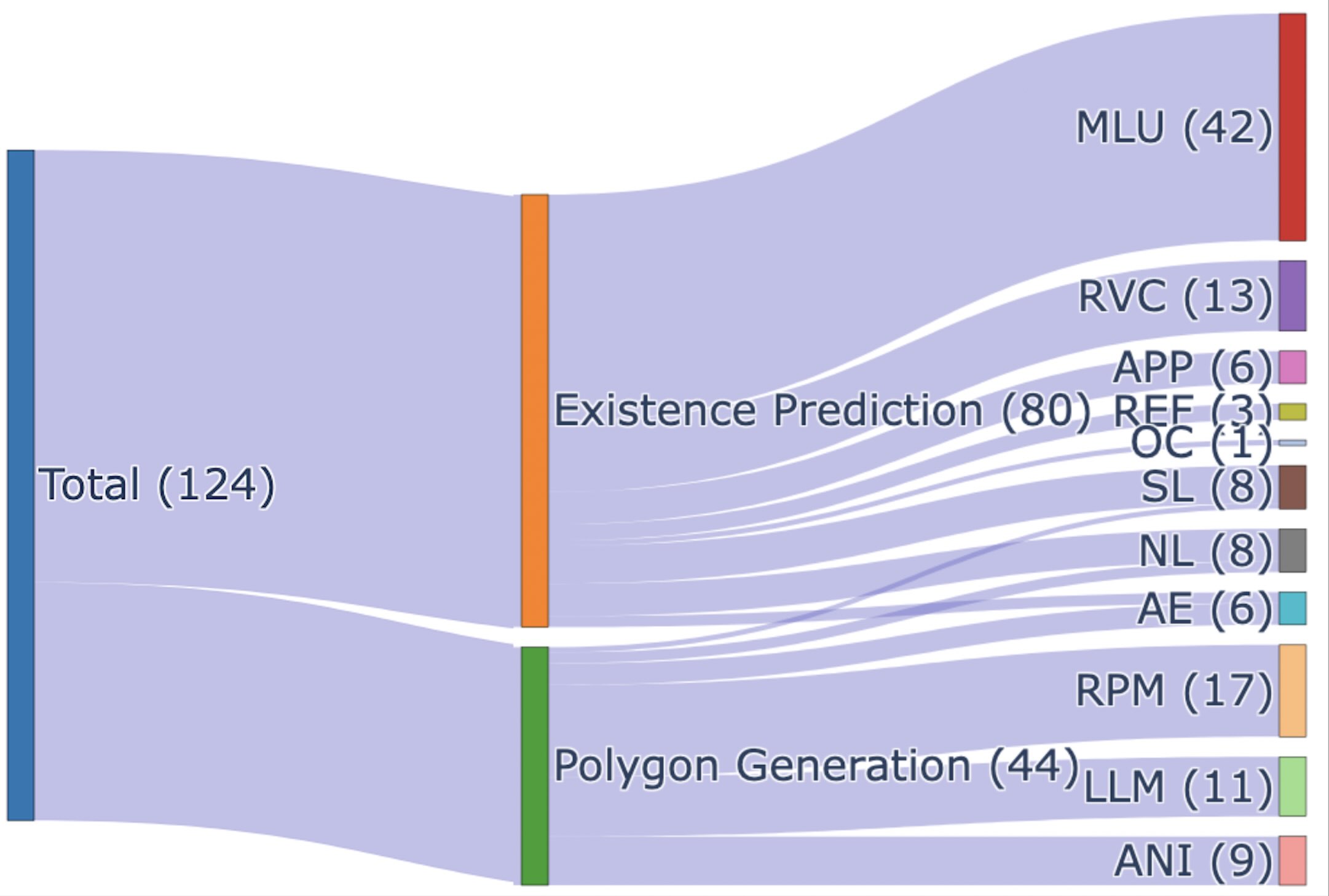}
  \caption{Categorization of failure modes. Error analysis results categorized into Cases (i) Existence Prediction and (ii) Polygon Generation.}
  \label{fig:error-analysis}
  \vspace{-8mm}
\end{wrapfigure}

\vspace{-2mm}
\section{Error Analysis}
\vspace{-2mm}
To investigate the limitations of GENNAV, we analyzed cases where the method did not perform as expected. 
In this study, failures are defined in two Cases:
(i) Existence Prediction: The prediction regarding the existence of a target region is incorrect (FP or FN).
(ii) Polygon Generation: The generated mask does not overlap with the ground truth mask of target regions (i.e., IoU = 0).
GENNAV failed in 187 samples for Case (i) and 44 samples for Case (ii) in the test set.

Fig \ref{fig:error-analysis} shows the results of the error analysis for Cases (i) Existence Prediction and (ii) Polygon Generation.
We randomly selected 40 samples each for false negative sample and true negative sample groups for Case (i). 
We classified them into the following eight categories for Case (i):
Multimodal language understanding for landmarks (MLU):
This refers to cases where the model incorrectly predicts the existence of landmarks that do not actually exist in the given image.

\noindent\textbf{Appearance misunderstanding of landmarks (AML).}
This category includes cases where errors in existence prediction occurred because of the misrecognition of appearance (e.g., although the instruction specifies ``next to the yellow car,'' the model incorrectly identified the target region as being next to the black car).

\noindent\textbf{Reduced visibility conditions (RVC).}
This category encompasses errors that arise from limited or poor visibility conditions, such as those caused by inadequate lighting, reflections from rain, or cloudy weather.
Fig. \ref{fig:RVC} shows a sample categorized as reduced visibility conditions. 
In each case, poor visual conditions are one of the factors that cause GENNAV to predict ``no-target'' despite the existence of a target region.

\noindent\textbf{Referring-expression misinterpretation (REF).} This category covers cases where visual information and navigation instruction sentences are interpreted incorrectly, particularly because of the misinterpretation of referring expressions.

\noindent\textbf{Small landmark (SL).} This involves cases where the landmark is too small or visually subtle, which causes the model to misunderstand the existence of target regions.

\noindent\textbf{No landmark (NL).} This category includes cases where errors arise because the instruction lacks any landmarks, which makes it difficult for the model to ground the action appropriately (e.g., vague instructions such as ``turn left'' or ``stay here'' without specifying a landmark).

\noindent\textbf{Annotation error (AE).} This category includes cases of discrepancies or mistakes in the annotation process itself, which can lead to misleading ground-truth data and subsequent model errors.

\noindent\textbf{Occlusion (OC).} This refers to cases where the target regions or landmarks are occluded, or located in visually accessible but physically unnavigable regions—such as across a guardrail or behind a transparent barrier.

Moreover, for Case (ii), we added three additional categories and classified the data accordingly.

\noindent\textbf{Relative position misunderstanding (RPM).}
This category includes cases where the model fails to understand the positional relationship between the landmark(s) and the intended target region appropriately (e.g., misunderstanding ``in front of'' vs. ``behind'').

\noindent\textbf{Landmark location misunderstanding (LLM).}
This refers to cases where the model misunderstands the location of one or more landmarks, leading to inaccurate polygon generation.

\noindent\textbf{Ambiguous navigation instruction (ANI).}
This refers to cases where the navigation instructions are ambiguous, because they lack sufficient clarity or specificity to uniquely identify the target region. 

Consequently, the model predicts a mask that does not overlap with the ground-truth target region, not because of incorrect reasoning, but because the ambiguity in the instruction leads to multiple plausible candidate regions.

% 6-14
As shown in Table \ref{fig:error-analysis}, the main bottleneck in Case (i) was MLU, presumably because of an insufficient semantic understanding of landmarks.
 A possible solution to address this issue is to overlay masks generated by more robust semantic segmentation models (e.g., Grounded SAM~\citep{ren2024grounded}, Detic~\citep{zhou2022detecting}, Grounding DINO~\citep{liu2023grounding} and Open-Vocabulary SAM~\citep{yuan2024ovsam}) onto the input images before handling them into LAPM. 
This approach aims to enhance the semantic understanding of landmarks and thereby mitigate the bottleneck in MLU.
By contrast, the main bottleneck for Case (ii) was RPM, which was likely to have arisen from an inadequate understanding of the 3D structure of environments, including vehicles.
A possible solution is to incorporate tasks such as vehicle orientation classification~\citep{vehicleorientation}  into the pre-training process.
These additions would strengthen the model's capability to understand 3D structural representations, thereby alleviating RPM-related limitations.

\section{Details of Real-World Experiments}

\subsection{Experimental Setup Details}

We used GPT-4o to generate navigation instructions. 
Specifically, we used Grounding DINO~\citep{liu2023grounding} to detect candidate landmarks based on the categories defined in the Talk2Car dataset, and generated visual inputs by overlaying bounding boxes on the identified landmarks. 
We then provided these images as input to GPT-4o, along with prompts instructing it to refer to detected landmarks when generating navigation instructions.
To evaluate whether the generated instructions incorporated the designated landmarks appropriately, we conducted manual inspections under both single-target and multi-target conditions. 
Additionally, for the no-target condition, we constructed samples by swapping instructions across different samples, following the approach used for the GRiN-Drive benchmark, to simulate scenarios in which no valid target landmarks were present.
In these cases, we manually verified that the generated instructions did not reference any specific landmarks inappropriately.
The instructions typically included referring expressions grounded in the visualized landmarks and described navigation tasks for directing a mobile agent to a designated destination.

\begin{table*}[t]
\centering
\setlength{\tabcolsep}{2pt}
\renewcommand{\arraystretch}{1.25}
\caption{Quantitative comparison between the proposed method and baseline methods in the real-world experiment. The best score for each metric is in bold. The ``Type'' column specify the segmentation approach of each method.}
\begin{tabular}{lcccccc}
\hline
Method & Resolution & Type & msIoU [\%]$\uparrow$ & $\mathrm{P}@0.1$ [\%]$\uparrow$ & $\mathrm{P}@0.2$ [\%]$\uparrow$ & Acc. [\%]$\uparrow$ \\
\hline
\citet{rufus2021grounding} & 224×224 & pixel   & 25.95 {$\scriptscriptstyle \pm 3.96$} & 8.50 {$\scriptscriptstyle \pm 2.60$} & 5.83 {$\scriptscriptstyle \pm 3.28$} & 41.83 {$\scriptscriptstyle \pm 8.09$} \\
LAVT \citep{yang2022lavt} & 224×224 & pixel   & 22.84 {$\scriptscriptstyle \pm 3.35$} & 10.00 {$\scriptscriptstyle \pm 3.06$} & 6.00 {$\scriptscriptstyle \pm 2.05$} & 55.00 {$\scriptscriptstyle \pm 5.17$} \\
TNRSM \citep{tnrsmral24} & 224×224 & pixel   & 23.11 {$\scriptscriptstyle \pm 5.88$} & 24.25 {$\scriptscriptstyle \pm 4.11$} & 13.25 {$\scriptscriptstyle \pm 4.56$} & 56.83 {$\scriptscriptstyle \pm 5.08$} \\
\hline
\citet{rufus2021grounding} & 640×640 & pixel   & 21.68 {$\scriptscriptstyle \pm 5.95$} & 3.50 {$\scriptscriptstyle \pm 2.97$} & 2.00 {$\scriptscriptstyle \pm 2.54$} & 40.42 {$\scriptscriptstyle \pm 6.02$} \\
LAVT \citep{yang2022lavt}  & 640×640 & pixel   & 30.44 {$\scriptscriptstyle \pm 1.63$} & 11.73 {$\scriptscriptstyle \pm 10.69$} & 5.00 {$\scriptscriptstyle \pm 5.80$} & 32.67 {$\scriptscriptstyle \pm 19.74$} \\
TNRSM \citep{tnrsmral24} & 640×640 & pixel   & 28.94 {$\scriptscriptstyle \pm 4.98$} & 7.25 {$\scriptscriptstyle \pm 12.10$} & 4.50 {$\scriptscriptstyle \pm 8.69$} & 47.43 {$\scriptscriptstyle \pm 18.60$} \\
\hline
Gemini~\citep{reid2024gemini} & 1600×900 & bbox    & 4.81 {$\scriptscriptstyle \pm 0.40$} & 4.25 {$\scriptscriptstyle \pm 0.69$}  & 3.00 {$\scriptscriptstyle \pm 0.69$} & 61.67 {$\scriptscriptstyle \pm 1.44$} \\
Qwen2-VL~\citep{Qwen2-VL} & 1600×900 & bbox    & 20.48 {$\scriptscriptstyle \pm 2.33$} & 5.75 {$\scriptscriptstyle \pm 2.59$}  & 1.50 {$\scriptscriptstyle \pm 1.05$} & 66.33 {$\scriptscriptstyle \pm 3.36$} \\
Qwen2-VL~\citep{Qwen2-VL} & 1600×900 & polygon & 11.17 {$\scriptscriptstyle \pm 0.46$} & 0.00 {$\scriptscriptstyle \pm 0.00$}  & 0.00 {$\scriptscriptstyle \pm 0.00$} & 50.33 {$\scriptscriptstyle \pm 0.75$} \\
\hline
\textbf{GENNAV (ours)} & 640×640 & polygon & \textbf{34.32 {$\scriptscriptstyle \pm 2.97$}} & \textbf{28.75 {$\scriptscriptstyle \pm 5.08$}} & \textbf{16.75 {$\scriptscriptstyle \pm 2.27$}} & \textbf{67.50 {$\scriptscriptstyle \pm 2.95$}} \\
\hline
\end{tabular}
\label{tab:quantitative-physical-full}
\vspace{-3mm}
\end{table*}
\subsection{Additional Quantitative Results} Table \ref{tab:quantitative-physical-full} shows the overall quantitative comparison between GENNAV and baseline methods in the real-world experiment.
The values in the table are the average and standard deviation over five trials.
The "Type" column specifies the segmentation approach of each method.
The msIoU of GENNAV and the baseline methods were as follows, grouped by input resolution:

In the 224×224 setting, the method by~\citet{rufus2021grounding}, LAVT, and TNRSM achieved msIoU scores of 25.95, 22.84, and 23.11, respectively.
Under the 640×640 condition, GENNAV achieved an msIoU of 34.32, whereas the method by Rufus et al., LAVT, and TNRSM achieved scores of 21.68, 30.44, and 28.94, respectively.
Notably, GENNAV outperformed the MLLMs that had high-resolution inputs.
These results demonstrate that GENNAV achieved the highest msIoU among the baseline methods.

Similarly, the $\mathrm{P}@0.1$ scores of the baseline methods by~Rufus et al., LAVT, and TNRSM in the 224×224 setting were 8.50, 10.00, and 24.25, respectively.
 At the 640×640 resolution, the $\mathrm{P}@0.1$ scores of GENNAV and the method by~Rufus et al., LAVT, and TNRSM were 28.75, 3.50, 11.73, and 7.25, respectively.
 Furthermore, the MLLM baselines, that is, Gemini, Qwen2-VL (bbox), and Qwen2-VL (polygon), achieved $\mathrm{P}@0.1$ scores of 4.25, 5.75, and 0.00, respectively.
These results demonstrate that GENNAV outperformed the baseline methods in terms of $\mathrm{P}@0.1$, demonstrating a 4.5 point improvement over the best-performing baseline, TNRSM (224×224).
Overall, GENNAV achieved higher accuracy (67.50\%) than the baseline methods in the real-world experiments conducted in a zero-shot manner.  

\subsection{Additional Qualitative Results} 
\begin{figure}[t]
    \centering
    \begin{minipage}{1.0\linewidth}
            \vspace{0pt}
            % Row 1
            \begin{minipage}{0.49\linewidth}
                \includegraphics[width=\linewidth]{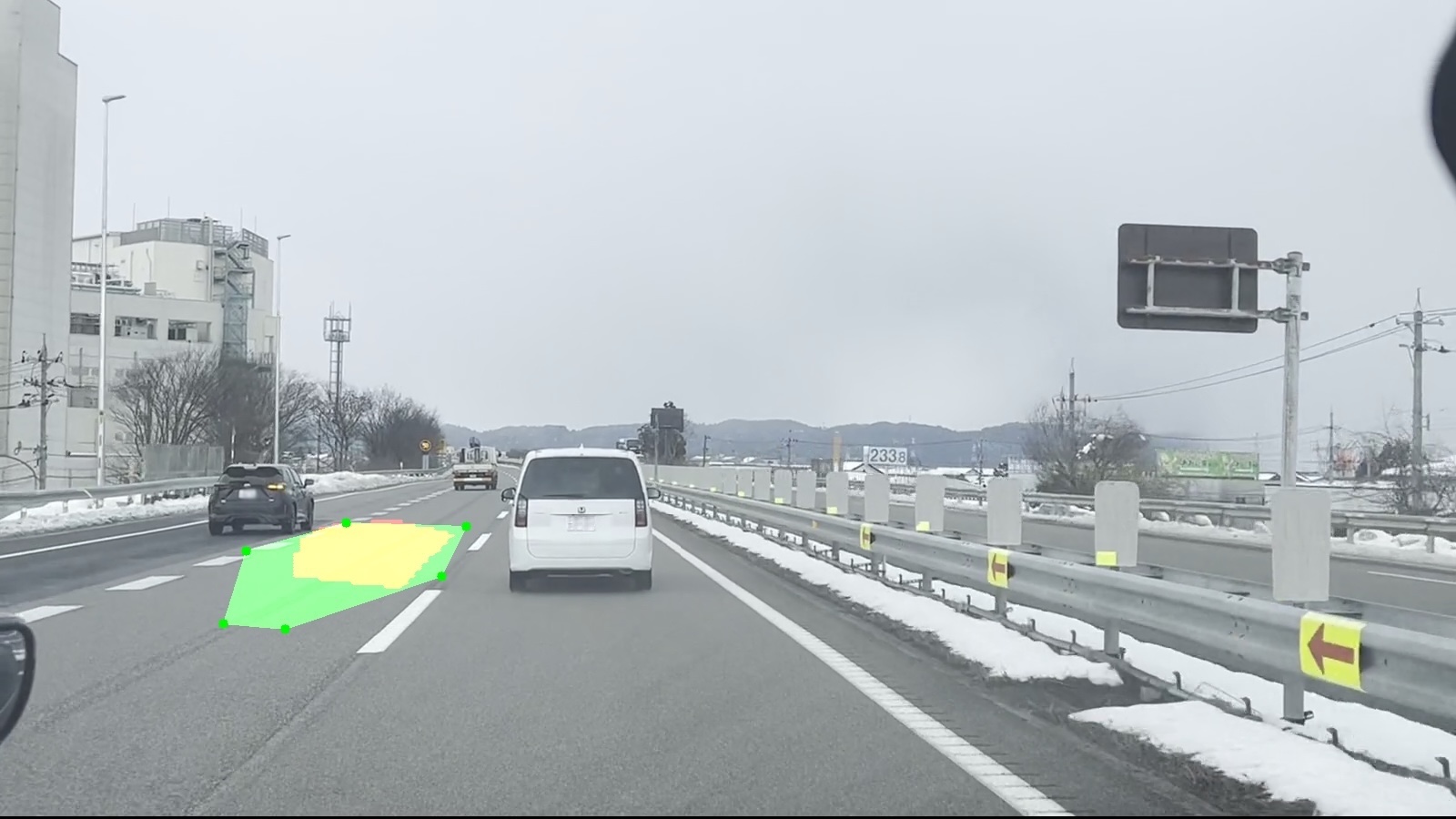}
            \end{minipage}
            \hfill
            \begin{minipage}{0.49\linewidth}
                \includegraphics[width=\linewidth]{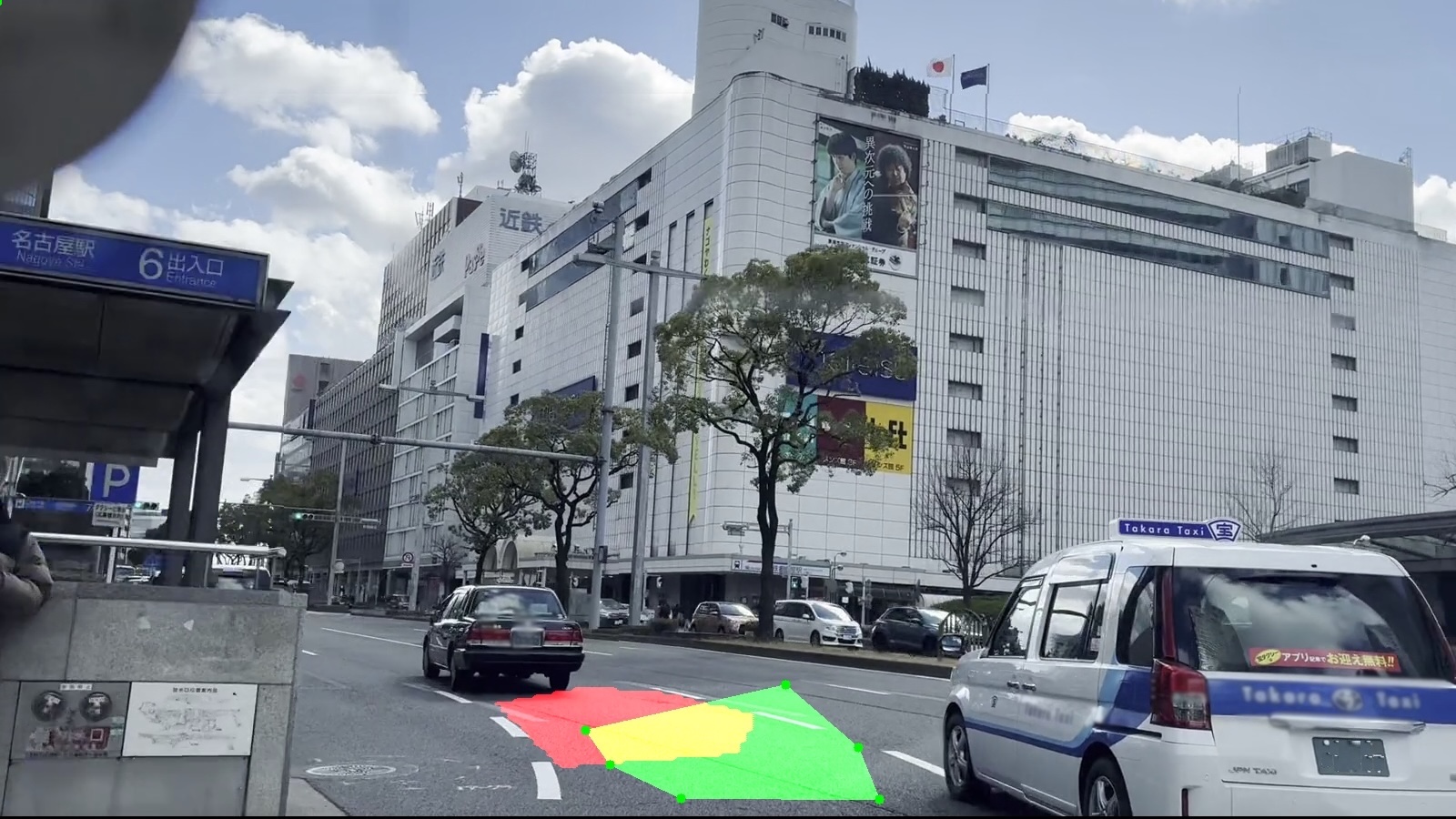}
            \end{minipage}
            \begin{minipage}[t]{0.5\linewidth}
                \vspace{3pt}
                \centering {(i)~$\bm{x}_\text{inst}$: ``Park next to the black car moving on the left lane.''}
            \end{minipage}
            \begin{minipage}[t]{0.5\linewidth}
                \vspace{3pt}
                \centering {(ii)~$\bm{x}_\text{inst}$: ``Stop behind a moving vehicle in the traffic lane.''}
            \end{minipage}
            % Row 2
             \begin{minipage}{0.49\linewidth}
                \includegraphics[width=\linewidth]{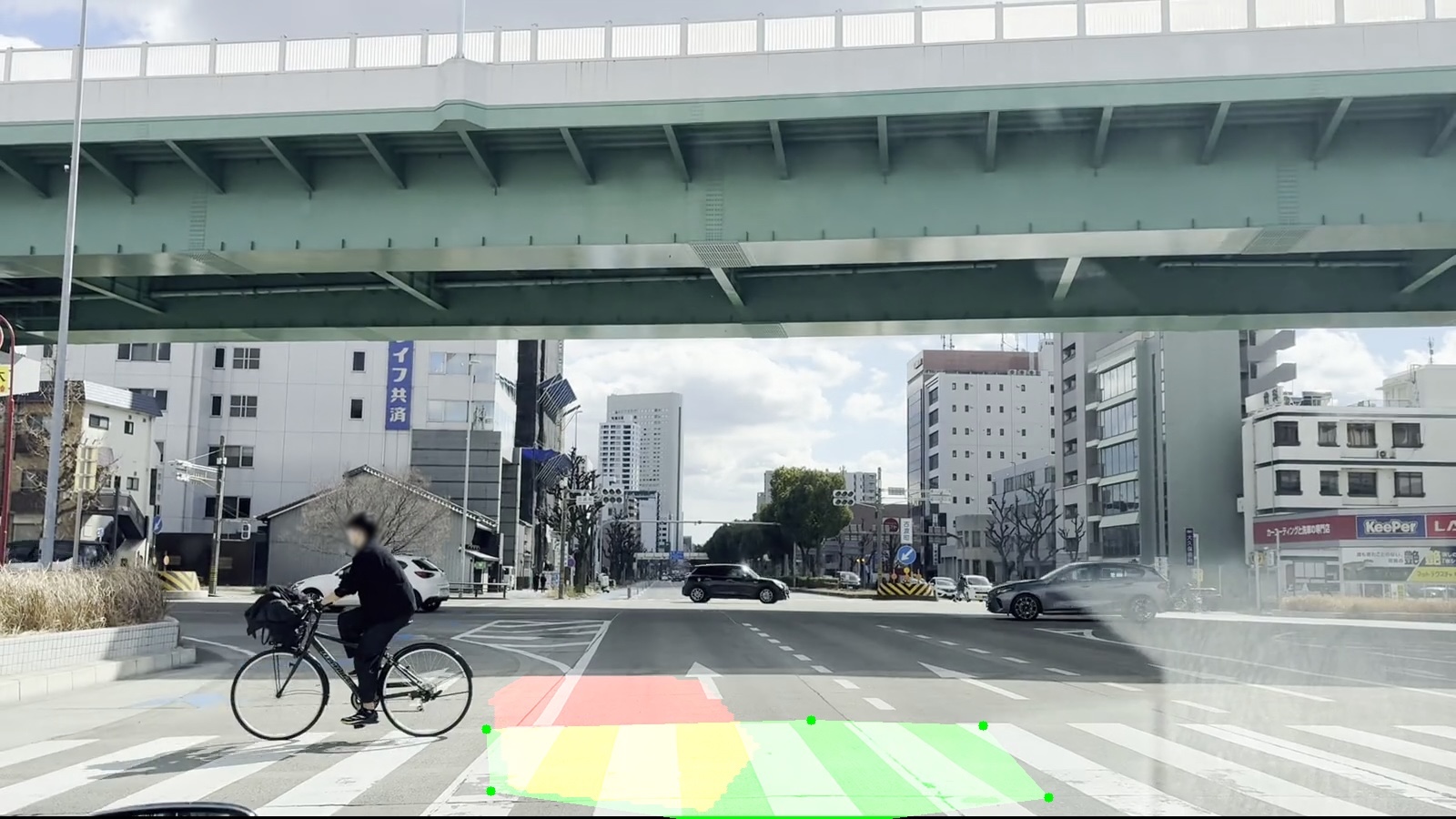}
            \end{minipage}
            \hfill
            \begin{minipage}{0.49\linewidth}
                \includegraphics[width=\linewidth]{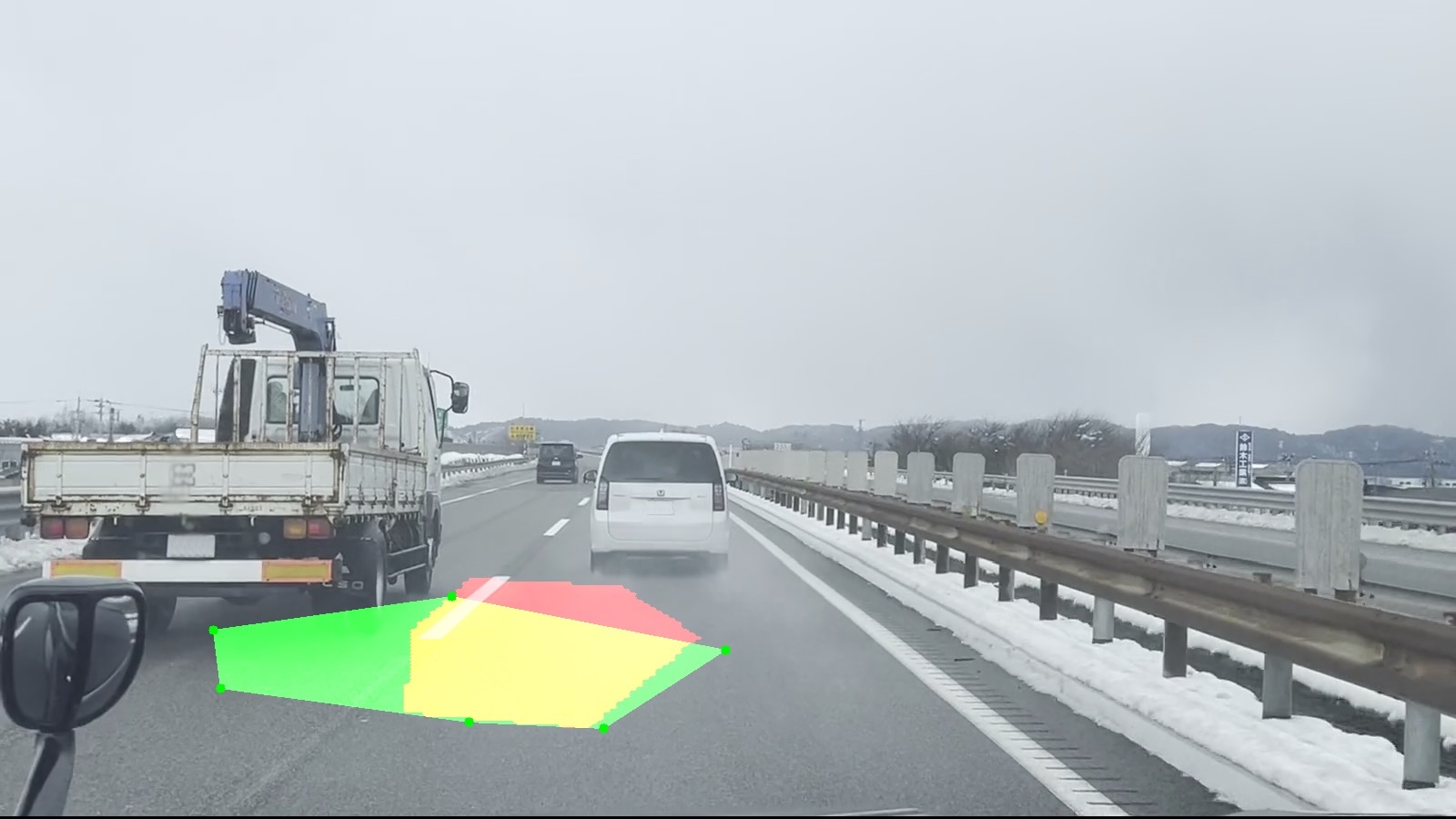}
            \end{minipage}
            \begin{minipage}[t]{0.5\linewidth}
                \vspace{3pt}
                \centering {(iii)~$\bm{x}_\text{inst}$: ``Stop around the cyclist crossing the street.''}
            \end{minipage}
            \begin{minipage}[t]{0.5\linewidth}
                \vspace{3pt}
                \centering {(iv)~$\bm{x}_\text{inst}$: ``Stop to the right of the large truck with the open bed.''}
            \end{minipage}

             \begin{minipage}{0.49\linewidth}
                \includegraphics[width=\linewidth]{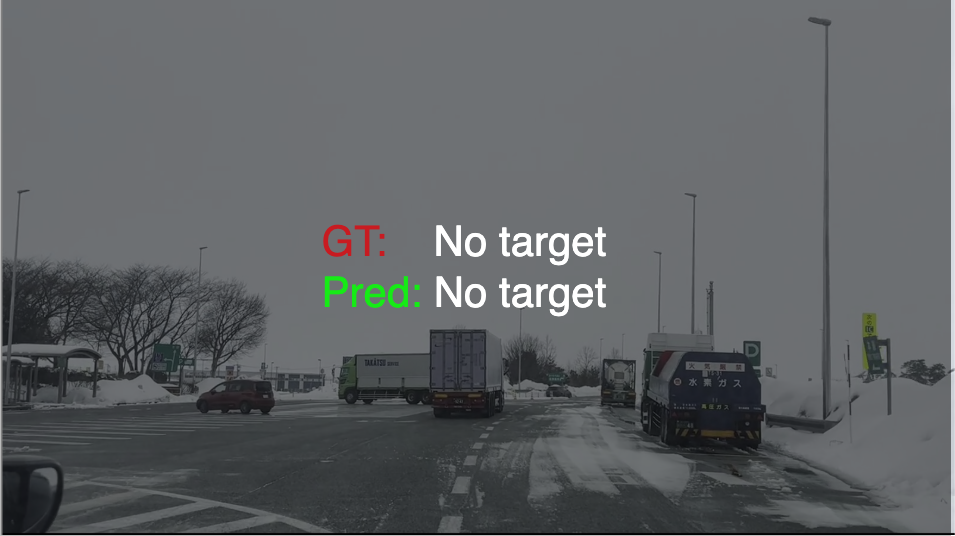}
            \end{minipage}
            \hfill
            \begin{minipage}{0.49\linewidth}
                \includegraphics[width=\linewidth]{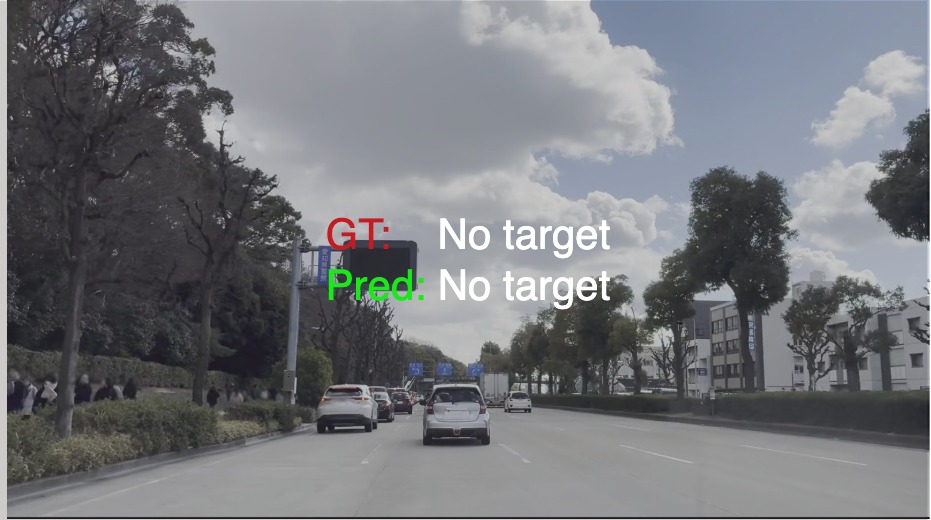}
            \end{minipage}
            \begin{minipage}[t]{0.5\linewidth}
                \vspace{3pt}
                \centering {(v)~$\bm{x}_\text{inst}$: ``Park behind that truck that is way up ahead.''}
            \end{minipage}
            \begin{minipage}[t]{0.5\linewidth}
                \vspace{3pt}
                \centering {(vi)~$\bm{x}_\text{inst}$: ``Stop to the left of the silver vehicle in the left lane.''}
            \end{minipage}
            % Row 2
             \begin{minipage}{0.49\linewidth}
                \includegraphics[width=\linewidth]{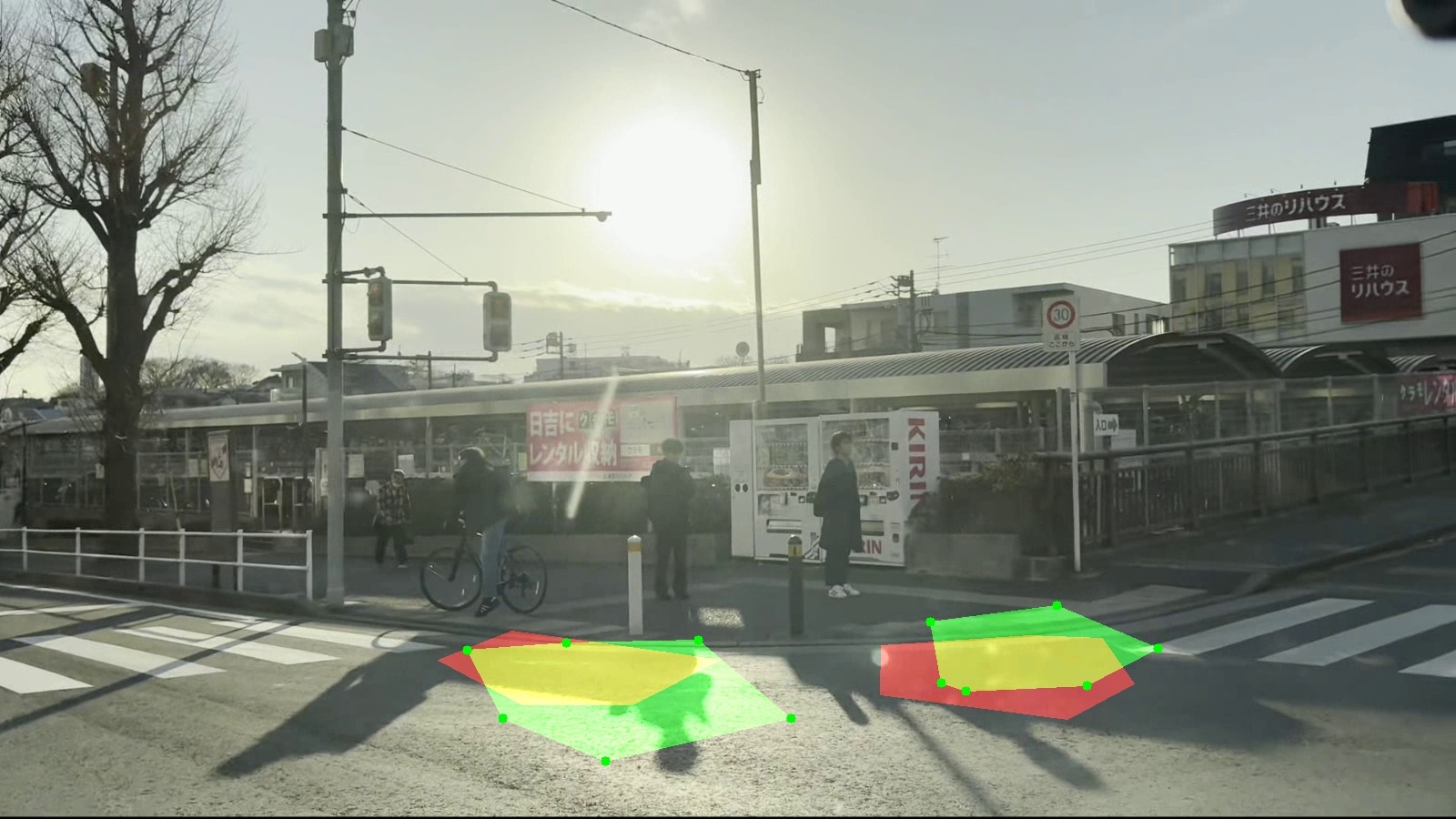}
            \end{minipage}
            \hfill
            \begin{minipage}{0.49\linewidth}
                \includegraphics[width=\linewidth]{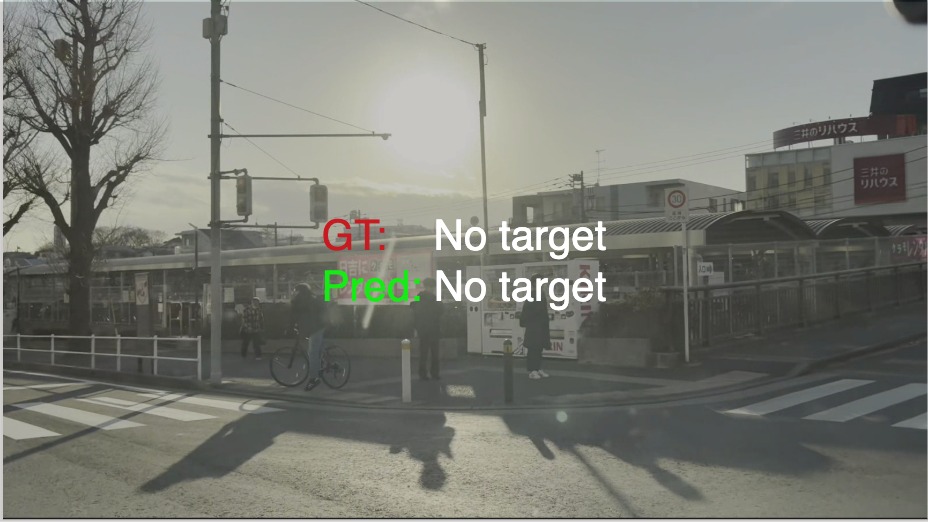}
            \end{minipage}
            \begin{minipage}[t]{0.5\linewidth}
                \vspace{3pt}
                \centering {(vii)~$\bm{x}_\text{inst}$: ``Park near the person standing near the intersection.''}
            \end{minipage}
            \begin{minipage}[t]{0.5\linewidth}
                \vspace{3pt}
                \centering {(viii)~$\bm{x}_\text{inst}$: ``Please follow the vehicle in front.''}
            \end{minipage}
            % Row 2
        \end{minipage}

    \caption{
      Additional qualitative results of the proposed method in the real-world experiments. The green and red regions indicate the predicted and ground-truth regions, respectively, while the yellow region represents the overlap between the predicted and ground-truth regions.
    }
    \vspace{-4mm}
    \label{fig:supp-qualitative-physical}
\end{figure}

Fig.~\ref{fig:supp-qualitative-physical} provides additional success examples of GENNAV on the real-world experiment. 
These results indicate that GENNAV can be effectively integrated into real-world systems.
Furthermore, as shown in Fig.~\ref{fig:supp-qualitative-physical}~(vii) and Fig.~\ref{fig:supp-qualitative-physical}~(viii), GENNAV is capable of generating different predictions for the same image when provided with different instructions.

% \clearpage
% \bibliography{reference}  % .bib

\end{document}